\documentclass[10pt,twocolumn,letterpaper]{article}
\usepackage{wacv}              
\usepackage{graphicx}
\usepackage{amsmath}
\usepackage{amssymb}
\usepackage{booktabs}
\usepackage{multirow}
\usepackage[pagebackref,breaklinks,colorlinks]{hyperref}
\usepackage[capitalize]{cleveref}
\crefname{section}{Sec.}{Secs.}
\Crefname{section}{Section}{Sections}
\Crefname{table}{Table}{Tables}
\crefname{table}{Tab.}{Tabs.}
\def\confName{WACV}
\def\confYear{2025}
\usepackage{xspace}
\newcommand{\cc}{Crowd Counting\xspace}
\newcommand{\density}{Volume Density Maps\xspace}
\newcommand{\ppdensity}{Per-Part Volume Density Maps\xspace}

\newcommand{\ourdataset}{ANTHROPOS-V\xspace} 

\newcommand{\imagei}{\ensuremath{I}}

\newcommand{\gtcounti}{Oracular\xspace\ensuremath{C(\imagei)}}
\newcommand{\datasetd}{\ensuremath{\mathcal{D}}\xspace}
\newcommand{\meanvold}{\ensuremath{\Bar{V}_{\datasetd}}\xspace}
\newcommand{\commas}{``}

\def\npourstrain{495}
\def\npoursval{64}
\def\npourstest{142}

\usepackage[dvipsnames]{xcolor}
\usepackage{cuted}

\begin{document}

\title{ANTHROPOS-V: benchmarking the novel task  of Crowd Volume Estimation}

\author{
    Luca Collorone$^*$, \hspace{1em} Stefano D'Arrigo$^*$, \hspace{1em} Massimiliano Pappa$^*$, \\  Guido M. D'Amely di Melendugno, \hspace{1em}
    Giovanni Ficarra, \hspace{1em} Fabio Galasso\\
    Sapienza University of Rome\\
    {\tt\small \{name.surname\}@uniroma1.it}\\
}

\maketitle
\renewcommand{\thefootnote}{\fnsymbol{footnote}} %
\footnotetext[1]{Authors contributed equally.}
\setcounter{footnote}{0}
\renewcommand{\thefootnote}{\arabic{footnote}}

\begin{abstract}

\noindent We introduce the novel task of \textbf{Crowd Volume Estimation (CVE)}, defined as the process of estimating the collective body volume of crowds using only RGB images. Besides event management and public safety, CVE can be instrumental in approximating body weight, unlocking weight-sensitive applications such as infrastructure stress assessment, and assuring even weight balance.
We propose the first benchmark for CVE, comprising \emph{\ourdataset}, a synthetic photorealistic video dataset featuring crowds in diverse urban environments.
Its annotations include each person's volume, SMPL shape parameters, and keypoints.
Also, we explore metrics pertinent to CVE, define baseline models adapted from Human Mesh Recovery and Crowd Counting domains, and propose a CVE-specific methodology that surpasses baselines. Although synthetic, the weights and heights of individuals are aligned with the real-world population distribution across genders, and they transfer to the downstream task of CVE from real images. 
Benchmark and code are available at \href{https://github.com/colloroneluca/Crowd-Volume-Estimation}{github.com/colloroneluca/Crowd-Volume-Estimation}.
\end{abstract}
\section{Introduction}

\noindent Dealing with large gatherings in public spaces presents significant challenges in crowd management: \textit{overcrowding} can jeopardize the safety, health, and comfort of individuals, while the assembly of crowds on structures not designed for high capacity poses risks of structural damage or collapse due to \textit{overloading} \cite{thompson1994developing}. 

Currently, the monitoring of crowds' risks based on head count \cite{sudharson2023proactive, fiandeiro2023modernized} tends to disregard potentially critical factors such as weight, occupancy, heat dissipation, and oxygen consumption, which are strongly correlated with individuals' body build~\cite{thompson1994developing, kappagoda1979comparison}. 
Additionally, these factors exhibit significant variability based on age, gender, ethnicity, and health conditions~\cite{jenkins2003patterns, 10.1093/oxfordjournals.aje.a008733, deurenberg1998body}.

A precise estimate of the \textit{crowd's total volume} offers a more reliable method for detecting space underuse or overcrowding by leveraging a priori knowledge of the available in-place volume. Additionally, this approach can significantly mitigate the risks associated with overloading, as volume serves as a robust proxy for estimating weight~\cite{durnin1974body, PfitznerLibra3D2015}.

\begin{figure}[!t]
    \centering
    \includegraphics[width=1\linewidth]{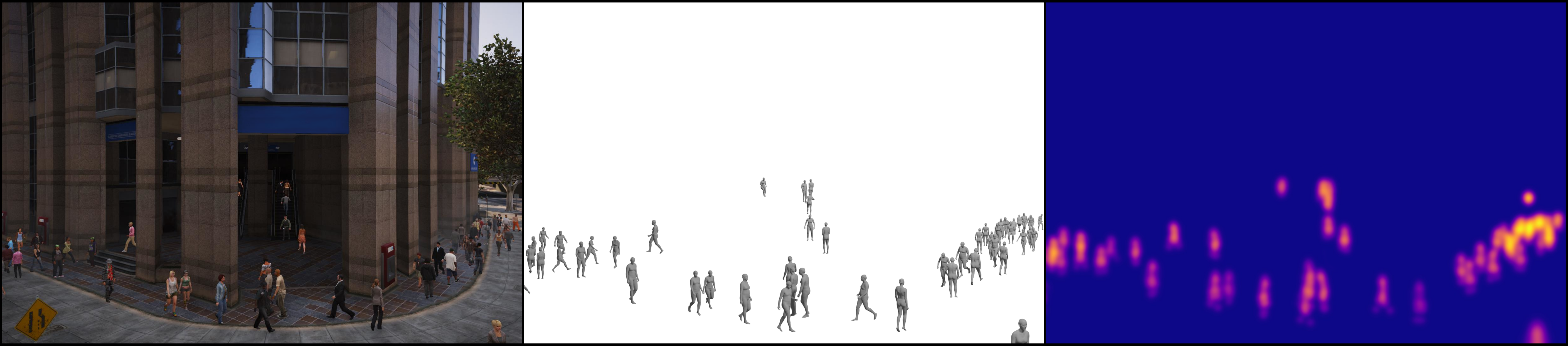}
    \caption{\ourdataset is the first dataset for the novel task of Crowd Volume Estimation. It features human crowds engaged in activities within real-world environments ({\it left}). Each individual in the dataset is labeled with ground-truth SMPL shape parameters ({\it center}) along with their volume labels. We employ novel \ppdensity as a superior supervision signal for training models to address this task ({\it right}).}
    \label{fig:teaser}
\end{figure}

Motivated by these insights, this paper introduces the novel \textit{Crowd Volume Estimation}  (CVE) task, which aims to estimate the collective volume occupied by groups of individuals directly from single RGB images. 

Recently, volume estimation (VE) has garnered significant research attention. However, previous works \cite{hu2023Human,leinen2021volnet,liu2017estimation,PfitznerLibra3D2015,pirker2009human} have focused primarily on estimating the volume of single individuals in controlled environments, often relying on expensive annotations.
These works rely on requirements and assumptions that render them impractical for estimating the volume of large crowds.
In addition, there are no datasets tailored for CVE, as existing datasets~\cite{patel2021agora,black2023bedlam} featuring scenes with multiple people only encompass a limited number of individuals ($\leq 15$) or lack necessary volume annotations~\cite{fabbri2018learning, fabbri2021motsynth}. We thus propose the first CVE benchmark, including baselines, a novel dataset, and metrics.\\
As for the baselines, we outline two research directions for crafting models for CVE: (1) grounding on methods from the Crowd Counting~\cite{song2021rethinking,hani2023steerer,lin2022boosting,ma2019bayesian} domain, or (2) repurposing the pipeline of Human Mesh Recovery (HMR)~\cite{wang2023refit,li2022cliff,black2023bedlam} combined with a human detector. In the case of (1), we extend the density-map strategy to a more specific form of supervision, the \textit{\density}, tailored to regress the human volumes in an image. 
For these models, we further define a more fine-grained form of supervision which we dub {\it \ppdensity} (rightmost picture in Fig.~\ref{fig:teaser}). This supervision allows the model to learn how to regress the volume of different human subparts (e.g., arms, chest, legs), resulting in a more accurate estimate of each individual's volume. In the case of (2), we show that HMR models with a human detection preprocessing can serve as volume estimators without any modification, as the volume estimate comes as a by-product after postprocessing the predicted meshes (middle picture in Fig.~\ref{fig:teaser}).

Moreover, to enable the training of CVE-specific models, we introduce ``ANTHROpometrics POse Shape and Volume estimation dataset'' (\ourdataset), a synthetic, large-scale, and video-realistic dataset representing large crowds in urban scenarios and reporting for the first time the annotated volume of each individual appearing in the scenes. 
\ourdataset is generated using the videogame engine of Grand Theft Auto V (GTA-V), which includes a large variety of realistic urban environments and a broad diversity of characters' ethnicities.

It is worth noting that acquiring real-world data with annotated anthropometric features for crowds requires gathering sensitive information from thousands of individuals in a wide range of environments, hence posing severe issues of feasibility, bias, and personal details disclosure. 
Thus, we collect a synthetic dataset to train models on CVE, evaluating their learned knowledge of both synthetic and real-world data.

Aiming at reducing the domain gap from synthetic to real images, we enhance the game's appearance and perform an in-depth analysis of the GTA-V default characters. Our findings reveal that the default characters exhibit a restricted and repetitive assortment of human anthropometrics, such as body sizes and heights. Therefore, we directly manipulate characters' 3D meshes to align them with the real-world human size distributions~\cite{owid-human-height}. As a result, we improve both the realism of the original GTA-V scenes and their annotations, featuring virtual characters whose height and weight distributions closely follow the authentic human variations~\cite{silverman2022exact}.

In summary, our contributions are four-fold:
\begin{itemize}
    \item we propose the novel task of \textit{Crowd Volume Estimation} (CVE) to regress the volume of large groups of people from RGB images;
    \item we release the first CVE benchmark, including metrics and baselines;
    \item we introduce \ourdataset, a dataset explicitly devised for CVE but also encompassing annotations for other human-centric tasks, with careful attention to mirror real-world anthropometric and gender statistics;
    \item we experiment with a novel volume-specific form of supervision, namely Per-Part Volume Density Maps, and use it to train our proposed model, STEERER-V, achieving superior results.
\end{itemize}

\section{Related Works}
 In this section, we review the existing literature that relates to the proposed CVE task. We discuss studies on single-subject Volume Estimation (VE) (Sec.~\ref{subsec:VolumeModels}), Crowd-Counting (Sec.~\ref{subsec:Counting}) and Human Mesh Recovery (Sec.~\ref{subsec:rwhmr}).

\subsection{Single subject VE in controlled environments}
\label{subsec:VolumeModels}
\noindent Previous literature explored volume estimation, targeting single-bodies~\cite{hu2023Human,leinen2021volnet,liu2017estimation,PfitznerLibra3D2015,pirker2009human} or objects~\cite{stereo2020,dehais2017two-view,lo2018food,lo2019point2volume,puri2009recognition,xu2013model,yang2021human} for applications in healthcare and nutrition.  
In particular, \cite{liu2017estimation} rely on 3D scans, while \cite{hu2023Human,lo2018food,lo2019point2volume,PfitznerLibra3D2015} exploit depth-maps or point clouds data.  
To deal with scale ambiguity, \cite{dehais2017two-view, hassannejad2017new,puri2009recognition,xu2013model,yang2021human} make use of reference objects, while \cite{stereo2020,pirker2009human} employ multiple images of the same object in different views. 
While all the mentioned works tightly depend on controlled environments, scans, or multiple inputs, and apply to a single subject at a time, we aim at estimating the total volume of human crowds in the wild. Notably, \cite{leinen2021volnet} proposed a large-scale video dataset displaying individual textured SMPL meshes \cite{loper2015smpl}, paired with body-part volume ground truths. However, scenes are designed by superimposing a single mesh onto 2D bedroom images, lacking realism and scale consistency between humans and the background. On the contrary, we propose a dataset of realistic scenes featuring large crowds.

\subsection{\cc}
\label{subsec:Counting}
\noindent \cc aims to estimate the number of people in images or videos. Typically, datasets in this domain showcase large crowds from bird-eye views \cite{idrees2013multi, zhang2016single, Idrees_2018_ECCV, sindagi2019pushing, sindagi2020jhu, wang2020nwpu}.
While seminal works~\cite{cho1999neural, marana1997estimation, kong2006viewpoint} cast this problem as a regression task,
recent literature~\cite{song2021rethinking, liang2022end, lin2022boosting} address crowd counting as a localization task. These methods regress the 2D positions of the heads in the images and estimate the total number by summing the retrieved outcomes after filtering the more uncertain predictions. 
Density-Map-based methods \cite{zhang2017fcn, li2018csrnet, Wang_2019_CVPR, wang2020distribution, ma2021learning, wan2021generalized, cheng2021decoupled, liu2022countr, hani2023steerer,ranasinghe2024crowddiff} express the ground truth density map $y$ for an image $x$ as a single-channel image of the same size, where each pixel is assigned $1$ if it contains the center of a person's head, $0$ otherwise; $y$ is subsequently smoothed with a Gaussian filter.
The Gaussian filtering operation is common in Counting tasks~\cite{liu2022countr, lempitsky2010learning, ranjanLearningCountEverything2021,
youFewshotObjectCounting2022, thanasutivesEncoderDecoderBasedConvolutional2021}, as it allows to treat the GT density map as a continuous function \cite{zhang2016single}, which, in turn, allows the end-to-end training of the network.
Bayesian-based approaches differ from conventional density-based methods as they estimate density maps without supervision upon ground truth density maps. For instance, \cite{lin2022boosting, ma2019bayesian} employ a bayesian-loss to construct a density contribution probability model starting from bare annotations.
We evaluate strong Density-Map \cite{hani2023steerer}, Localization \cite{song2021rethinking}, and Bayesian models \cite{ma2019bayesian, lin2022boosting} for the CVE task. 
We introduce baselines capitalizing on an adaptation of the Density Maps approach for CVE, namely \density: each pixel containing a person's head is assigned with the whole person's body volume instead of \(1\). 

\subsection{Human Mesh Recovery for Few Individuals} \label{subsec:rwhmr}
Human Mesh Recovery (HMR) regresses the human 3D shape and pose from single RGB images. CLIFF~\cite{li2022cliff} pairs cropped image features with their bounding box information, enabling the accurate prediction of global rotations. In BEDLAM-CLIFF~\cite{black2023bedlam}, the authors train CLIFF on their dataset and improve its performance. ReFit~\cite{wang2023refit} exploits a recurrent updater that iteratively adjusts a parametric human model to align with image features. The recent TokenHMR~\cite{dwivedi2024tokenhmr} uses a tokenized representation of the human pose and reformulates the problem as a token prediction. DPMesh~\cite{zhu2024dpmesh} leverages a diffusion model to meliorate robustness to occlusions. 
Crowd3DNet~\cite{wen2023crowd3d} focuses on mesh reconstruction of people within crowds in wide-field images, though tightly assuming the existence of a common plane where all the actors lie; such assumption does not hold for the complex scenes of \ourdataset. Similarly, \cite{huang2023reconstructing} exploits pseudo-GT to model the relations and interactions of the individuals and improve pose and localization estimation; however, this work does not focus on human shape, as no shape-related metric is employed. Contrarily, CVE requires precise volume/shape GT for correct computation (Sec. \ref{metrics}).
Thus, we repurpose \cite{li2022cliff,black2023bedlam,wang2023refit} for CVE, pairing them with a human detector.\\

\vspace{-1em}
\section{Measuring Crowds Volumes}
\label{sec:CVE}

In this section, we formalize CVE (cf. Sec.~\ref{problem_formalization}) and define the metrics for the novel CVE benchmark (cf. Sec.~\ref{metrics}).

\subsection{Problem Formalization}
\label{problem_formalization}

\noindent We define \textbf{Crowd Volume Estimation} as the task of estimating the undergarment total body volume occupied by human bodies in a given scene. While CVE can be applied to videos (that we make available in \ourdataset{}), we define the CVE task to be benchmarked per frame. 

Let $I$ be an image and $V_{tot}$ the label of the actual total volume of human bodies represented in $I$. We define the objective of CVE as  $\displaystyle\min_{\theta}||V_{tot} - M_{\theta}(I)||$, where $M_\theta$ is a crowd volume estimation function parameterized by $\theta$.
This definition is intentionally general and designed to be independent of any specific methodology. Indeed, this formulation enables its application as an objective for both \cc and HMR models, as well as for our proposed method.

\begin{figure}[t!]
    \centering
    \includegraphics[width=0.75\linewidth]{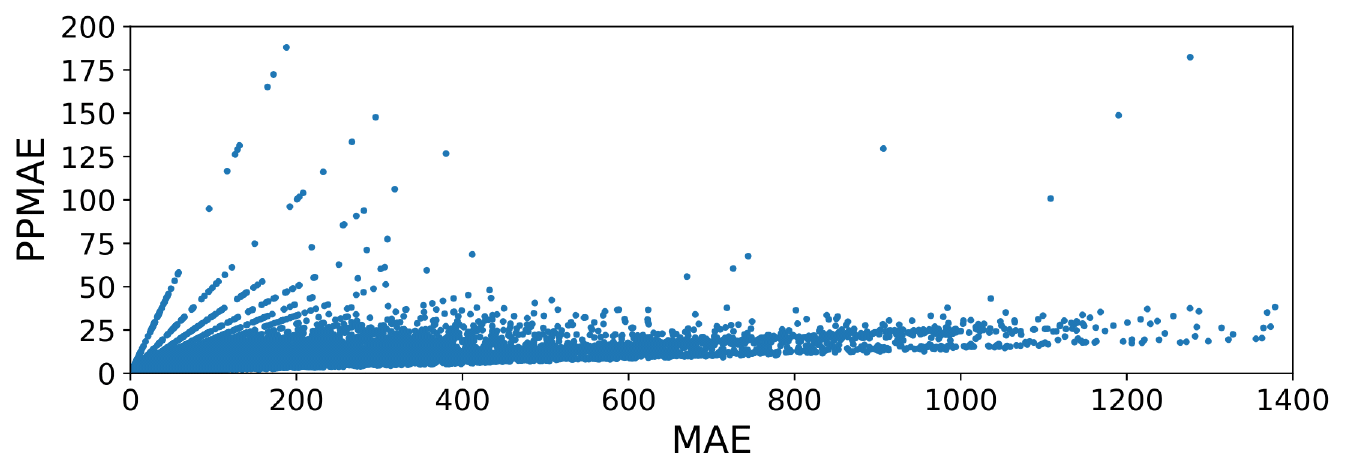}
    \caption{STEERER-V's PP-MAE$/$MAE computed on \ourdataset{} test set.}
    \label{fig:mae_vs_ppmae}
    \vspace{-1.0em}
\end{figure}

\subsection{Proposed Metrics}
\label{metrics}
Leveraging earlier research in \cc~\cite{zhang2016single, Idrees_2018_ECCV, wang2020nwpu, hani2023steerer}, we propose to measure the volume estimation error with the following suite of metrics:\\
\noindent \textit{Mean Absolute Error} (MAE), as a standard measure to assess the quality of the estimations.
Given a set of images \(\{I_k\}\), we indicate with \(\{V_k\}\) the total volume associated to each image and with \(\{\hat{V}_k\}\) the estimated one. Hence:
\begin{equation}
        \mathrm{MAE} = \frac{1}{K} \sum_{k=1}^{K} |V_k - \hat{V}_k|
    \label{eq:mae}
\end{equation}
\noindent This measure estimates the accuracy of the predictions on the whole test set as $K$ represents the total count of the images belonging to the test set.\\
\noindent \textit{Per-Person Mean Absolute Error} (PP-MAE), to measure the average error for each individual. Its formulation can be expressed as:
\begin{equation}
    \mathrm{PP\text{-}MAE} =  \frac{1}{K} \sum_{k=1}^{K} \frac{|V_k - \hat{V}_k|}{n_k}
\end{equation}
where $n_k$ is the number of persons in the $k$-th image.
PP-MAE highlights the estimator's mean error per person, enabling evaluation across scenes and datasets with different numbers of individuals. This is advantageous in CVE, as crowd sizes vary widely among frames.
It is worth noting that PP-MAE is related to the Normalized Absolute Error metric used in Crowd Counting \cite{hani2023steerer}, which normalizes the mean counting error by the total number of persons.

\noindent Although PP-MAE and MAE appear to be closely related, Fig.~\ref{fig:mae_vs_ppmae} confirms the absence of a direct correlation between the two metrics. 
The lines composed of aligned points in Fig.~\ref{fig:mae_vs_ppmae} correspond to frames with a fixed human count (as their slope MAE$/$PP-MAE equals the number of individuals). Thus, comparing MAE vs. PP-MAE facilitates identifying the frames whose estimation error marginally depends on the number of individuals and more likely stems from other latent variables (\eg  camera position, weather and lighting condition).
\section{Crowd Volume Estimation}\label{sec:exp_CVE}

In this section, we outline how we repurpose Human Mesh Recovery (HMR) (Sec.~\ref{sec:hmr_volume}) and adapt Crowd Counting (Sec.~\ref{sec:cc_volume}) baselines for our task. 
In Sec.~\ref{sec:pp_volume}, we describe the intuitions and methodologies of our proposed approach.

\subsection{From HMR to CVE}\label{sec:hmr_volume}

We adapt CLIFF~\cite{li2022cliff}, BEDLAM-CLIFF~\cite{black2023bedlam}, and ReFit~\cite{wang2023refit} to the task of CVE according to the following pipeline: (1)~identifying human occurrences using a human detector (HD) model, (2)~determining the mesh for each individual in the scene, and (3)~calculating the volume of each mesh. Finally, the total crowd volume is obtained by aggregating the individual volumes. We dub these baselines as HD+HMR. A shortcoming of this approach is that these methods rely upon an upstream human detector, which can fail when multiple human instances populate an image, as in the case of crowds. To marginalize this issue, we also consider an oracular baseline that replaces the predicted bounding-box locations of humans in the scene with the ground truth ones. 

\subsection{From Crowd Counting to CVE}\label{sec:cc_volume}

To assess whether CVE can be naively solved without the adoption of any specific strategy, we set a baseline whose volume estimation stems from $C_{B+}(\imagei)\times\meanvold$, where $C_{B+}$\xspace is a \cc model\footnote{We use Bayesian+~\cite{ma2019bayesian}. In the Supplementary Material, we demonstrate that, for counting purposes, it performs best on \ourdataset.}, $I$ is the input image, and \meanvold is the average per-person volume in the dataset \datasetd. As a statistical reference, we further experiment with an oracular version of this baseline that replaces $C_{B+}(\imagei)$\xspace with the ground-truth count of image \imagei, namely \gtcounti$\times$\meanvold.

Additionally, we adapt relevant \textit{Localization}, \textit{Bayesian}, and \textit{Density Map} approaches from \cc (cf. Sec.~\ref{subsec:Counting}).
As for the \textit{Localization} approach, we select P2P-Net~\cite{song2021rethinking}.
For CVE purposes, we adjust its architecture to predict an array of (2+1) scalars, where 
the additional coordinate represents the volume of the target person.

As \textit{Bayesian} approaches we consider Bayesian+~\cite{ma2019bayesian} and MAN~\cite{lin2022boosting}.
We adapt them for CVE by appending an additional branch that takes the estimated density map as input and regresses the total volume in the input frame (cf. Sec.~6.2 and Fig.~6 in the Supplementary Material).

For the {\it Density Map} method, we adopt the recent STEERER~\cite{hani2023steerer}. Our adaptation preserves the original network architecture while modifying the model's supervision technique: instead of using conventional counting density maps that label a pixel representing a person's head with the value $1$, we use {\it \density}, where we annotate the pixel with the person's total volume. This Volume Density Map is then smoothed using a Gaussian filter.

\subsection{Per-part Volume Density Maps}\label{sec:pp_volume}

In our proposed approach, we leverage \ourdataset per-part volume annotations, discussed in Sec. \ref{sec:smpl_fitting}. Driven by the insight that volume is distributed throughout the human body, we enhance the proposed \density approach to incorporate this concept. Specifically, since \ourdataset provides fine-grained annotations of body parts volumes of each character, we introduce {\it \ppdensity}, where specific keypoints of each person are assigned a portion of the total body volume. For instance, each of the five torso keypoints will be attributed with \( \frac{1}{5} \) of the torso-only volume.  After smoothing this local map, the volume is distributed over the interested body parts (see the second column of  Table~\ref{tab:qual_results} for visualization). We train a STEERER-like model from scratch with these annotations and dub this model as STEERER-V.
\vspace{-0.3em}
\section{Experiments}\label{sec:experiments}

In this section, we evaluate all methods' performance on \ourdataset quantitatively and qualitatively (Secs.~\ref{subsec:results_baselines_cve} and ~\ref{sec:qualitatives}).
Sec.~\ref{sec:real_world} provides the results of our best model on real-world datasets.

\subsection{Experimental Results}\label{subsec:results_baselines_cve}

Table~\ref{tab:anthropos_results} reports results on the test set of \ourdataset for the CVE task. 
All the baselines are trained on \ourdataset. The HD+HMR baselines' human detectors are YOLOv7\cite{wang2023yolov7} instances fine-tuned on our dataset. \\
\noindent{\bf HD-HMR methods do not perform well.} The upper part of Table~\ref{tab:anthropos_results} shows that HD+HMR methods report suboptimal performance in CVE tasks. These approaches are limited by the heterogeneous scales of individuals within crowd scenes and the limitations of the HD, whose accuracy is significantly marred by severe occlusions and challenging environmental conditions. Note that HD not only fails to generate a bounding box for some individuals, but it can also propose multiple bounding boxes for the same person, resulting in redundant volume estimations for the same individual. Replacing the HD with an oracle that provides GT bounding boxes 
yields a marked reduction in volume estimation error, still reporting a rather large PPMAE with respect to the Crowd Counting adapted baselines. This is probably due to the elevated number of occlusions in \ourdataset, which hampers the exact body shape reconstructions (cf. Sec.~6.1 of Sup. Mat.). Our proposed STEERER-V demonstrates superior performance, 
surpassing all the oracle-enhanced HMR approaches.
\\
\noindent{\bf Density maps help in CVE.} 
STEERER demonstrates superior performance among the methodologies adapted from Crowd Counting and trained on Volume Density Maps (second block in Table~\ref{tab:anthropos_results}). This result suggests that Bayesian~\cite{ma2019bayesian, lin2022boosting} and localization~\cite{song2021rethinking} techniques exhibit suboptimal efficacy in CVE compared to architectures purely based on density prediction. Indeed, STEERER-V, a model built up from STEERER that benefits from the proposed \ppdensity during training (cf. Sec.~\ref{sec:pp_volume}), reports the best performance. This superiority is attributed to its capacity for fine-grained predictions, enabling effective management of significant occlusions inherent in crowded scenarios. Quantitatively, STEERER-V reports a minimal average error of 6.73 dm$^3$ per individual, representing a 53.36\% improvement over the most effective Crowd Counting adapted model, STEERER. \newline
\begin{table}[!t]
    \centering 
    \resizebox{1\linewidth}{!}{
    \begin{tabular}{ll|c|cc|c}
        \toprule
        & \textbf{Model} & \textbf{Oracle} & \textbf{MAE} & \textbf{PPMAE} & \textbf{Inf. time} \\
        \midrule
        \multirow{6}{*}{\rotatebox[origin=t]{90}{\parbox[t]{2.5cm}{\centering \textbf{HMR}}}} & CLIFF~\cite{li2022cliff} & & 673.7 & 21.41 & 145.7 \\
        & BEDLAM-CLIFF~\cite{black2023bedlam} & & 656.4 & 21.17 & 137.9 \\
        & ReFit~\cite{wang2023refit} & & 595.2 & 18.79 & 170.0 \\
        & \textcolor{gray}{CLIFF}~\cite{li2022cliff} & \textcolor{gray}{\textbf{\checkmark}} & \textcolor{gray}{370.2} & \textcolor{gray}{12.89} & \textcolor{gray}{56.68} \\
        & \textcolor{gray}{BEDLAM-CLIFF}~\cite{black2023bedlam} & \textcolor{gray}{\textbf{\checkmark}} & \textcolor{gray}{364.7} & \textcolor{gray}{12.15} & \textcolor{gray}{49.73} \\
        & \textcolor{gray}{ReFit}~\cite{wang2023refit} & \textcolor{gray}{\textbf{\checkmark}} & \textcolor{gray}{346.8} & \textcolor{gray}{11.31} & \textcolor{gray}{108.1} \\
        \midrule
        \multirow{6}{*}{\rotatebox[origin=t]{90}{\parbox[t]{2cm}{\centering \textbf{Counting}}}}
        & Bayesian+~\cite{ma2019bayesian} & & 578.09 & 17.31 & \bf 37.50\\
        & P2P~\cite{song2021rethinking} & & 590.91 & 17.07 & 61.16\\
        & MAN~\cite{lin2022boosting} & & 557.90 & 17.03 & 81.84\\
        & STEERER~\cite{hani2023steerer} & & 506.94 & 14.43 & 105.1\\
        & $C_{B+}(\imagei)\times\meanvold$ & & 507.97 & 14.39 & \bf 37.50\\
        & \textcolor{gray}{\gtcounti$\times$\meanvold} & \textcolor{gray}{\textbf{\checkmark}} & \textcolor{gray}{191.50} & \textcolor{gray}{5.32} & \textcolor{gray}{-} \\ 
        \midrule
        & STEERER-V~\cite{hani2023steerer} & & \textbf{205.59} & \textbf{6.73} & 105.1 \\ 
        \bottomrule
        \end{tabular}}
    \caption{Results on \ourdataset, reported in dm${^3}$. Inference time (ms) is measured on an NVIDIA A100. Methods are divided into HD+HMR, Crowd Counting, and our proposed approach. $C_{B+}(\imagei)$ refers to the headcount given by~\cite{ma2019bayesian}, while \meanvold is the average human volume. Grayed-out lines use oracular information and should not be directly compared with the other results.}\label{tab:anthropos_results}
    \vspace{-1.5em}
\end{table}
\noindent \textbf{Crowd Counting is not enough for CVE}. 
Additionally, Table~\ref{tab:anthropos_results} presents the results of the $C_{B+}(\imagei)\times\meanvold$ approach (Sec. \ref{sec:cc_volume}).
This method underperforms when compared to STEERER-V and its oracular counterpart. This is due to the compounded error arising from substituting individual body volume estimations with the average volume, \meanvold, as well as the inherent detection inaccuracies of the counting model, $C_{B+}$. This comparison underscores the significant performance degradation that would result \emph{in practice} from naively applying a Crowd Counting strategy to CVE, highlighting the necessity for a specialized approach designed specifically for VE. \\
When considering \gtcounti$\times$\meanvold, which mimics a perfect human detector, something unattainable in practical applications, the error is reduced. This comparison emphasizes the magnitude of the error introduced by the imperfect detection carried out by $C_{B+}(\imagei)\times\meanvold$.   \\
It is worth noting that although STEERER-V does not leverage any privileged information and consequently exhibits imperfect detection, it is comparable with \gtcounti$\times$\meanvold.
This indicates that STEERER-V compensates for its detection inaccuracies achieving a robust per-person volume estimation, making it the best candidate for practical CVE application. This quality primarily originates from STEERER-V's training strategy, which integrates body part detection, mitigating false negatives caused by occlusions, with expert knowledge of the volume contribution of each body segment. An additional experiment where we separately assess the contributions of the volume estimation and detection errors to the total error is available in Supplementary Material's Sec.~11.\\

\subsection{Qualitative Evaluation}\label{sec:qualitatives}

Table~\ref{tab:qual_results} presents the qualitative results of our proposed method, STEERER-V, alongside the Per-Part Volume Density Map, which serves as its training supervision. Furthermore, we provide qualitative results of STEERER and BEDLAM-CLIFF.
STEERER-V stands out as the top performer because of its robustness to occlusion and its capacity to generalize. The first row demonstrates that both STEERER and STEERER-V perform well when heads are visible and occlusions are minimal. However, STEERER tends to hallucinate volume along the branches of trees, probably because the model learned that such an object may hide human heads. Contrarily, STEERER-V, designed to distribute volume across the entire body, does not suffer from this side effect, as it does not detect bodies in such scenarios.
As occlusions intensify, particularly with multiple people overlapping at a distance (second and third rows), the performance gap between STEERER and STEERER-V becomes more pronounced, with STEERER-V being notably better. In the case of the dark image in the fourth row, STEERER fails to recognize the volume of the person in the foreground because their head merges with the background, while STEERER-V focuses on visible body parts, such as arms or legs, thus reducing the error. Additional qualitative results are available in the Supplementary Materials.

\begin{table}[t]
\centering 
\label{tab:performance_comparison}
\resizebox{\linewidth}{!}{
\begin{tabular}{l|c|c|c|c|c|c}
    \toprule
    & $C_{B+}(\imagei)\times\meanvold$ & \textbf{Refit} & \textbf{B-CLIFF} & \textbf{CLIFF} &\textbf{STEERER} &  \textbf{STEERER-V} \\
    \midrule
    \textbf{3DPW} & 71.3/43 & 125/75 & 64.0/40 & 69.8/43 & 102/89 &  40.4/25\\
    \textbf{CH} & - & -30.0 & -10.1 & +1.00 & -7.30 &  -3.40 \\
    \bottomrule
\end{tabular}
}
    \caption{Evaluation on 3DPW and CrowdHumans (CH). Reported metrics are MAE/PP-MAE (3DPW) and the difference between the average real-world per-person volume and the predicted per-person one (CH). B-CLIFF stands for BEDLAM-CLIFF.}
    \label{table:real-world}
\vspace{-1.3em}
\end{table}

\begin{table*}[t!]
  \centering
  \setlength{\tabcolsep}{1pt} 
  \renewcommand{\arraystretch}{0.4} 
  \begin{tabular}{ccccc}
    {\footnotesize ANTHROPOS-V} &
    {\footnotesize GT Per-Part} &
    {\footnotesize STEERER } &
    {\footnotesize STEERER-V } &
    {\footnotesize BEDLAM-CLIFF} \\
    
      \includegraphics[width=0.2\linewidth]{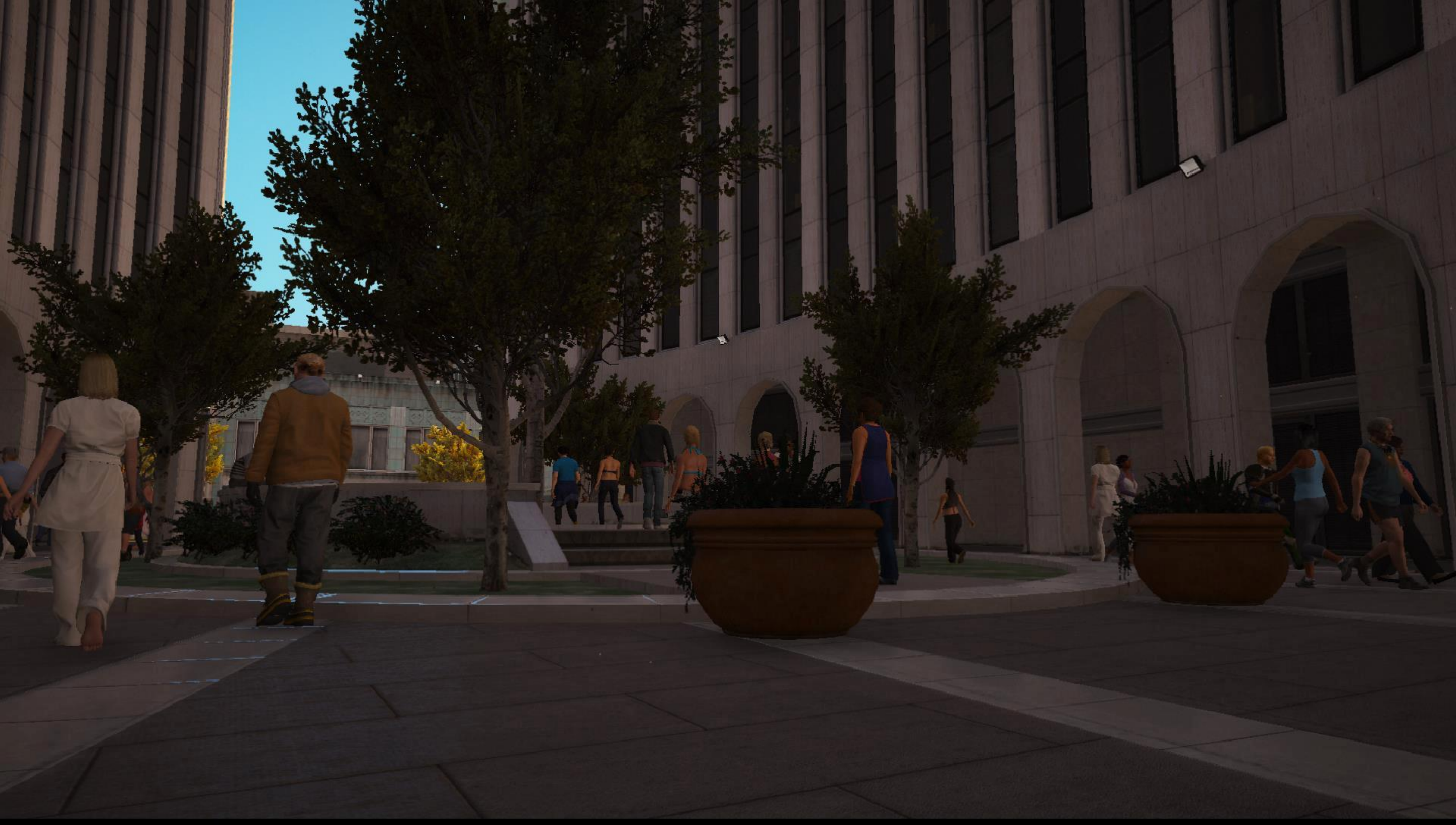} &
      
      \includegraphics[width=0.2\linewidth] {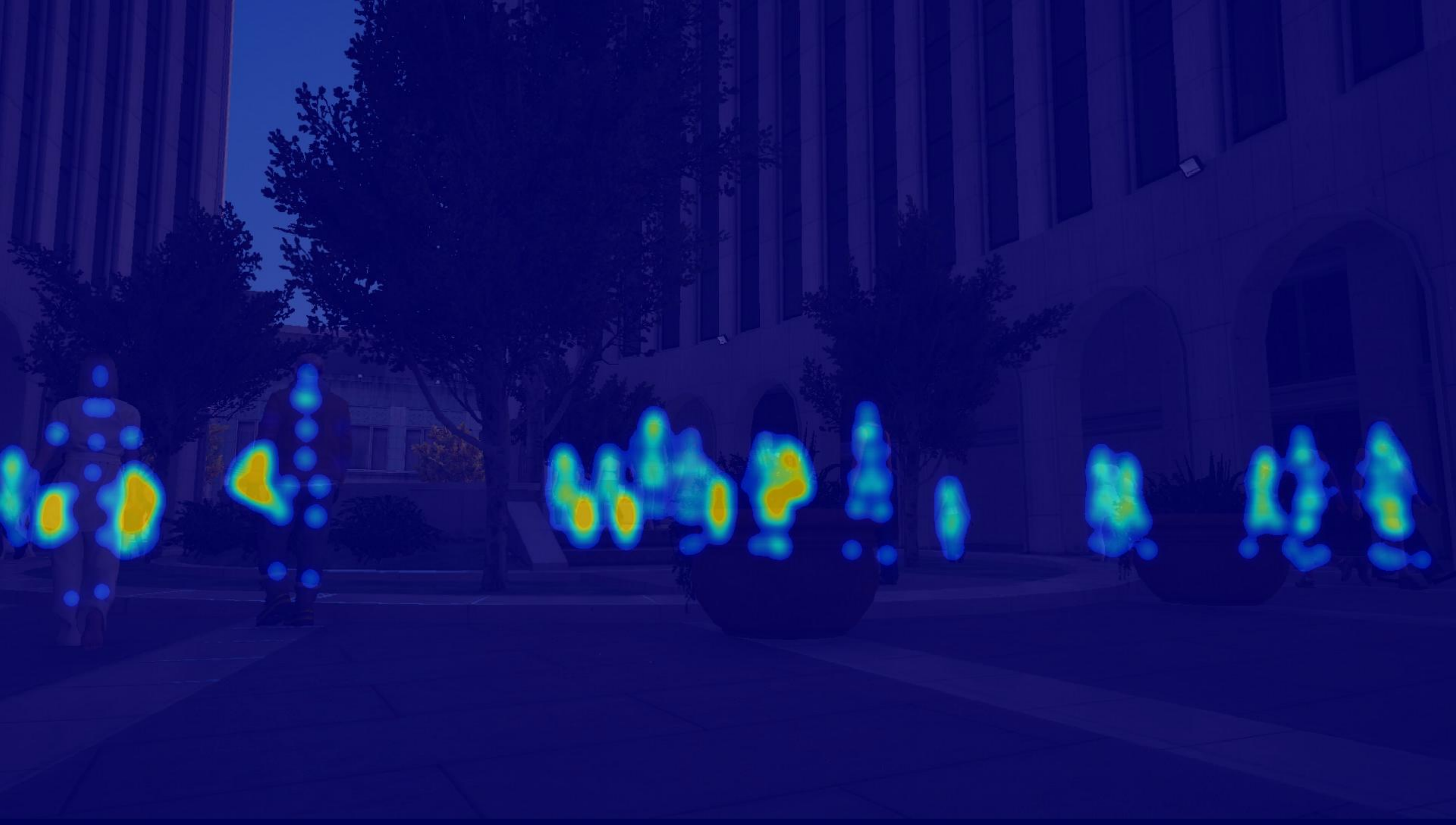} &

      \includegraphics[width=0.2\linewidth]{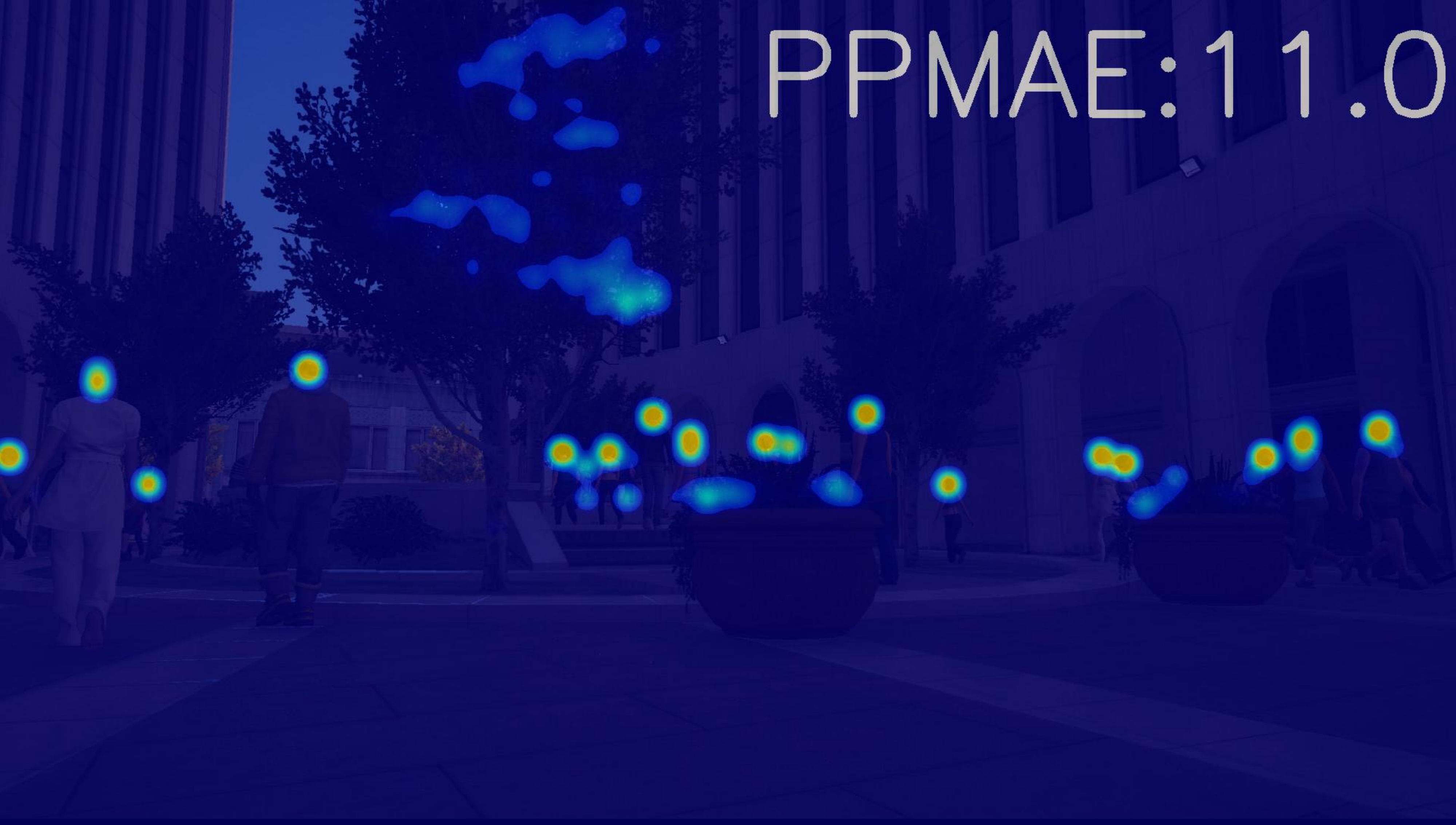} &

      \includegraphics[width=0.2\linewidth]{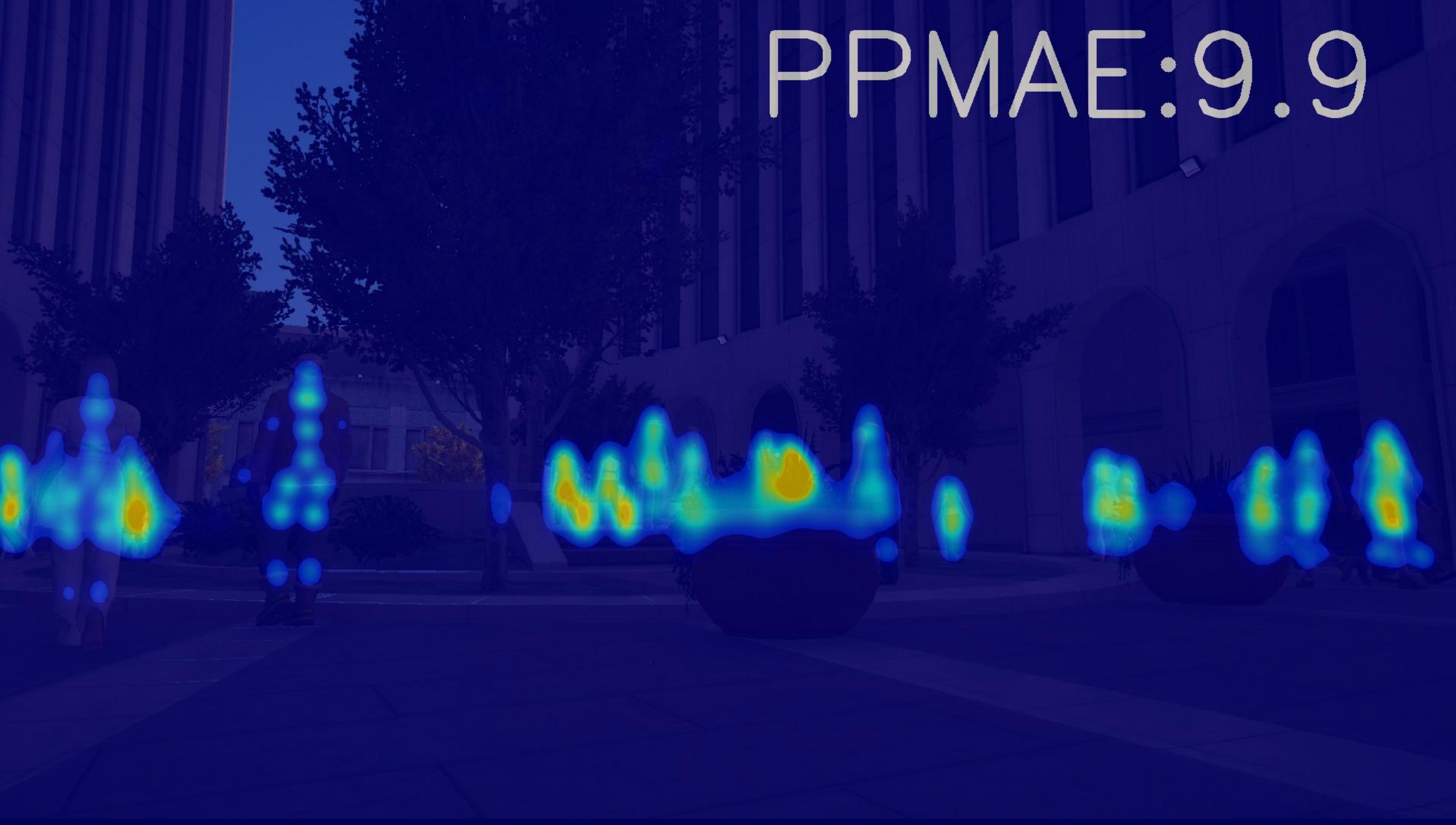} &
    
      \includegraphics[width=0.2\linewidth]{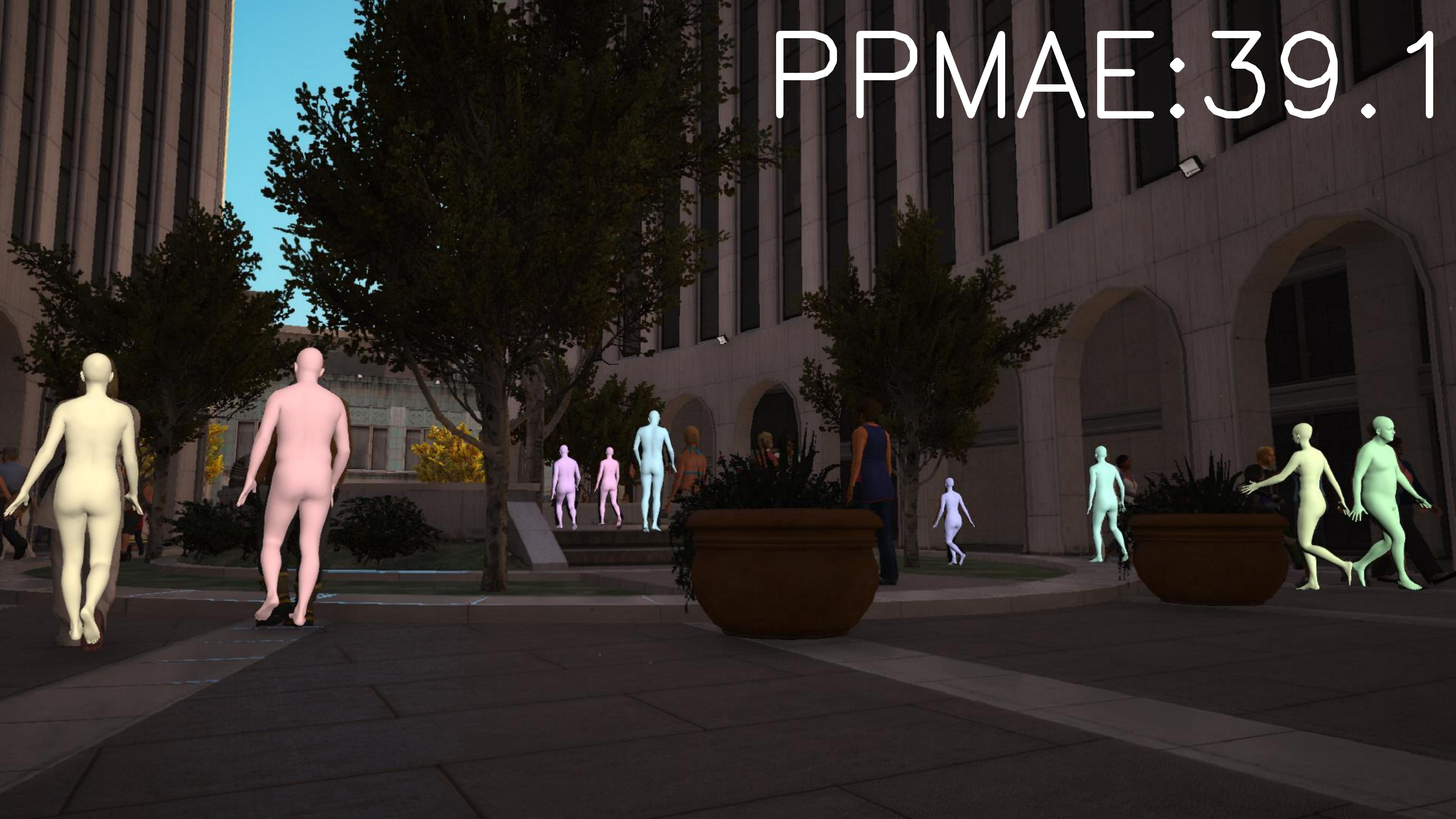} \\

      \includegraphics[width=0.2\linewidth]{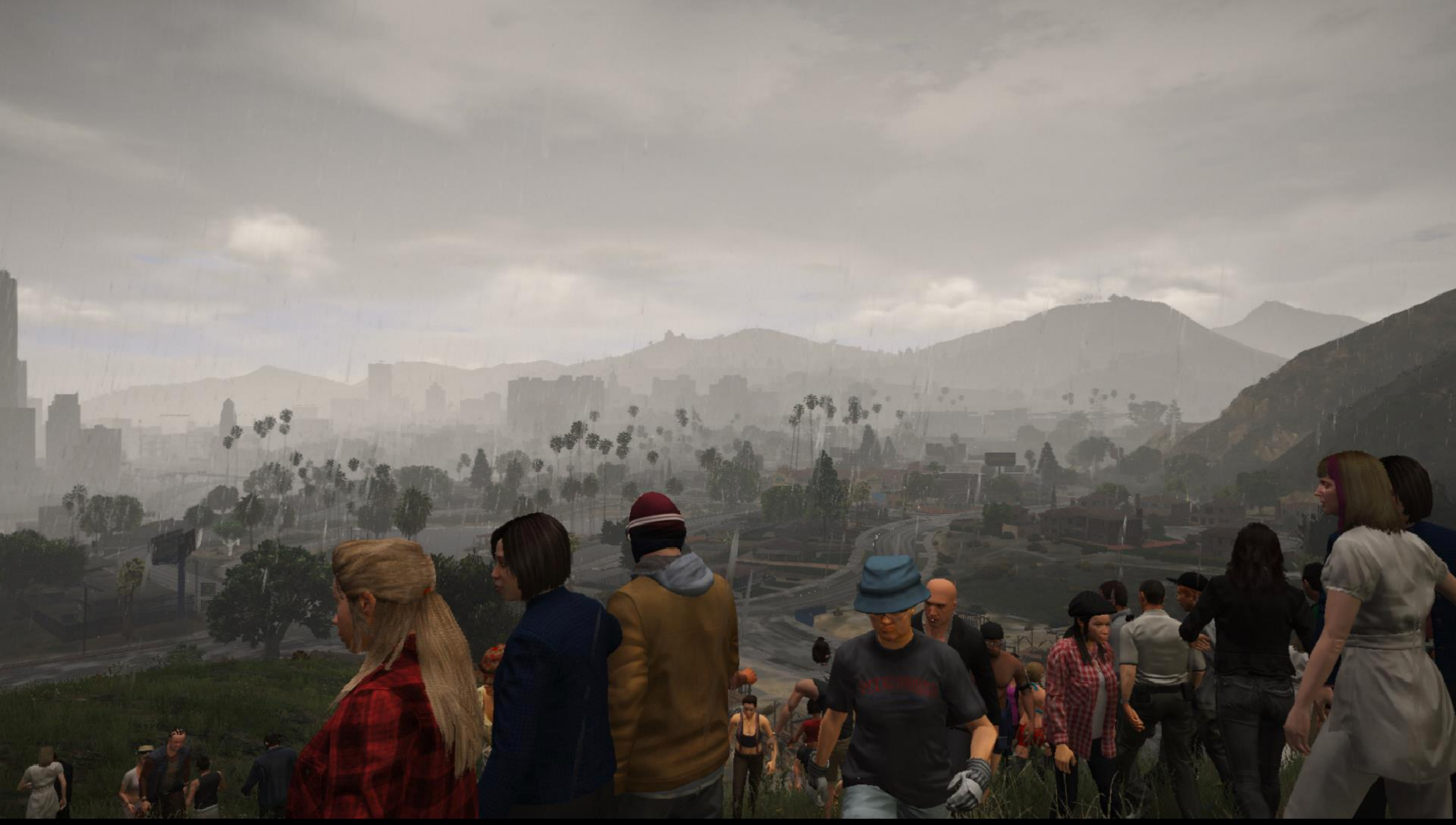} &
      
      \includegraphics[width=0.2\linewidth] {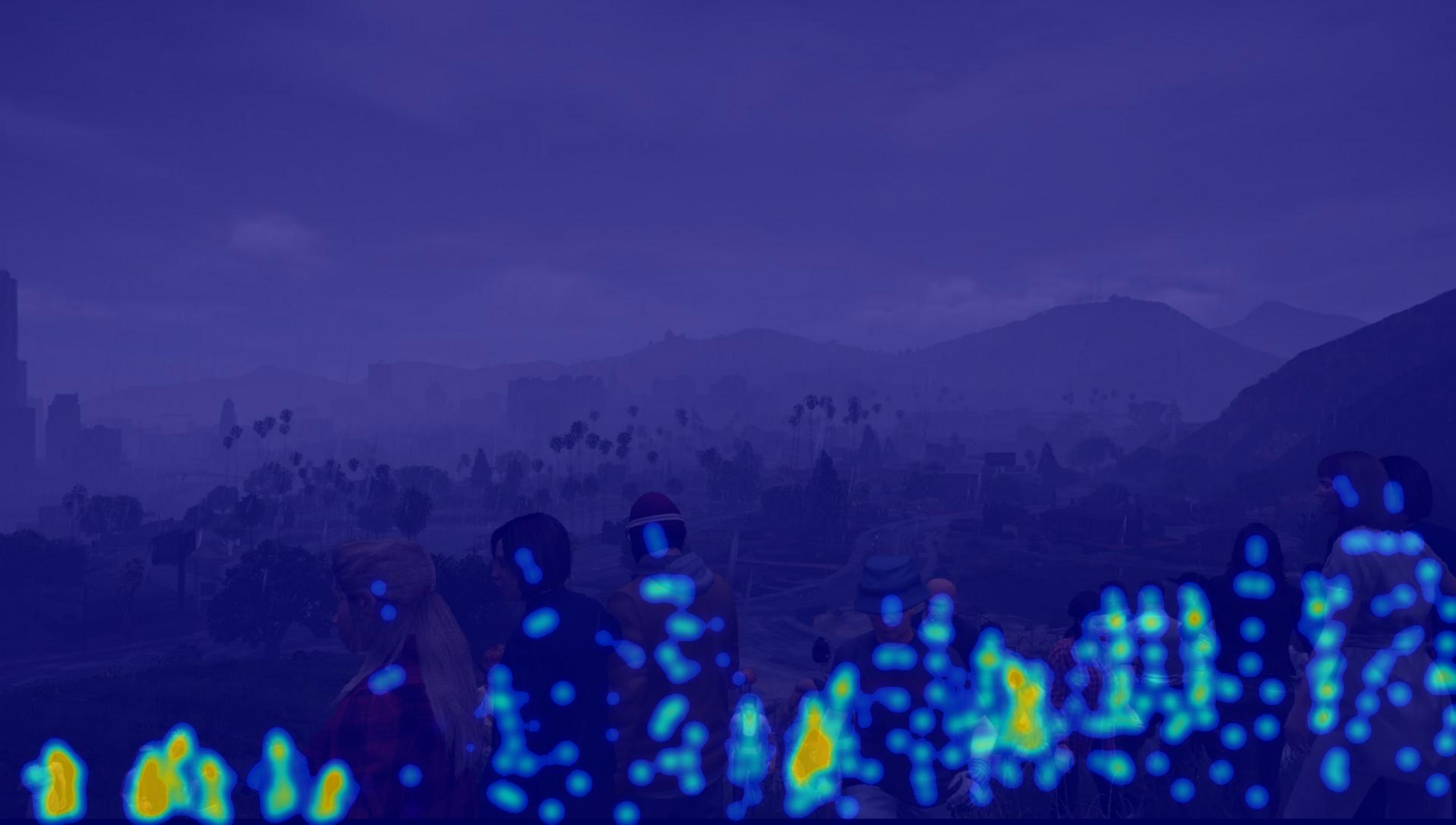} &

      \includegraphics[width=0.2\linewidth]{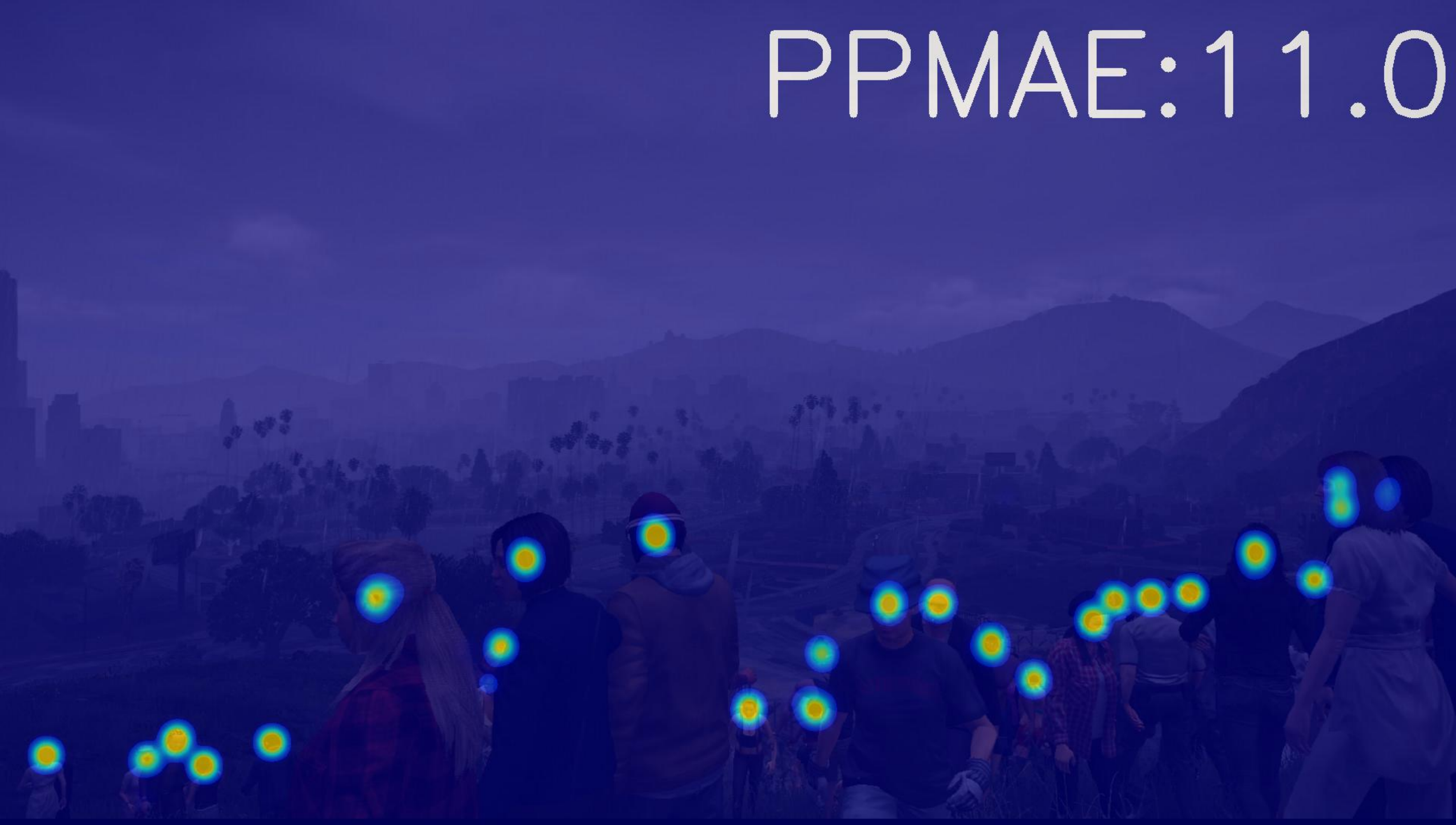} &

      \includegraphics[width=0.2\linewidth]{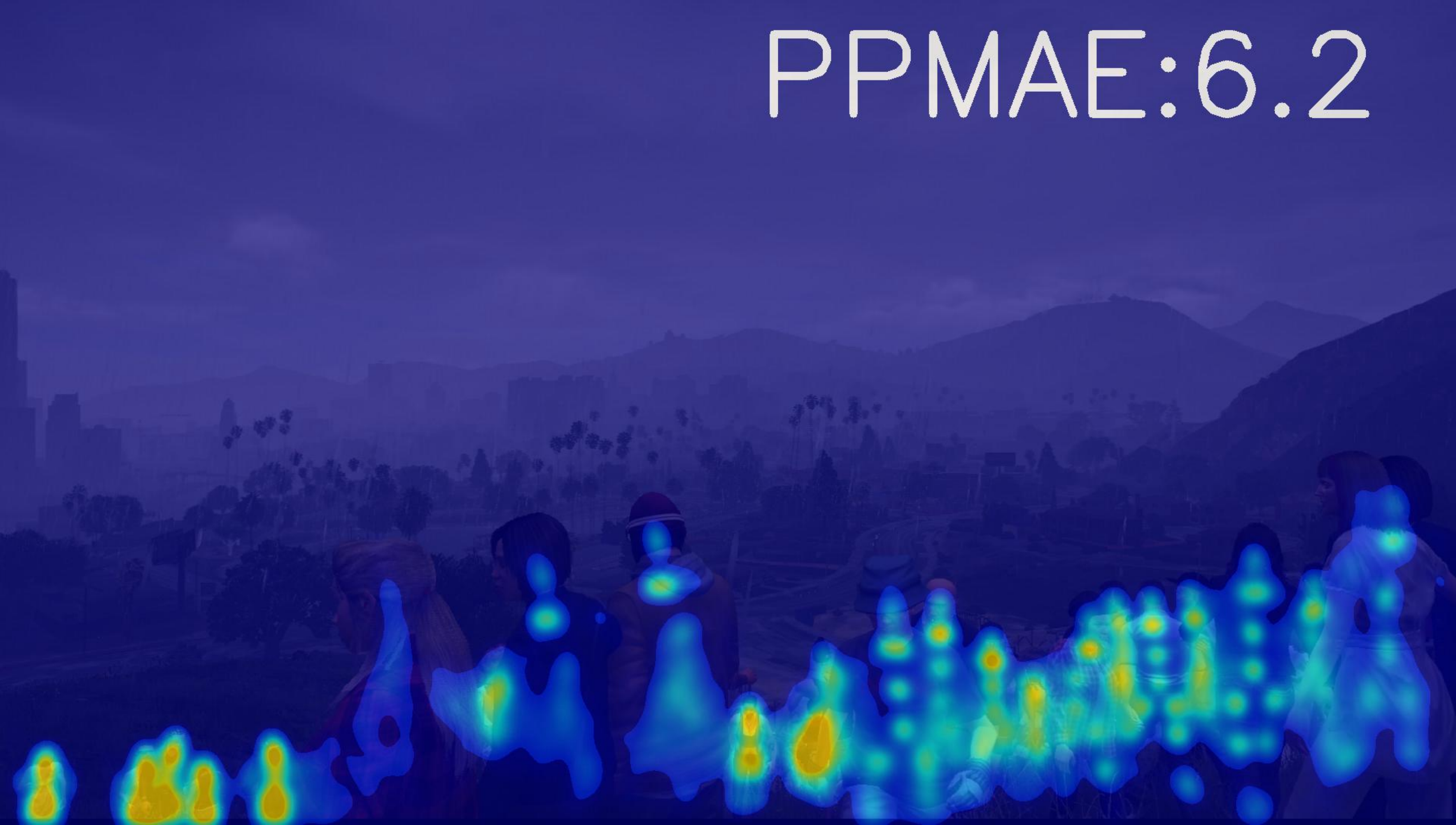} &
    
      \includegraphics[width=0.2\linewidth]{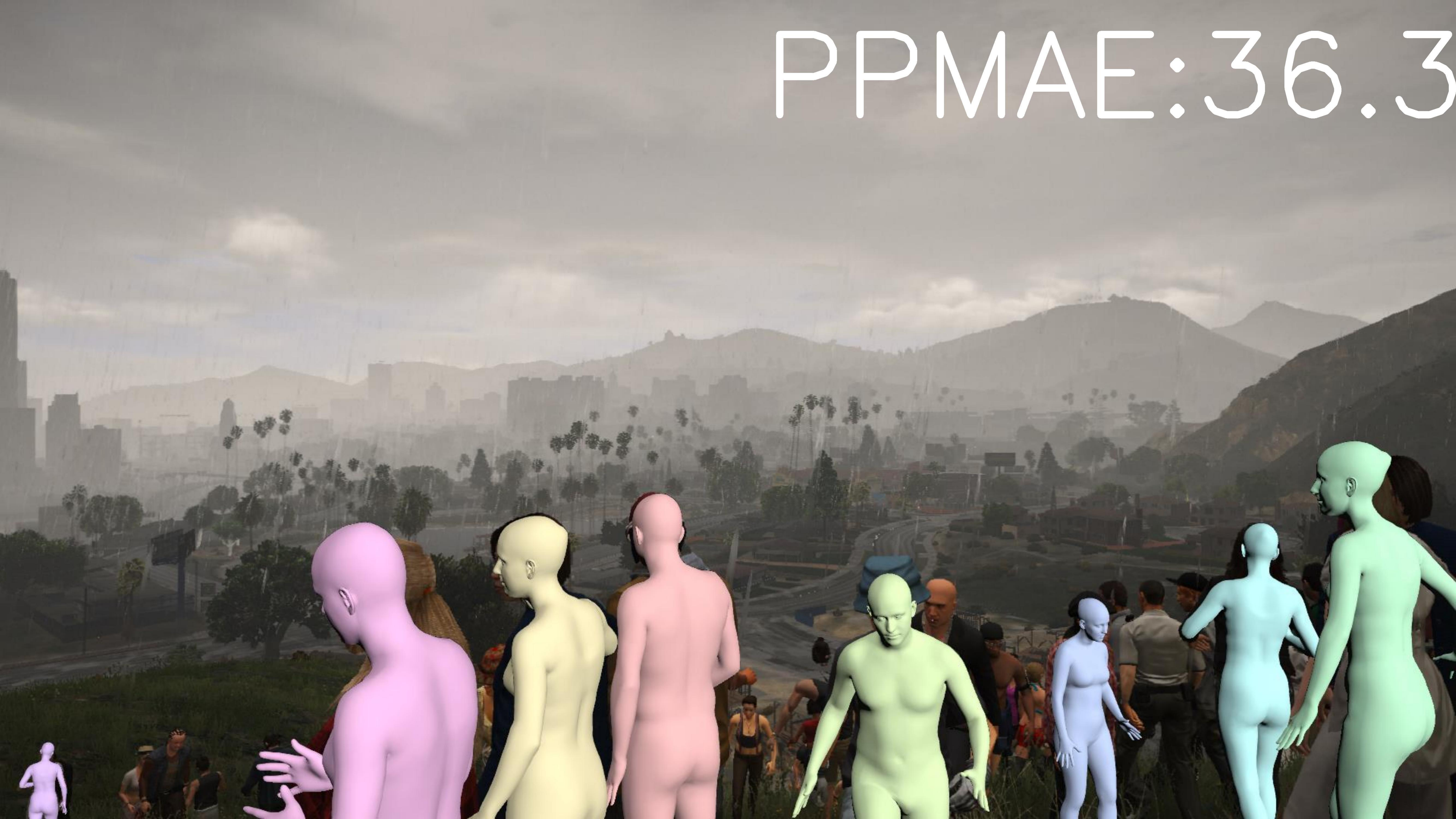} \\

            
      \includegraphics[width=0.2\linewidth]{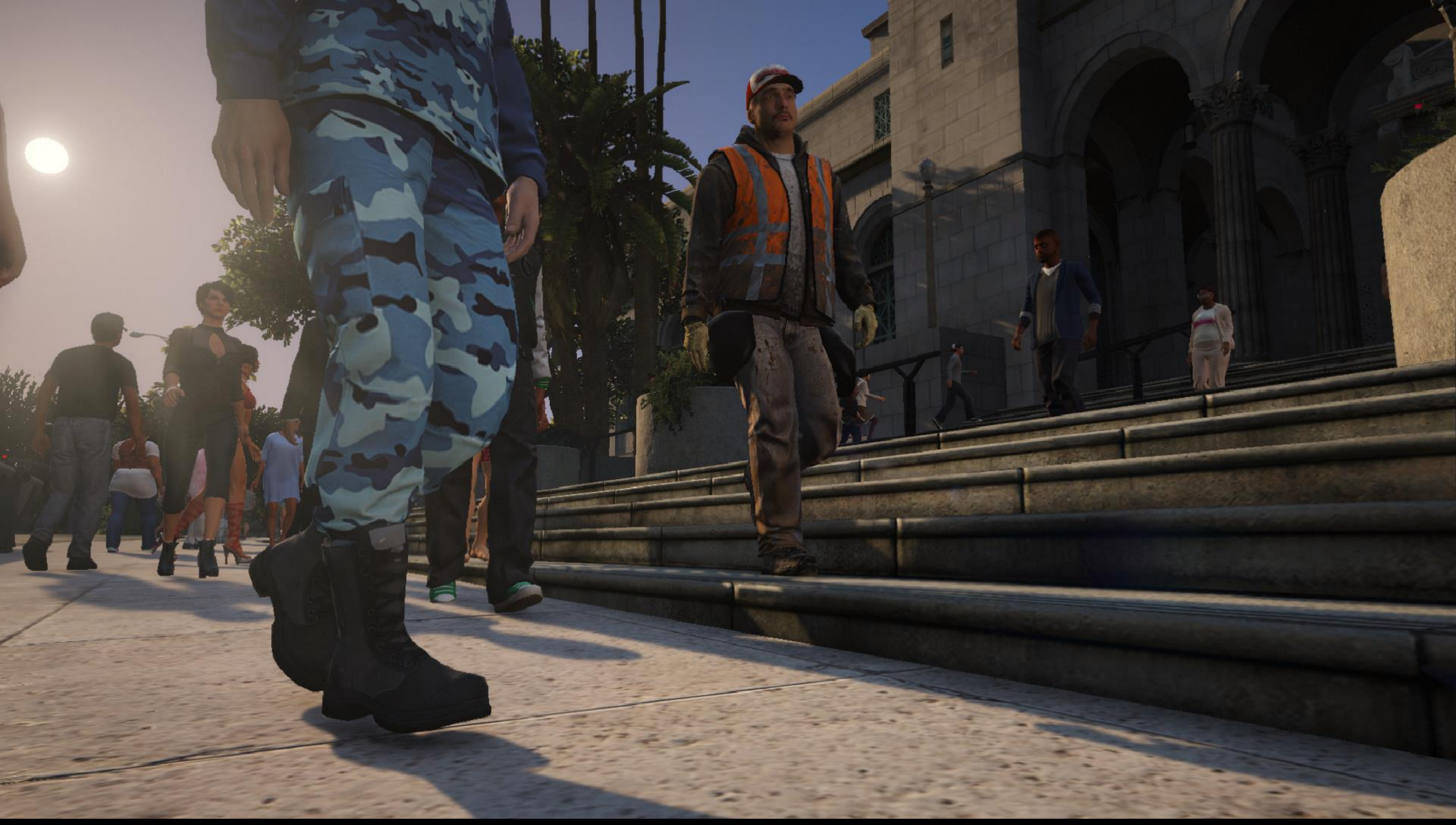} &
      
      \includegraphics[width=0.2\linewidth] {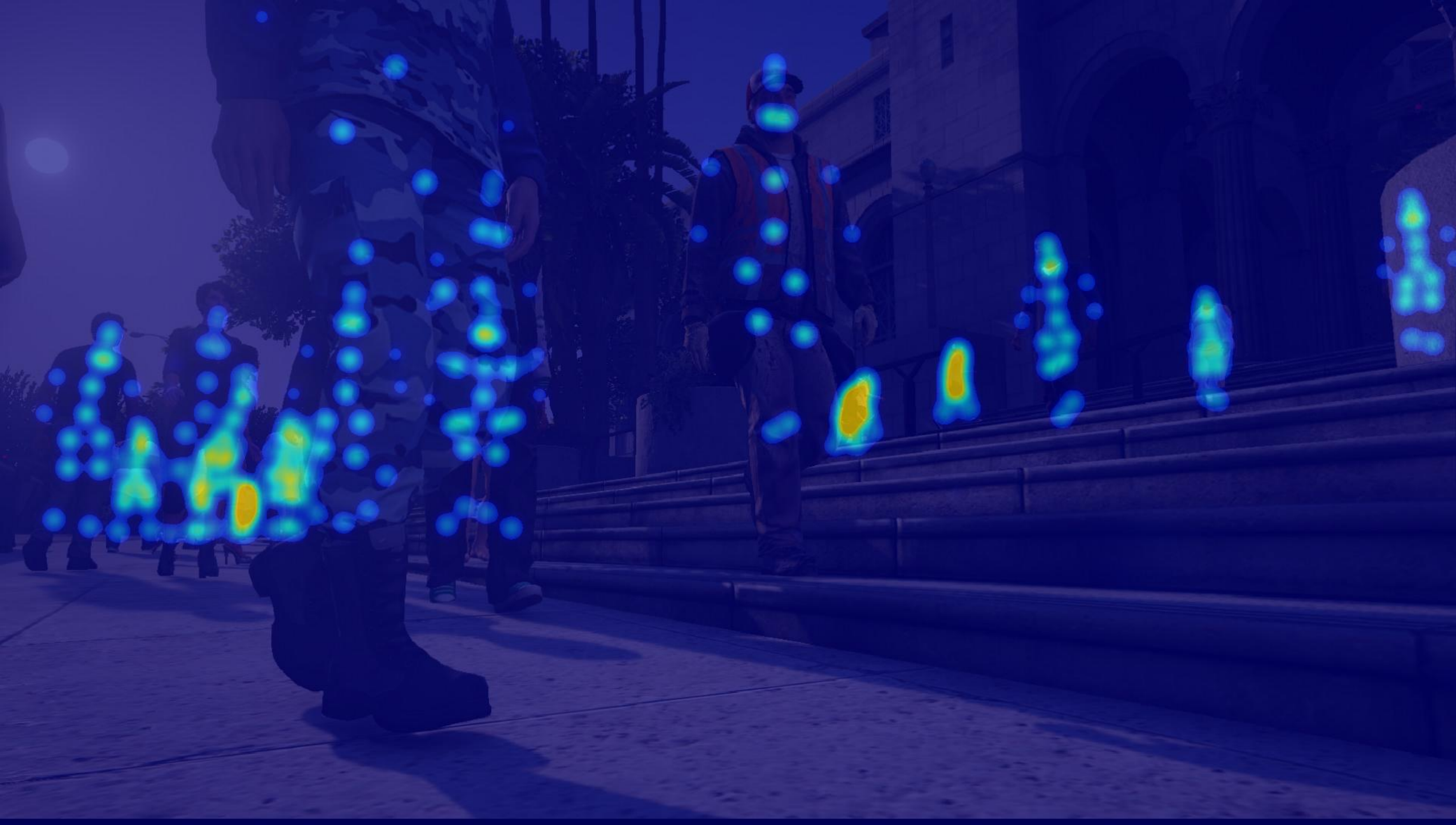} &

      \includegraphics[width=0.2\linewidth]{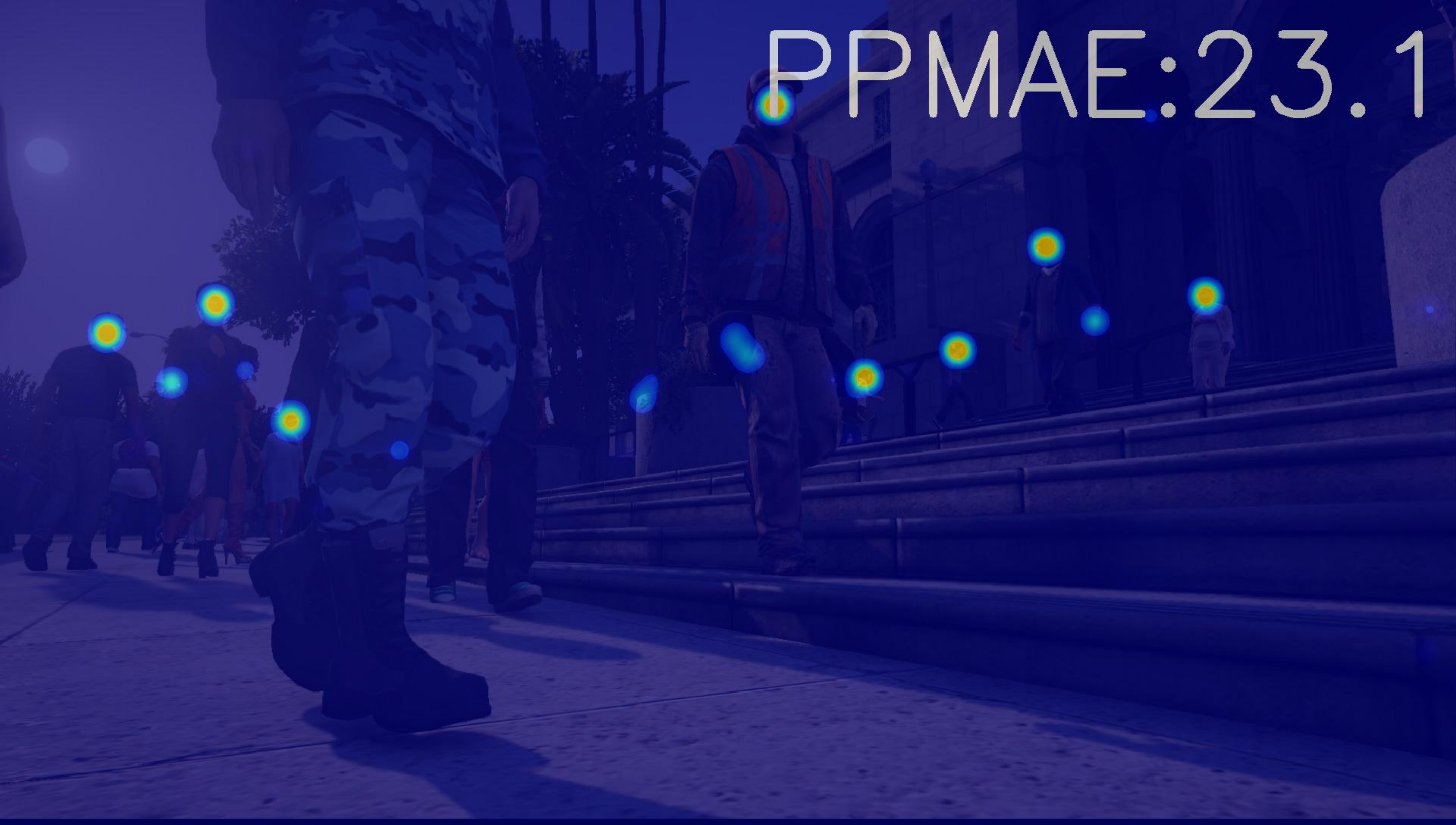} &

      \includegraphics[width=0.2\linewidth]{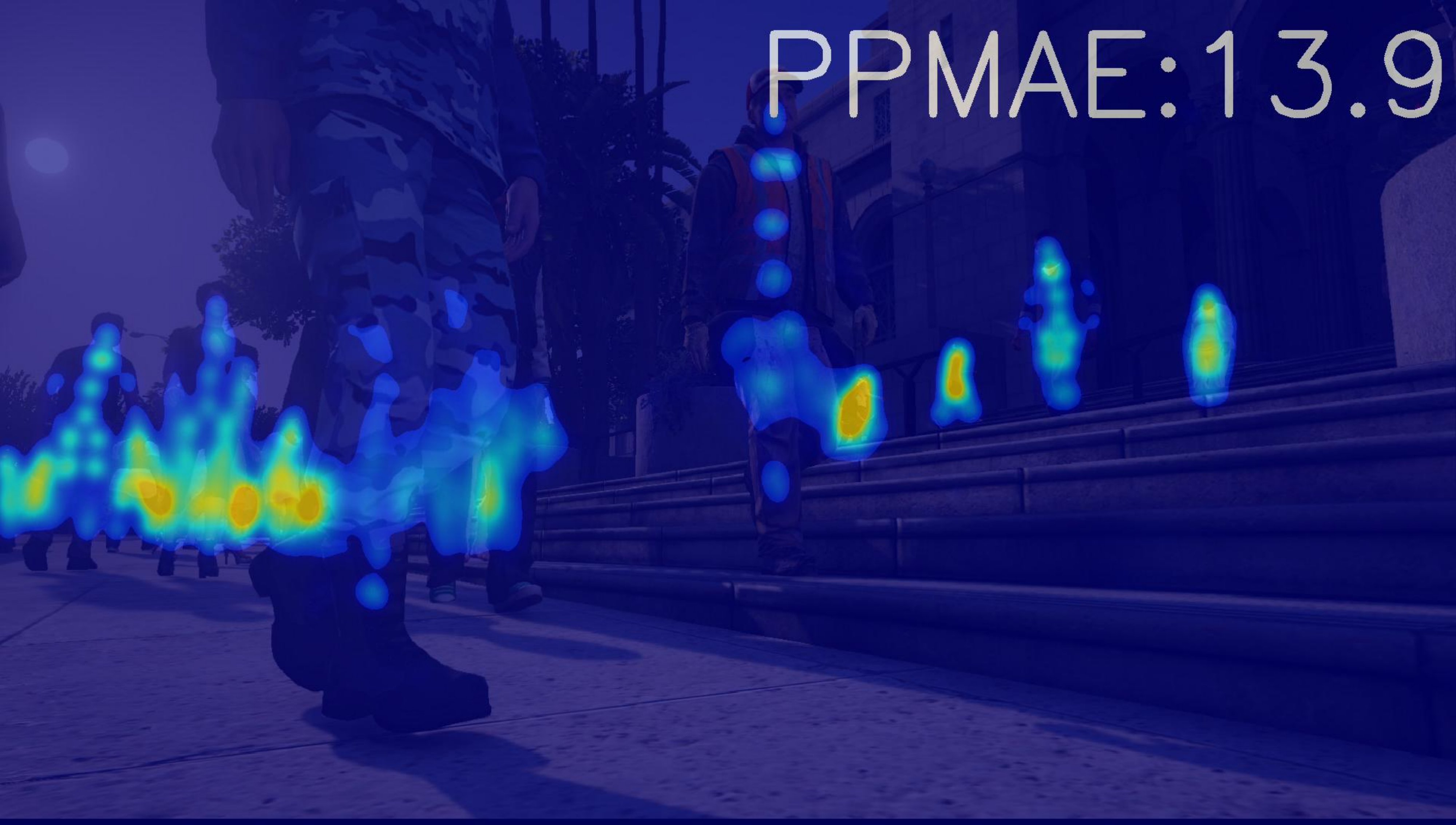} &
    
      \includegraphics[width=0.2\linewidth]{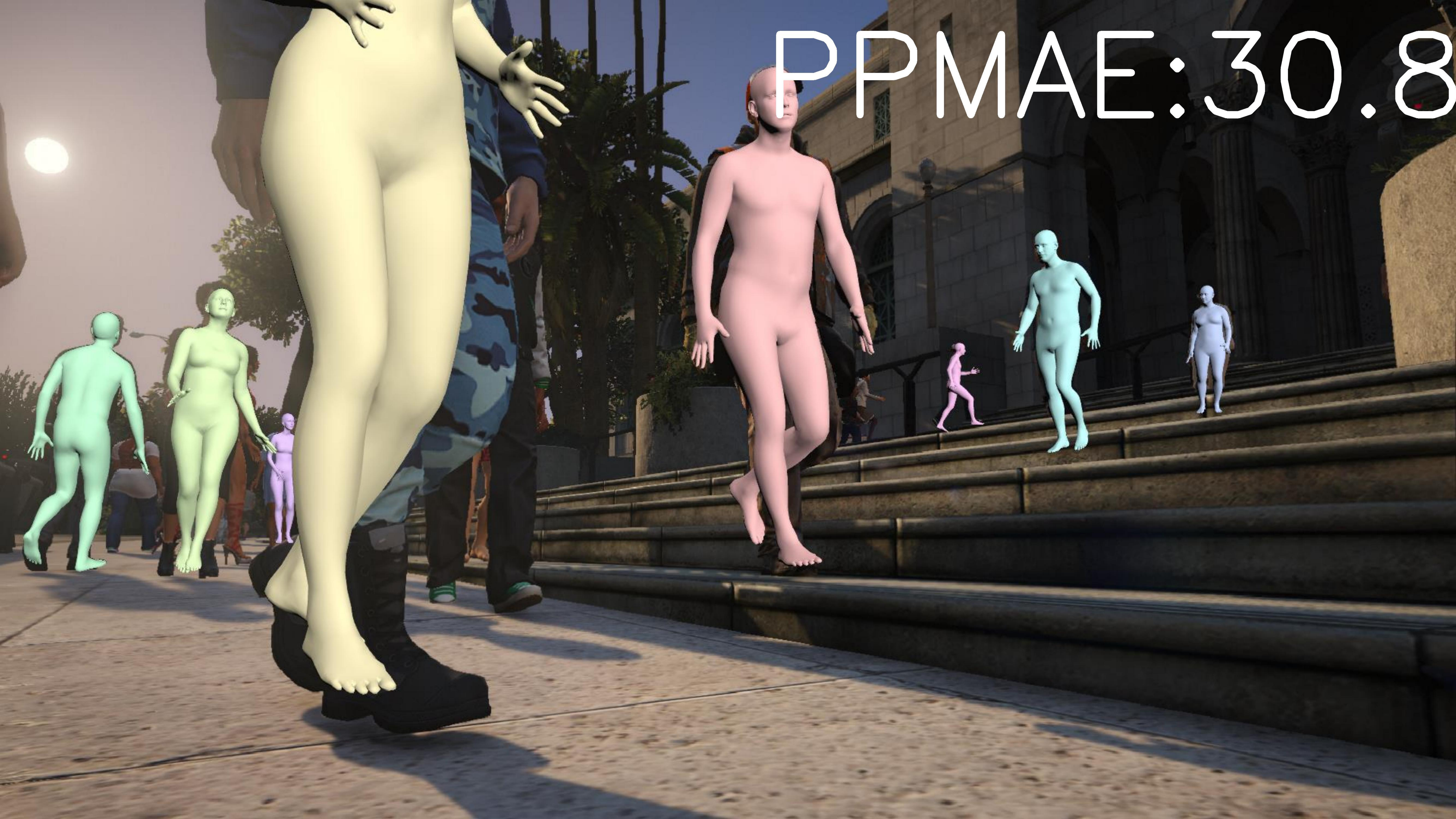} \\

      \includegraphics[width=0.2\linewidth]{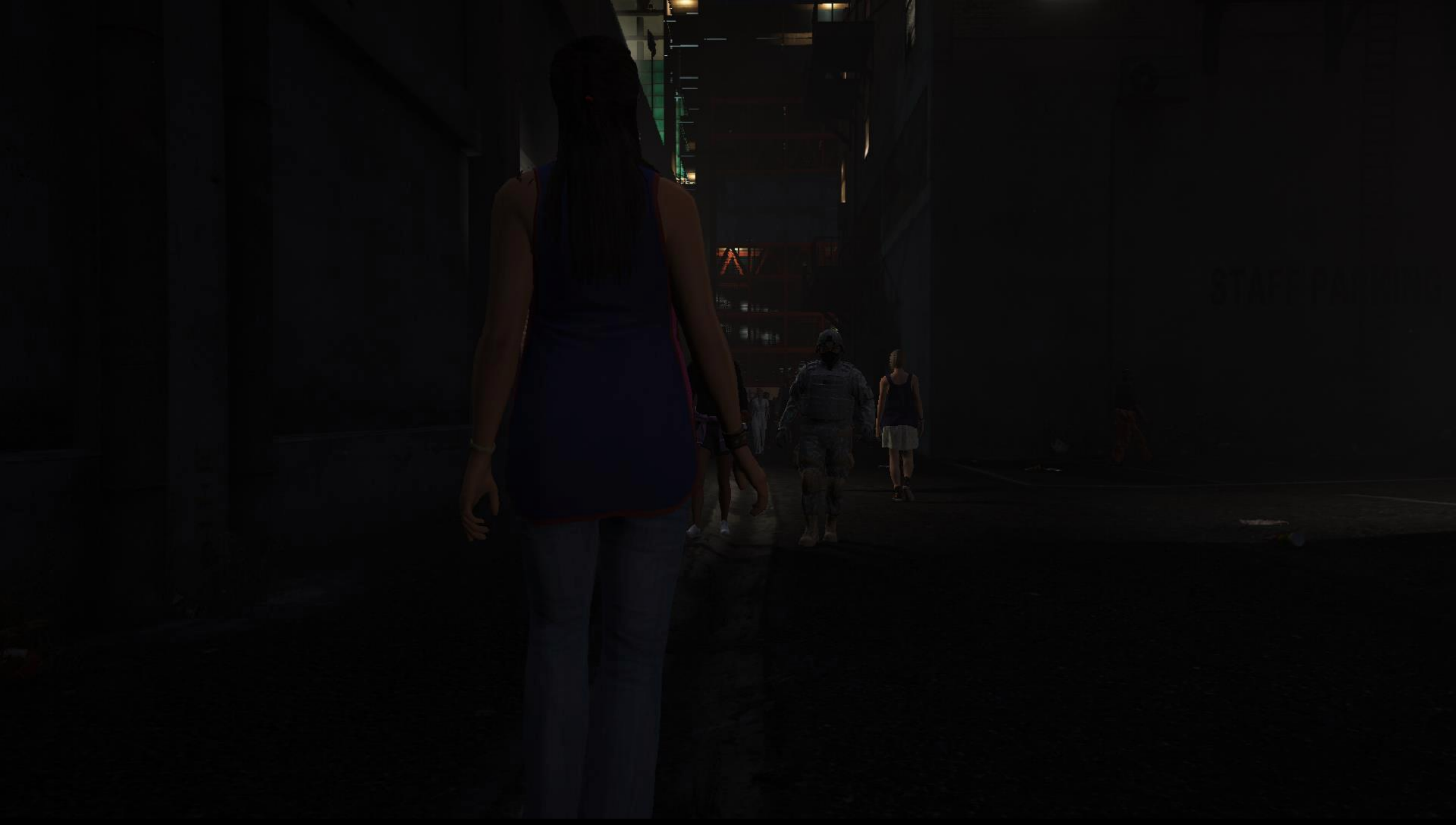} &
      
      \includegraphics[width=0.2\linewidth] {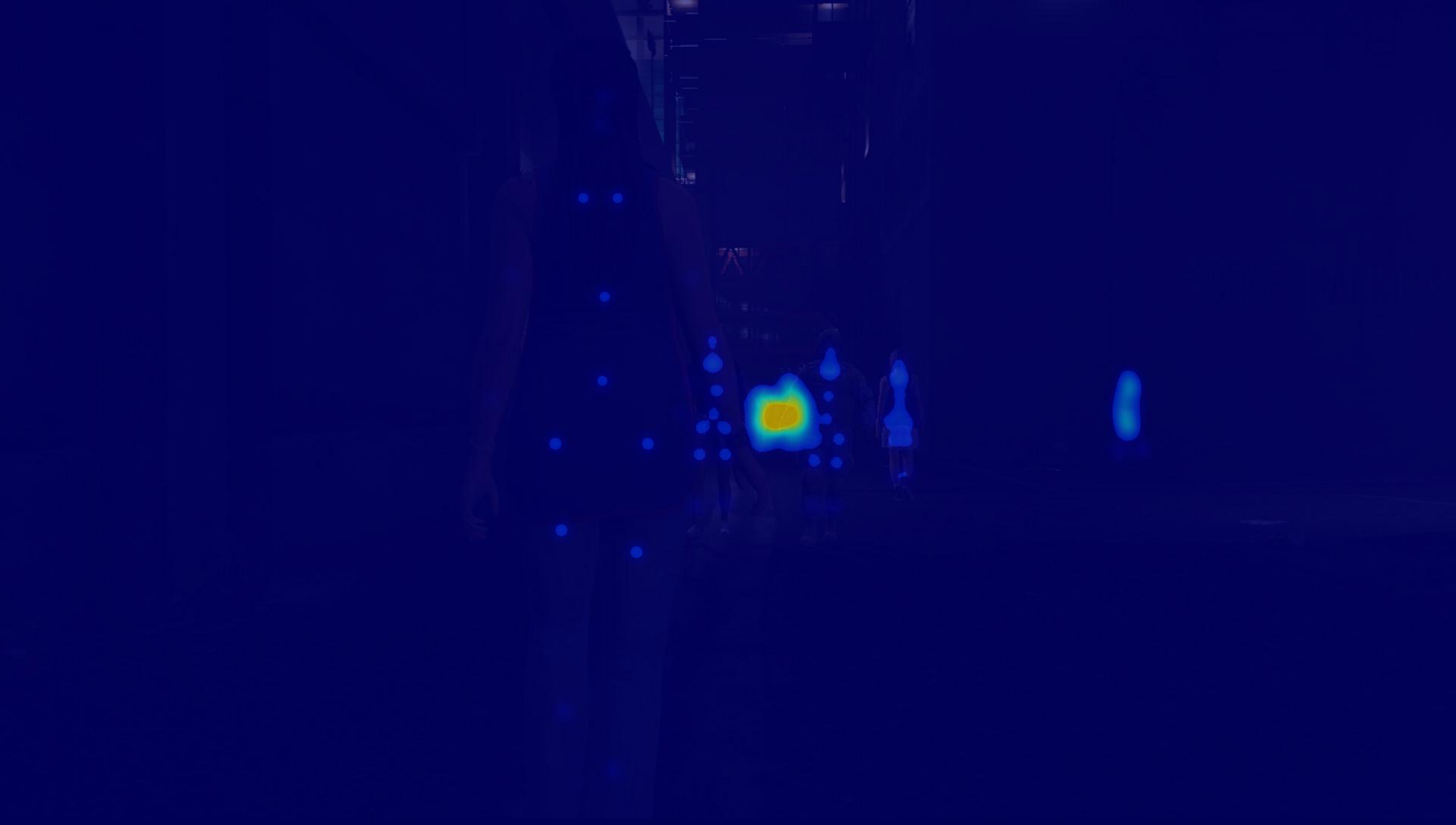} &

      \includegraphics[width=0.2\linewidth]{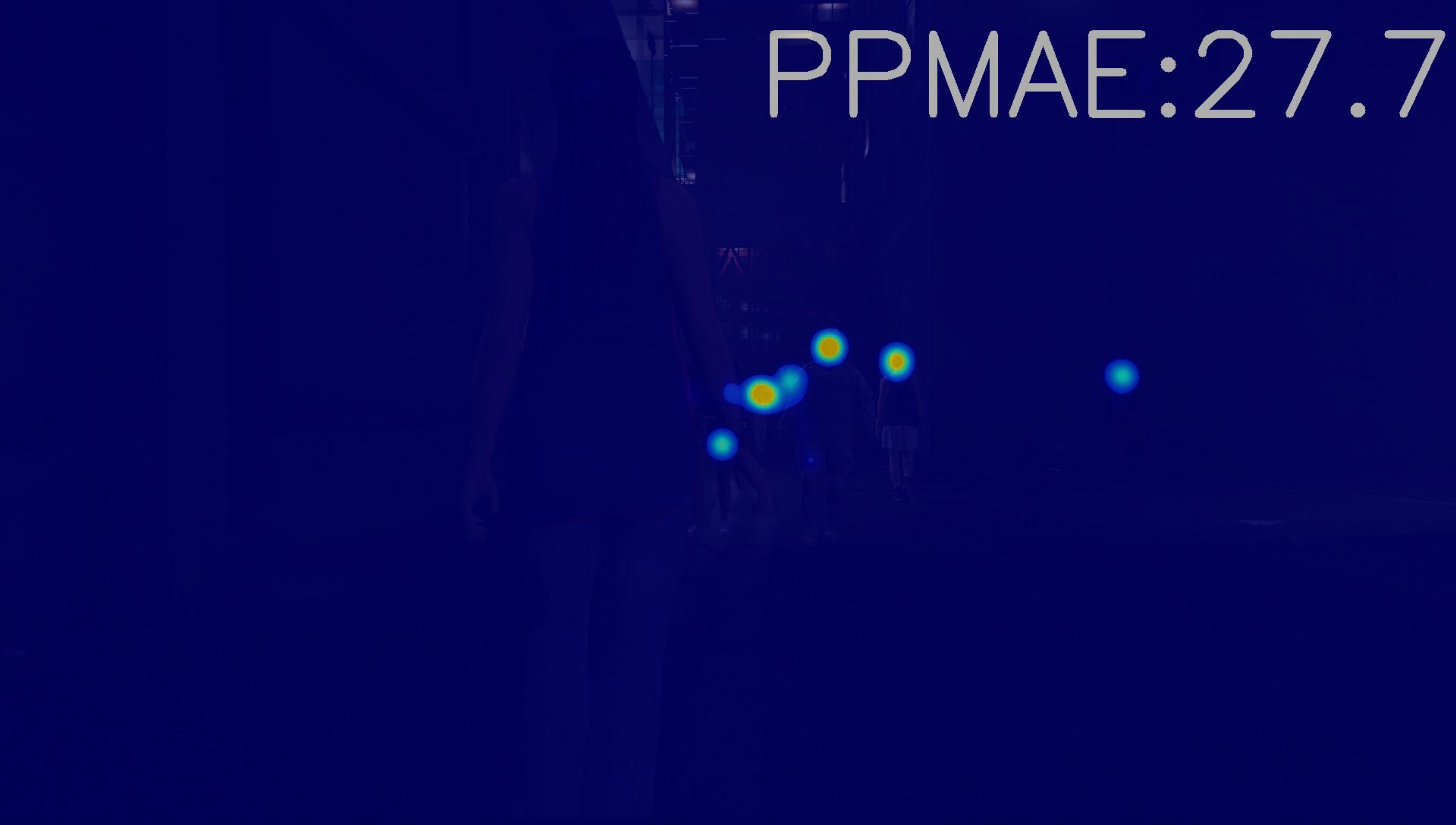} &

      \includegraphics[width=0.2\linewidth]{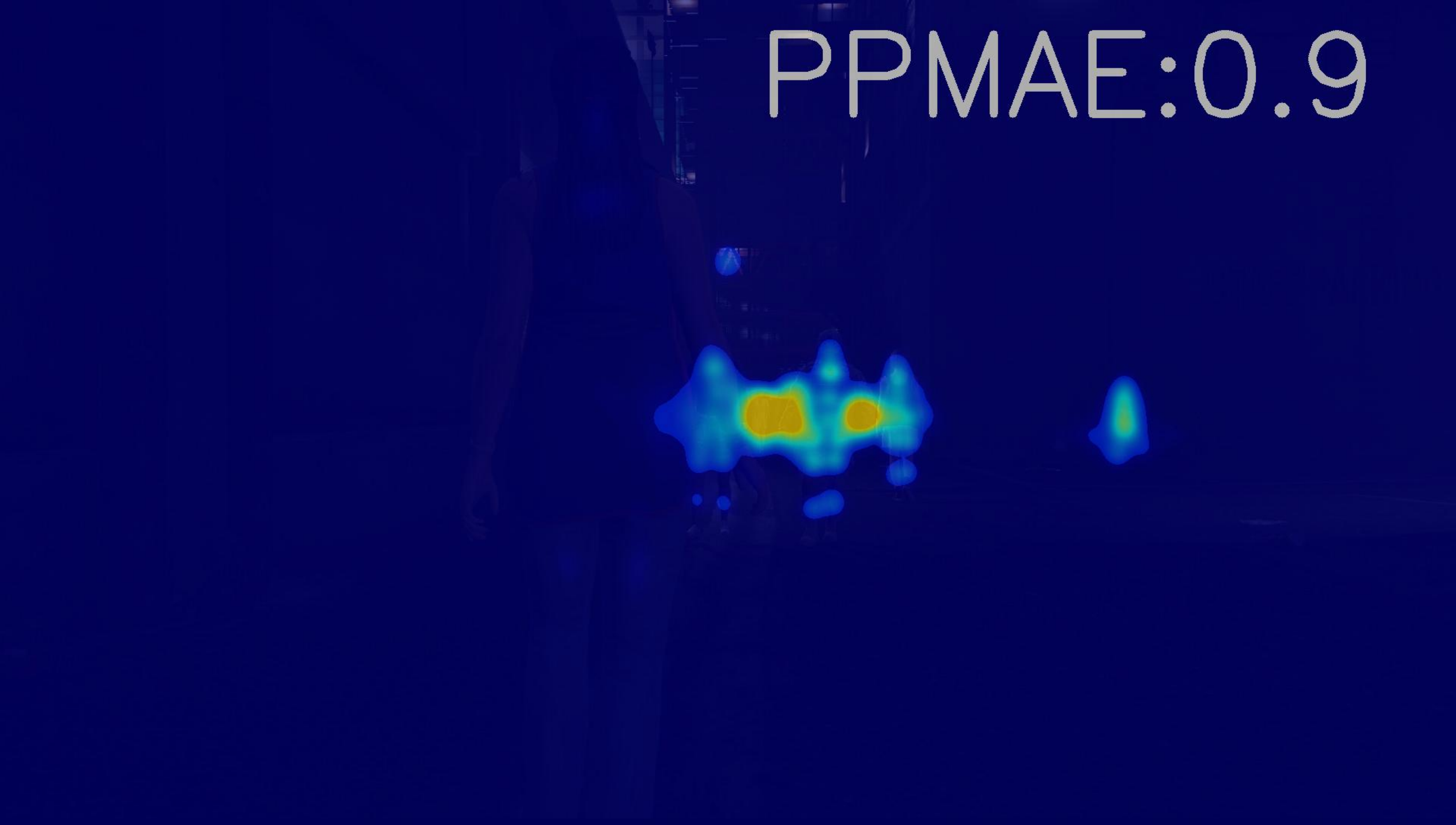} &
    
      \includegraphics[width=0.2\linewidth]{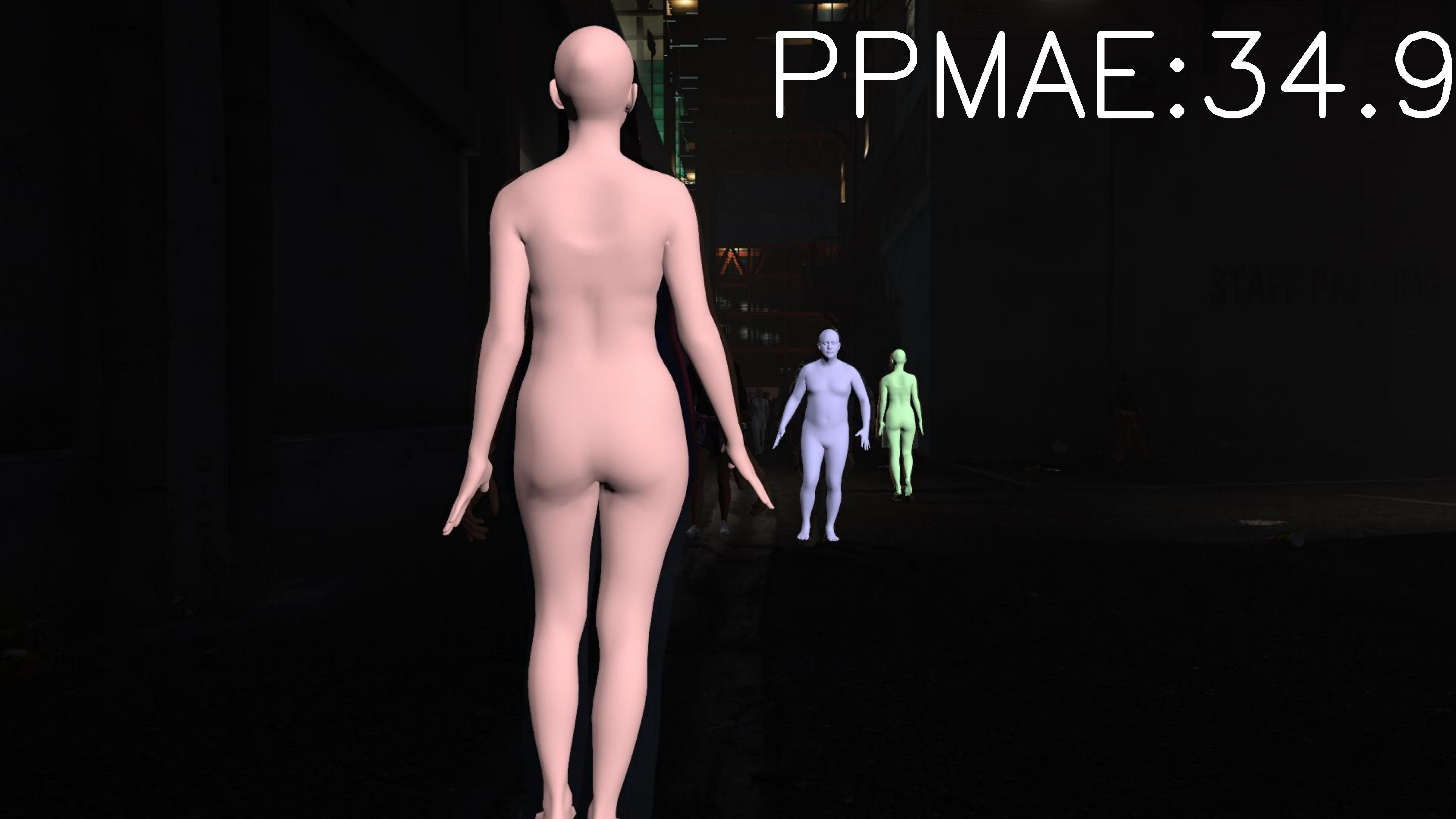} \\
       
\end{tabular}
\caption{Visual results of baseline models and STEERER-V on ANTHROPOS-V, along with the Ground Truth Per-Part-Volume Density Maps (GT Per Part). STEERER's concentrates the volume on the heads, whereas STEERER-V distributes it across the entire body.}\label{tab:qual_results}
\end{table*}

\subsection{From \ourdataset to real images}\label{sec:real_world}

We assess the transferability of models trained on \ourdataset to real imagery for CVE. Given the absence of suitable real-world datasets with volume annotations for crowds, we employ a bifurcated evaluation approach: we use crowd-centric real-world datasets, such as CrowdHumans~\cite{shao2018crowdhuman}, which lack volume annotations, and mesh-based real-world datasets, such as 3DPW~\cite{3dpw_von2018recovering}, which allow ground-truth volume computation but do not feature crowds.

For CrowdHumans~\cite{shao2018crowdhuman}, we address the lack of volume labels by imputing the average real-world volume~\cite{silverman2022exact} for each individual in the images. We compare these estimates with each model's predictions (Table \ref{table:real-world}). This experiment assesses the alignment of each model's predictions with expected crowd volumes, with STEERER-V and CLIFF being the most aligned. STEERER-V underestimates the expected volume by 3.40 dm³ per person, while CLIFF overestimates it by 1.00 dm³. Qualitative results on CrowdHumans are provided in the Supplementary Materials. 

For 3DPW~\cite{3dpw_von2018recovering}, we compare each model's predictions against ground-truth mesh volumes. However, several 3DPW images include unannotated persons, such as cameramen or unscripted passers-by. Since no ground-truth is available for these individuals, we manually excluded these images from our test set, reducing the original test set to 6989 images. Results indicate that STEERER-V trained on \ourdataset outperforms all baseline models (Table \ref{table:real-world}), with MAE and PPMAE registering at 40.40 and 25.28, respectively. 

Additionally, we evaluated STEERER-V trained on datasets from \cite{black2023bedlam} and \cite{patel2021agora} on 3DPW. In this scenario, STEERER-V continues to showcase superior results, with its counterparts presenting increased MAE and PPMAE to (59.72, 37.43) and (44.47, 29.15), respectively.
\begin{figure*}[t!]
    \begin{subfigure}[b]{0.3281\textwidth}
        \includegraphics[width=\textwidth]{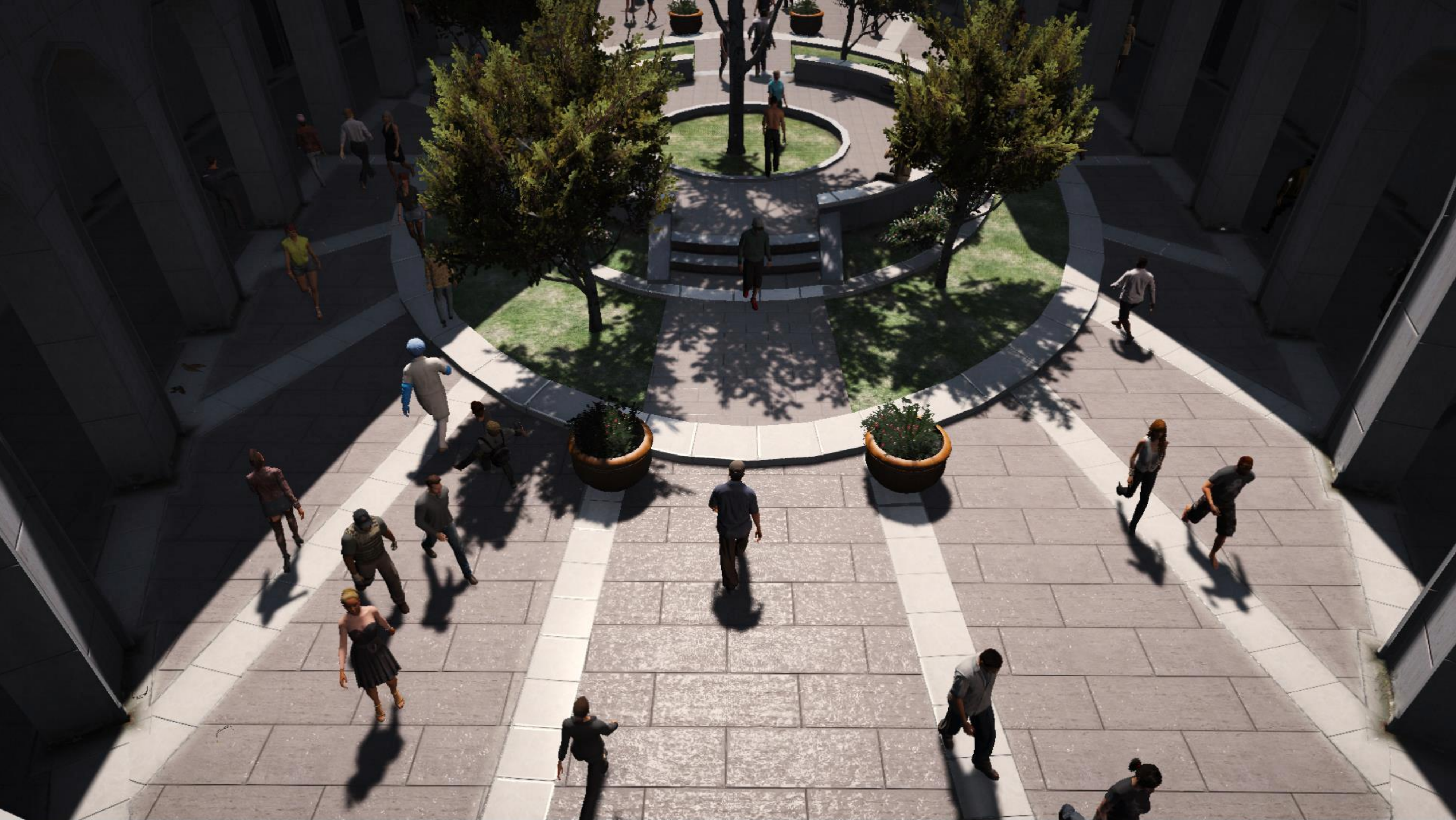}
        \label{fig:qualitative4}
    \end{subfigure}
    \hfill
    \begin{subfigure}[b]{0.3281\textwidth}
        \includegraphics[width=\textwidth]{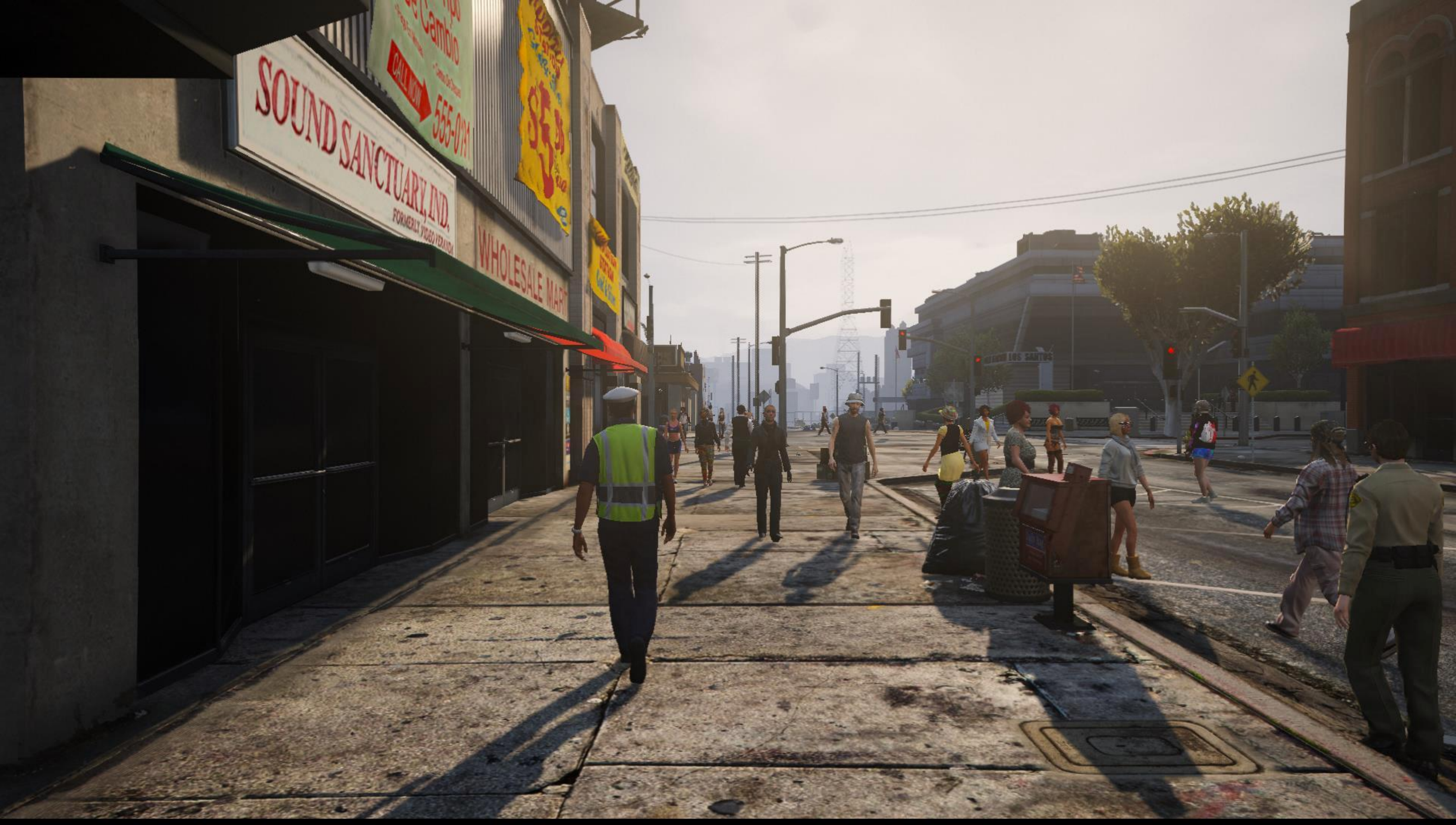}
        \label{fig:qualitative2}
    \end{subfigure}
    \hfill
    \begin{subfigure}[b]{0.3281\textwidth}
        \includegraphics[width=\textwidth]{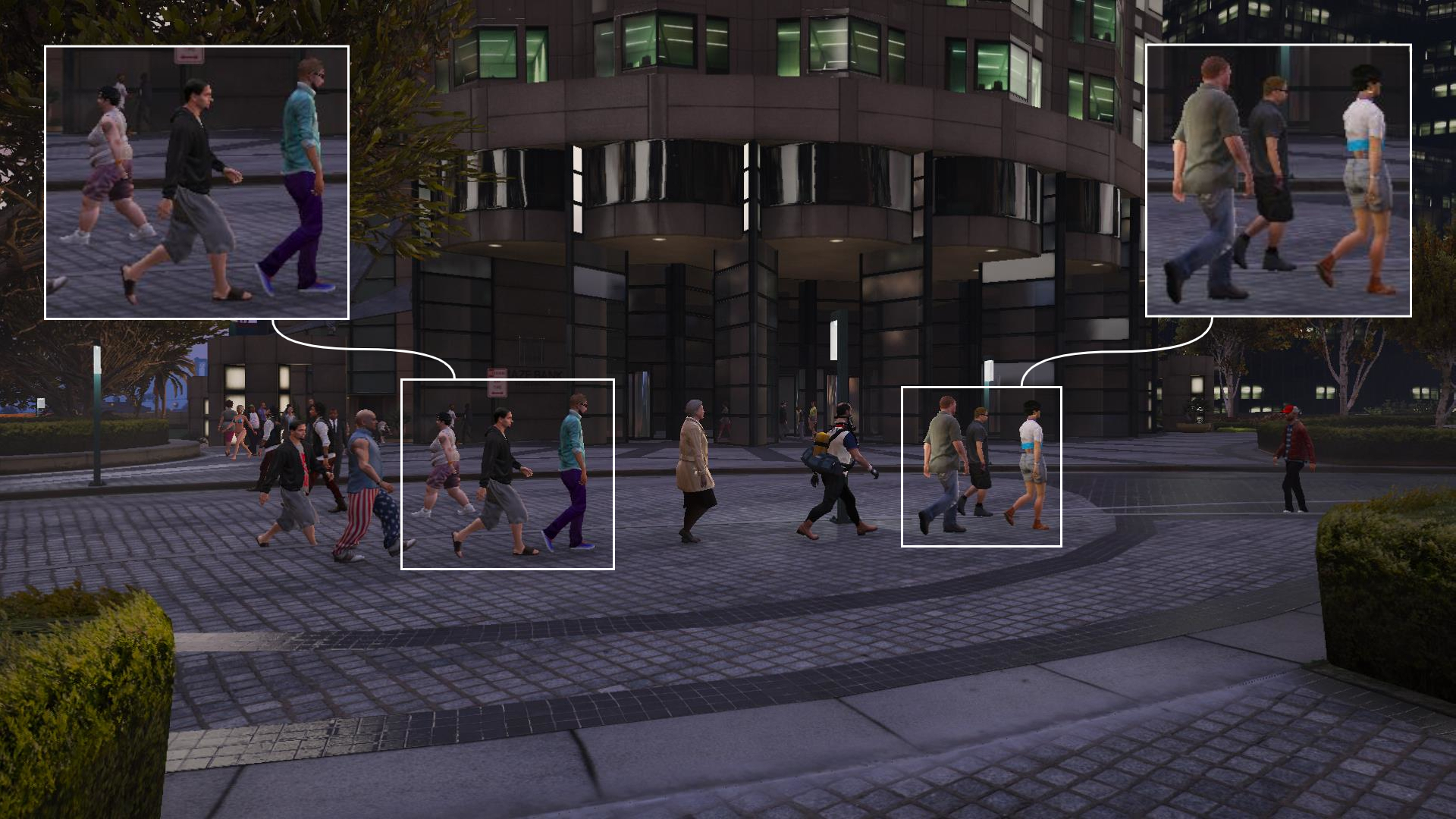}
        \label{fig:qualitative1}
    \end{subfigure}
    \vspace{-1.0em}
    \caption{Examples from ANTHROPOS-V, showcasing several lighting and weather conditions, camera angles, and a variety of physiques. The crops in the zoomed boxes depict persons with differences in statures and body shapes.}
    \label{fig:qualitatives-1-6}
    \vspace{-1.0em}
\end{figure*}

\vspace{-1em}
\section{The \ourdataset dataset}

Here we describe the generation of the proposed \ourdataset (Sec.~\ref{subsec:our-dataset}). We detail how we align in-game meshes to the real-world statistics (Sec.~\ref{sec:alignment}) and how we obtain SMPL meshes (Sec.~\ref{sec:smpl_fitting}). We also comment on \ourdataset statistics and annotations (Sec.~\ref{subsec:stats}).

\subsection{Dataset Generation}\label{subsec:our-dataset}

We construct \ourdataset{} exploiting the tools introduced in \cite{fabbri2018learning, fabbri2021motsynth}, which, leveraging the game engine from Grand Theft Auto V (GTA-V), allow us to create densely crowded scenes within photorealistic environments. GTA-V provides several 3D urban settings, with different weather and lighting conditions during day and night, and a broad array of characters with diverse appearances, as depicted in Fig.~\ref{fig:qualitatives-1-6}.
In addition, differently from previous GTA-based datasets~\cite{fabbri2018learning,fabbri2021motsynth}, to achieve a higher degree of photorealism, we use a professionally designed mod~\cite{gta-redux} that enhances the game graphics and improves the behavior and interaction among characters. Moreover, it offers additional atmospherical conditions and improves the physics in the scenes. 

\subsection{Alignment to real-world body-types}\label{sec:alignment}

\begin{figure*}[t!]
\centering
    \begin{subfigure}{0.8\linewidth}
        \includegraphics[width=\linewidth]{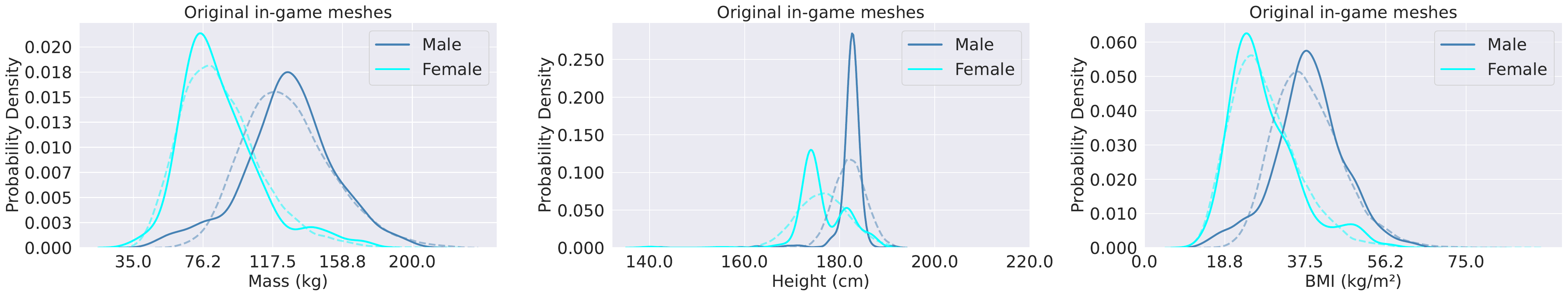}
        \caption{}\label{subfig:original_stats}
    \end{subfigure}
    \begin{subfigure}{0.8\linewidth}
        \includegraphics[width=\linewidth]{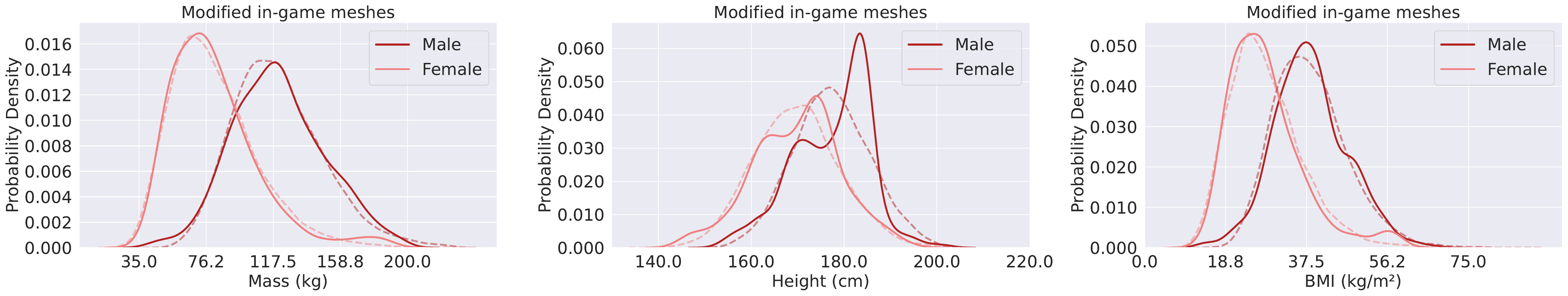}
        \caption{}\label{subfig:modified_stats}
    \end{subfigure}
    \begin{subfigure}{0.8\linewidth}
        \includegraphics[width=\linewidth]{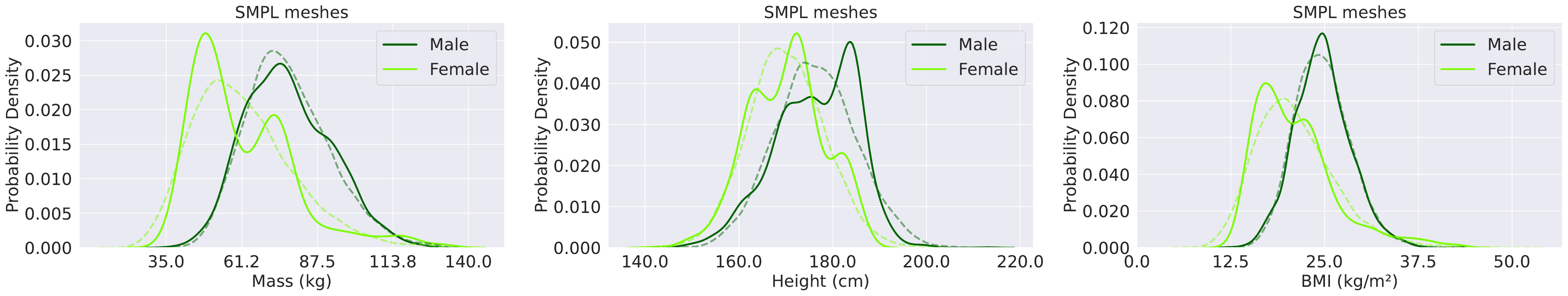}
        \caption{}\label{subfig:smpl_stats}
    \end{subfigure}
    \vspace{-0.5em}
    \caption{Statistical analysis of the distributions of mass, height, and \textit{Body Mass Index} (BMI) of the individuals in \ourdataset{}. Solid curves depict the empirical distributions, while dashed curves refer to the theoretically expected ones~\cite{silverman2022exact}. (\ref{subfig:original_stats}) Distribution of the body features of the characters' meshes in GTA-V without any manipulation. (\ref{subfig:modified_stats}) Distribution of the body features of the characters' meshes in GTA-V after applying some geometrical transformations. (\ref{subfig:smpl_stats}) Distribution of the body features of the resulting fitted SMPL meshes.}
    \vspace{-0.5em}
    \label{fig:statistical_analysis}
    \vspace{-0.5em}
\end{figure*}

The original GTA-V meshes exhibit a narrow range of variations in anthropometric features, with utterly repetitive heights and volumes and a noticeable imbalance in gender representation. To address this limitation, we carefully revise the in-game meshes and code and generate a distribution of individuals that closely mirrors the real-world one~\cite{owid-human-height} concerning height, volume, and gender.

To achieve this purpose, first, we conduct an in-depth statistical analysis of the distribution of the characters' anthropometrics in GTA-V. We consider the mass\footnote{medical literature refers to the body mass as \commas weight'', which in physics refers to another quantity; we stick with the physics definition.}, height, and \textit{Body Mass Index} (BMI) of the male and female characters in the game (from now on referred to as \commas in-game meshes''). We estimate the mass by multiplying the body volume by the average body density ($1000$ kg/m$^3$ as in \cite{durnin1974body, PfitznerLibra3D2015}).

As theoretically proven by \cite{silverman2022exact}, such body features can be represented with random variables $M, H, B$ that follow a log-normal distribution $\Lambda(\mu, \sigma^2)$:
\begin{equation}
    M \sim \Lambda(\mu_M, \sigma_M^2), \quad H \sim \Lambda(\mu_H, \sigma_H^2), \quad B = \frac{M}{H^2}
\end{equation}
Fig.~\ref{fig:statistical_analysis} shows the empirical distributions (solid lines) as opposed to the expected distributions (dashed lines). 

The body features of the original in-game meshes do not adhere to the theoretically expected ones, especially for the height that varies in a narrow range around the mean, as evident in the middle plot in Fig.~\ref{subfig:original_stats}. To mitigate such mismatch and increase the variance, we scale the in-game meshes along the three axes with scaling factors $\alpha, \beta, \gamma$ that we independently sample from truncated normal distributions; we carefully choose the hyperparameters for this step to avoid unfeasible and unnatural bodies and to end up with meshes that appear realistic (qualitative results of the scaling are reported in the Supplementary Material). The anthropometrics of the resulting meshes follow a distribution that improves the approximation  (Fig.~\ref{subfig:modified_stats}). Quantitatively, the Kullback-Leibler divergence between the empirical and the expected distributions, averaged across genders, decreases by 27.9\%, 63.3\%, and 19.8\% for mass, height, and BMI, respectively.
The SMPL fitting process (Sec.~\ref{sec:smpl_fitting}) disregards the clothing, thereby producing meshes that more closely match the real-world distributions of height~\cite{owid-human-height} and BMI~\cite{worldmean} (Fig.~\ref{subfig:smpl_stats}). As a final remark, the BMI of the SMPL meshes in \ourdataset{} ranges in $[10, 50]$ kg/m$^2$, representing also underweight and obese individuals. 
\vspace{-0.05em}
\subsection{SMPL Fitting}\label{sec:smpl_fitting}
To label each character with accurate ground truth volume, we employ a technique akin to the one described in \cite{patel2021agora}. 
The fitting procedure ensures that the SMPL mesh tightly conforms to the character mesh's uncovered body parts while allowing a looser fit on clothed parts. Details about this process are described in the Supplementary Materials.
We report that our SMPL meshes have an average per skin vertex error of $7.32$ mm and a penetration error of $10$ mm for clothed vertices, where a looser fit is desired. This measure indicates how much these vertices extend beyond the GTA-V mesh. Finally, we use the obtained meshes to compute ground-truth volume labels for each character. 
Notably, besides offering labels for the total body volume, \ourdataset{} includes annotations for the volume of individual body parts obtained by slicing the SMPL meshes. We divide the estimated meshes into nine sections: head, torso, thighs, left and right arms, forearms, and calves. 
We then calculate the volume of each of these parts separately.

\subsection{Dataset Statistics}\label{subsec:stats}

\ourdataset features 768 FHD videos with annotated volumes, SMPL shape parameters, keypoints, and camera parameters and position. 
Videos are recorded at 30 fps and display crowds moving in diverse urban scenarios.
\ourdataset features 701 distinct characters, each with a variable number of outfits, resulting in over 3k unique appearances, interacting in 384 diverse scenarios with different camera angles and weather conditions. To propose a fair split, we divide characters into three disjoint sets of \npourstrain, \npoursval, and \npourstest\ that we distribute in different train, validation, and test videos, respectively. Within crowded scenes, characters engage with each other and with the environment, undertaking interrelated actions. For instance, they avoid collisions and form queues to navigate stairs or enter confined areas.
 
\vspace{-0.5em}
\section{Limitations and Future Works}

As the first endeavor to establish a benchmark for Crowd Volume Estimation (CVE), our work lays the groundwork for this emerging field.
However, we acknowledge some aspects of our work that present opportunities for future refinement.

We introduced \ourdataset{} aiming to bridge the gap between synthetic and real-world data. While testing the transferability of the learned knowledge on real images without fine-grained and precise volume annotations may suffice to make an initial point on the validity of the dataset, future work should embark on acquiring detailed volume estimates of real images. Moreover, it may pursue even larger crowds, increasingly complex and diverse interactions, and estimates of objects other than people (e.g. backpacks, bags, etc.).

The current output of our model provides a single per-frame number representing the total crowd volume. While suitable for many applications, this approach encourages exploration into more granular spatial analyses that could further benefit fields such as civil engineering, where detailed volume distribution information might be valuable.
\\
Finally, we acknowledge that the ethical implications of CVE from images present complex challenges. Primary among these is the privacy issue in public spaces, which intersects with concerns about data security and the potential for misuse, as the underlying data could be adapted for unintended surveillance purposes.

Furthermore, bias in volume estimates due to potential underrepresentation in training data could lead to discriminatory applications. 
As CVE technology evolves, these ethical considerations underscore the critical need for robust guidelines and transparent deployment protocols to ensure that the benefits of CVE can be realized while safeguarding individual rights.

\vspace{-0.7em}
\section{Conclusion}

In this study, we have established the first benchmark for Crowd Volume Estimation. We introduced relevant metrics and developed a dataset specifically designed for this task, focusing on human crowds in real-world-like environments. Additionally, we evaluated baseline and oracular models adapted from Crowd Counting and Human Mesh Recovery domains. Furthermore, we proposed a novel supervision approach called \ppdensity, which we utilized to train STEERER-V, achieving superior results.
Given the challenges in gathering real-world datasets for CVE, we anticipate that introducing this new task and benchmark will ignite interest in the research community and inspire future endeavors in the field.

\section*{Acknowledgements}
We acknowledge financial support from
the PNRR MUR project PE0000013-FAIR and from the Sapienza grant RG123188B3EF6A80 (CENTS). Also, we acknowledge WSense and Chiara Petrioli for partially funding this work. Finally, we extend our gratitude to Matteo Fabbri for providing the materials essential for the initial setup of the dataset generation. This work has been carried out while Stefano D’Arrigo and Massimiliano Pappa were enrolled in the Italian National Doctorate on Artificial Intelligence run by Sapienza University of Rome.

{\small
\bibliographystyle{ieee_fullname}
\bibliography{main}
}

\end{document}


\title{Supplementary Materials: \\ ANTHROPOS-V: benchmarking the novel task  of Crowd Volume Estimation}

\maketitle

We supplement the main paper by outlining further notes on the SMPL fitting process and an additional experiment on estimating volumes of single body parts (Sec.~\ref{sec:smpl_fitting},~\ref{sec:analysis}). 
We complement Sec.~3.2 of the main paper with additional remarks on the task's evaluation metrics (Sec.~\ref{sec:notes_metrics},~\ref{sec:mae_vs_ppmae_qualitative}).
In addition, we show that \ourdataset{} can also serve as a benchmark for the tasks of \textit{Crowd Counting} and \textit{Human Mesh Recovery} (HMR) (Sec.~\ref{sec:other_tasks}).
Then, we illustrate more details on the implementation of baselines (Sec.~\ref{sec:baselines}), and we provide additional qualitative results, encompassing both success and failure cases on real and synthetic images (Sec.~\ref{sec:add_qualitatives},~\ref{sec:additional_results_steerer_v}). Additionally, we include some sample images from \ourdataset\ (Sec.~~\ref{sec:anthropos_qualitatives}). 
Finally, we present a cross-dataset evaluation, other remarks on CVE vs. Crowd Counting, and a tentative approach to leverage temporal information in CVE (Secs.~\ref{sec:cross}, \ref{sup:decoupling}, \ref{temporal}).

\section{Further notes on the SMPL fitting process}\label{sec:smpl_fitting}

The process of fitting SMPL meshes to characters, particularly in complex environments such as the Grand Theft Auto V (GTA-V) game, involves a complex combination of techniques from 3D modeling, computer vision, and machine learning. 

\begin{figure}[t!]
    \centering
    \includegraphics[width=0.8\linewidth]{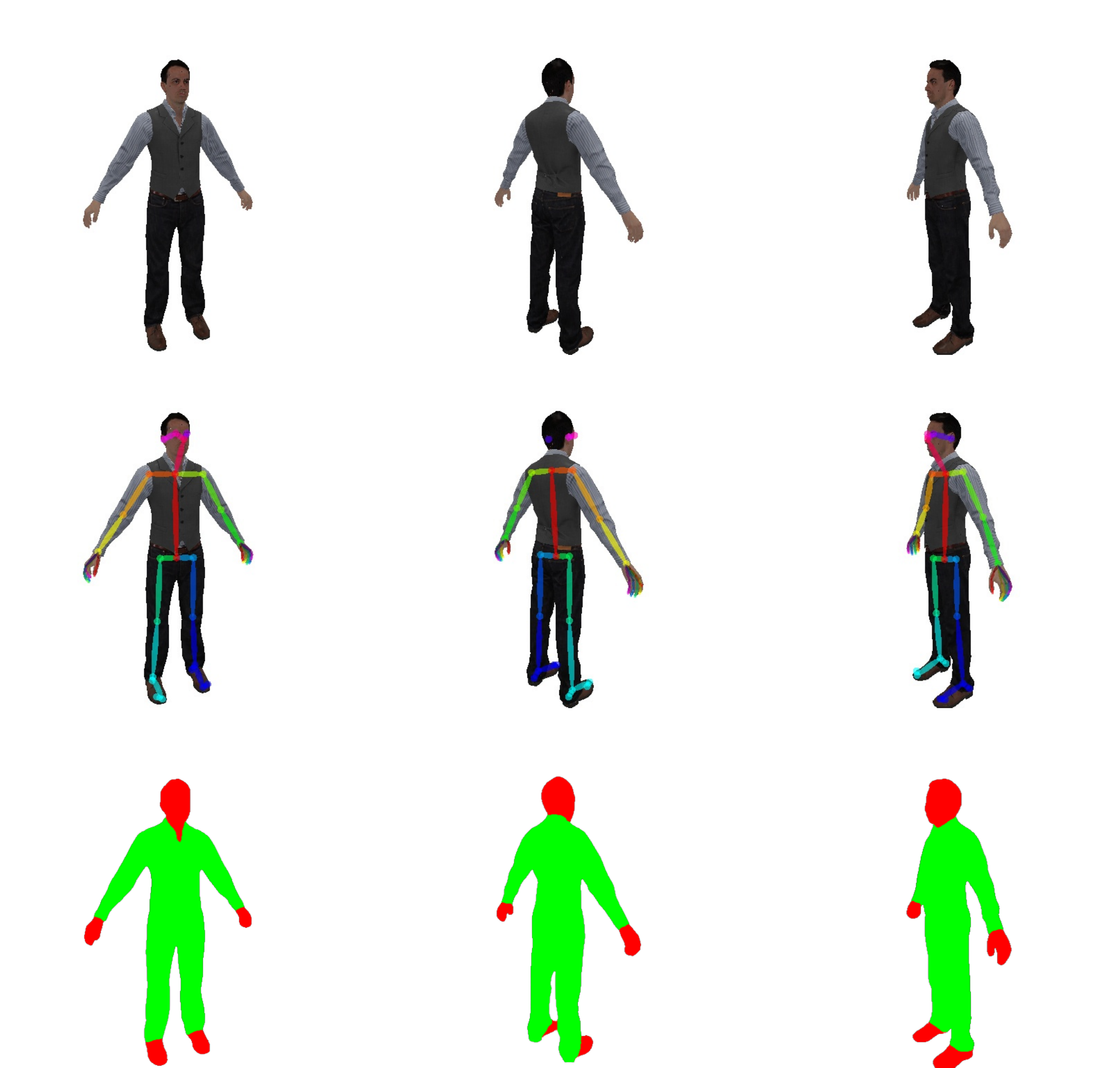}
    \caption{Renders from the SMPL mesh fitting process on a single character. The first line represents the renders, the second shows the estimated 2D pose for each render and the third is the Graphonomy output of skin segmentation (marked in red), as opposed to the dressed body segmentation (marked in green).
    Shoes are forced to be \commas{}skin'' points to improve the fitting.}
    \label{fig:renders}
\end{figure}

\begin{figure*}[ht!]

    \begin{subfigure}{0.50\linewidth}
        \includegraphics[width=\linewidth]{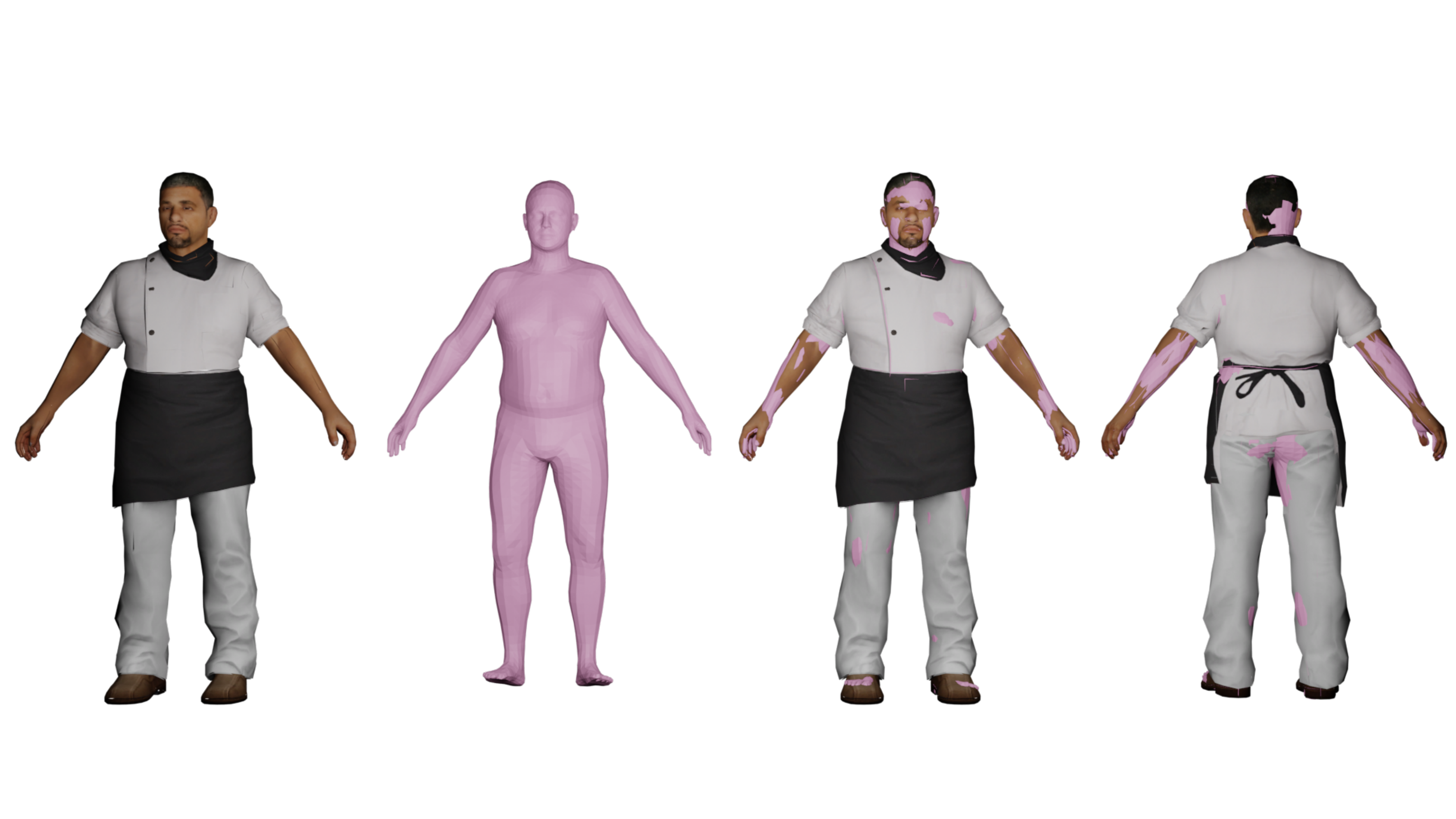}
    \end{subfigure}
    \begin{subfigure}{0.50\linewidth}
        \includegraphics[width=\linewidth]{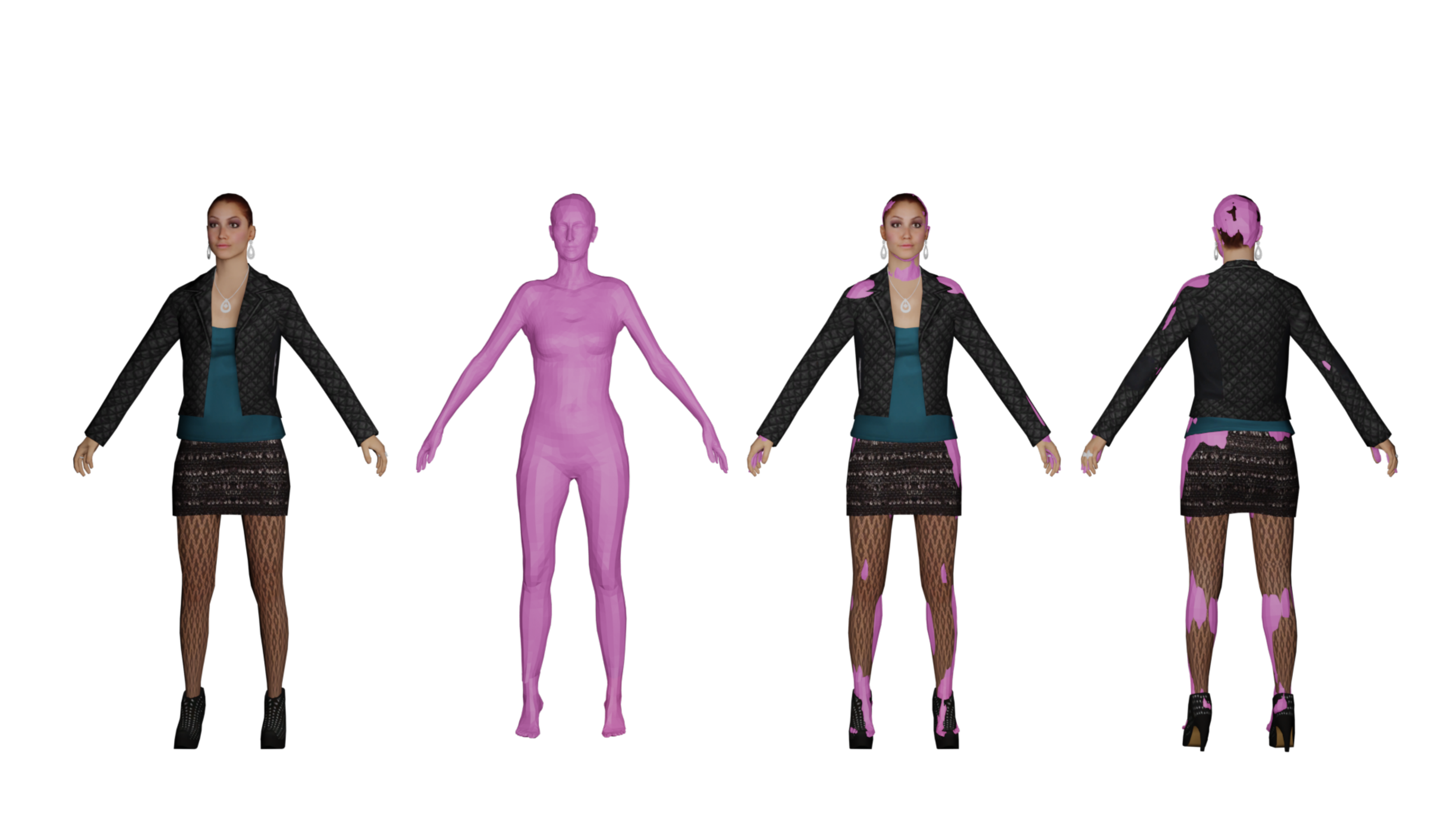}
    \end{subfigure}
    \caption{ Qualitative outcomes of the SMPL fitting process. Each image features: the original GTA-V character (first mesh), the output of the SMPL fitting process (second mesh), and an overlap of the GTA-V character with the SMPL result, in a front-facing view (third mesh), and in a backward-facing view (fourth mesh).}
    \label{fig:fitting}
\end{figure*}

We begin by collecting all the pre-existent meshes in the GTA-V game. The characters are identified by a name and a list of eleven variations that, in turn, express the contingent appearance of the character. It is worth noting that characters with the same name and different appearances do not necessarily share the same volume. Hence, we fit an SMPL mesh to all characters' variations appearing in each scene.
Initially, our fitting method retrieves characters' data, including their 3D models and texture information. The 3D models are then converted into the widely-used OBJ format (see Fig.~\ref{fig:fitting}, the first image of each sequence) accompanied by MTL files, which are required for defining the materials and textures of the model.
As in \cite{patel2021agora}, our objective is to achieve a tight fit that closely conforms to exposed bare-skin body parts such as the head or uncovered arms. Simultaneously, we seek a more relaxed fit in clothed body regions to diminish the impact of the added thickness introduced by clothing on the overall body volume.
To perform this fitting process, we need both a 3D pose prior and knowledge of which vertices in the GTA-V mesh represent skin or clothing. Thus, we initiate the process by generating 10 visual renders of the GTA-V 3D characters. This is achieved by moving the camera around the textured 3D mesh of the characters, as depicted in the first line of Fig.~\ref{fig:renders}.

Then, the pose estimation process exploits \cite{openpose} to predict the character's 2D pose in each rendered image, as depicted in the second line of Fig.~\ref{fig:renders}. This 2D pose data is lifted into a three-dimensional space, giving a complete spatial representation of the character's posture.
Next, the process of dividing a character's mesh into skin and clothes vertices leverages \cite{graphonomy} to segment each of the 10 renderings (see Fig.~\ref{fig:renders}, third line). The resulting segmentation is reprojected onto the mesh to label each vertex. 

Before fitting the SMPL mesh, the character's gender is determined, which ensures the accuracy of the SMPL model, as these models are gender-specific. The SMPL fitting involves aligning a standard human body model to the character's 3D pose and shape. This step requires meticulous adjustments to ensure that the SMPL mesh accurately follows the contours and posture of the character. Indeed, following~\cite{patel2021agora}, we employ two different loss functions to constrain the SMPL within the GTA-V mesh. The first loss is applied to the retrieved skin vertices, where we impose a severe fitting. The second applies to the clothes vertices, where we aim to have a looser fit so that these vertices would not penetrate the original mesh while remaining sufficiently close to it.
Once the fitting process is over, the volume of the fitted SMPL mesh is computed using Blender~\cite{blender} Python API, which calculates the volume within a mesh. 
A qualitative assessment of our fitting is present in Fig.~\ref{fig:fitting}. It is worth noting that our SMPL meshes typically lie beneath the attire of the GTA-V characters, with minimal penetration occurring primarily at skin vertices. This penetration is expected since we want a tighter fit in these specific areas.

\begin{figure}[t!]
  \centering
  
  \includegraphics[width=0.5 \linewidth]{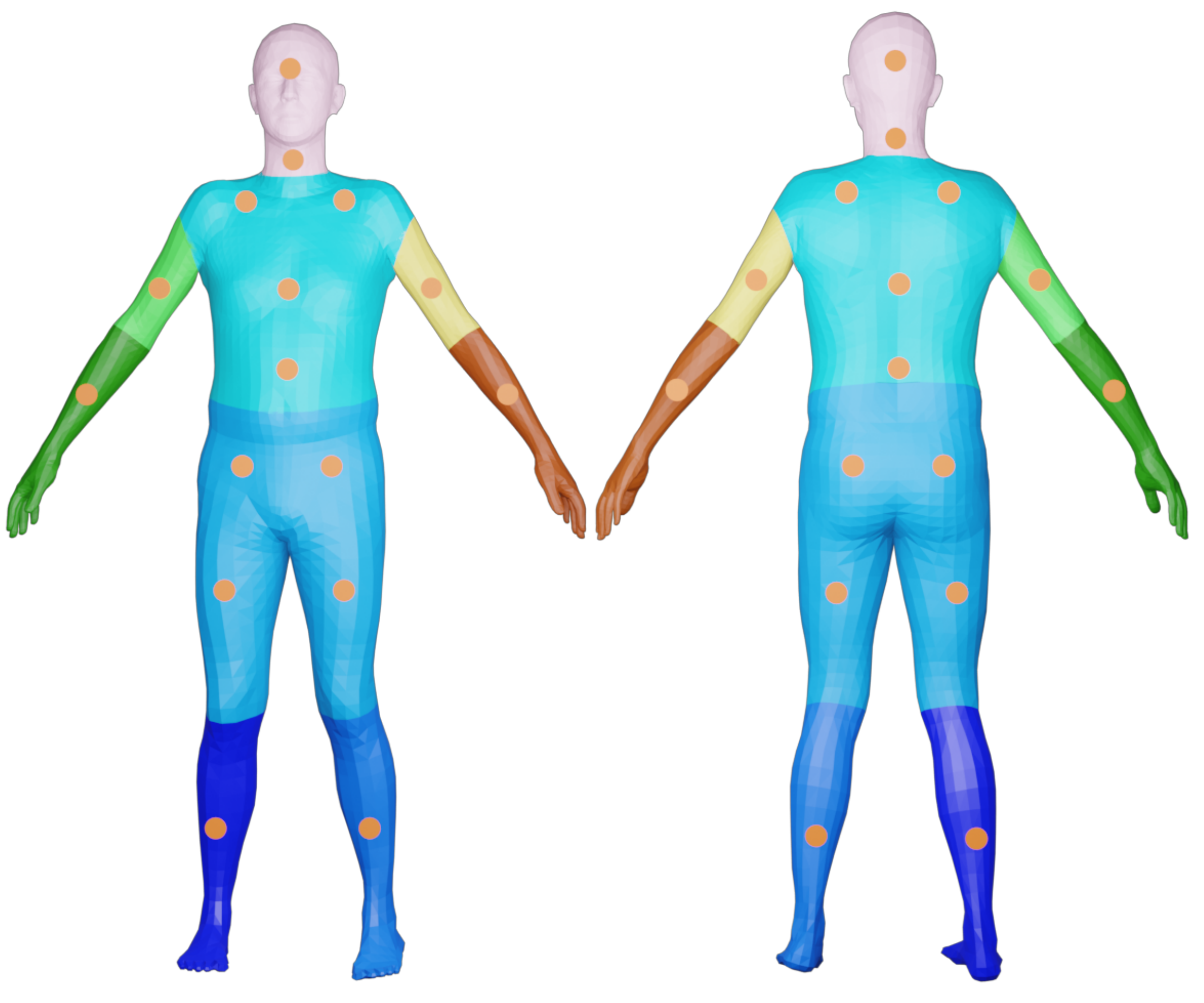}
  \caption{A SMPL mesh from \ourdataset{}. Different body parts are highlighted in different colors. Orange dots represent the keypoints associated with each body part.  
  }
  \label{fig:split}
\end{figure}

Furthermore, to assign volume labels to individual body segments, we partition SMPL meshes into nine 3D parts, as illustrated in Fig.~\ref{fig:split}. The segmentation process relies on the body segmentation mappings presented in \cite{loper2015smpl}. These mappings provide the indices of vertices corresponding to each body part, enabling the identification of boundary vertices situated between adjacent body parts. We split the meshes into disjointed body parts along these identified boundaries. Since boundary vertices often do not lie on a common plane, we identify the plane that traverses the maximum number of them while reporting the minimum distance from the non-traversed boundary vertices. Finally, we employ these planes to split the meshes into distinct body parts and compute their volumes. 
\section{Further Analysis of Crowd Volume Estimation Models}\label{sec:analysis}

\begin{table}[ht]

\centering
\resizebox{\linewidth}{!}{
\begin{tabular}{l|cccccc}
\toprule
 & \multicolumn{6}{c}{\textbf{Body Part}} \\
\textbf{Metric} & Head & Arms & Forearms & Torso & Thighs & Calves\\
\midrule
MAE & 19.161 & 29.230 & 55.504 & 398.73 & 90.072 & 280.22\\
PP-MAE & 1.1036 & 1.6920 & 3.6455 & 21.504 & 6.7060 & 20.707 \\
\bottomrule
\end{tabular}}
\vspace{0.5em}
\caption{Volume error for each part of the body. The results are reported in dm\(^3\).
}\label{tab:vol_err_poi}
\end{table}

To further investigate the performance of STEERER-V on Crowd Volume Estimation, we conducted additional experiments to assess its ability to localize volume within images. Specifically, we divided the images into random-scale patches and evaluated whether STEERER-V could accurately allocate the correct volume to each patch. The results (MAE: 130.2, PP-MAE: 5.8) are consistent with those from our main experiments, demonstrating that the model effectively distributes volume across the correct individuals.

Additionally, we assess STEERER-V capability to estimate the volume of single body parts.  Specifically, we train our proposed model to estimate the volume of the single body parts' split presented in Sec.~4.3 of the main paper.
Results of this experiment are reported in Table~\ref{tab:vol_err_poi}.
While the error on the estimated volume of the head and arms is low, other parts like the torso and thighs expose a greater error due to a superior volume occupancy and loose-fitting clothes, rendering the correct estimation more challenging.
Other body parts like calves and forearms have a high probability of being partially occluded or self-occluded, leading to a higher error compared to body parts with nearly the same volume coverage. 

\section{Further notes on the metrics}\label{sec:notes_metrics}

\begin{table}[!t]
    \centering
    \resizebox{0.7\linewidth}{!}{
    \begin{tabular}{l|c}
        \toprule
        \textbf{Model} & \textbf{RMSE}\\
        \midrule
        CLIFF~\cite{li2022cliff} & 862.4\\
        BEDLAM-CLIFF~\cite{black2023bedlam} & 827.1\\
        ReFit~\cite{wang2023refit} & 708.3\\
        \midrule
        \textcolor{gray}{Oracular CLIFF}~\cite{li2022cliff} & \textcolor{gray}{473.9}\\
        \textcolor{gray}{Oracular BEDLAM-CLIFF}~\cite{black2023bedlam} & \textcolor{gray}{459.5}\\
        \textcolor{gray}{Oracular ReFit}~\cite{wang2023refit} & \textcolor{gray}{412.7}\\
        \midrule
        \textcolor{gray}{$C_{B+}(I) \times \bar{V}_\mathcal{D}$} & \textcolor{gray}{638.29}\\ 
        Bayesian+~\cite{ma2019bayesian} & 904.23\\
        P2P~\cite{song2021rethinking} & 743.81\\
        MAN~\cite{lin2022boosting} & 915.64 \\
        STEERER~\cite{hani2023steerer} & 643.10\\
        \midrule
        \textcolor{gray}{Oracular $C(I) \times \bar{V}_\mathcal{D}$} & \textcolor{gray}{254.91}\\
        \midrule
        STEERER-V~\cite{hani2023steerer} & \textbf{269.39}\\
        \bottomrule
        \end{tabular}
    }
    \caption{
    Results on \ourdataset, reported in dm$^3$. Methods are divided into HD+HMR, Crowd Counting, and our proposed approach. Gray-out lines rely on some oracular information and shouldn’t be directly compared with the other results.}\label{tab:rmse_results}
\end{table}

In Sec.~3.2 of the main manuscript, we introduced the minimal set of metrics for the Crowd Volume Estimation (CVE) task, particularly \textit{Mean Absolute Error} (MAE) and \textit{Per-Person Mean Absolute Error} (PP-MAE). We are aware that some literature on the task of \textit{Crowd Counting} also reports the \textit{Root Mean Squared Error} (RMSE). We argue that given a set of images $\{I_k\}$, RMSE is redundant for CVE, as it is proportional to MAE. We show this in Eq.~\ref{eq:proportional}, where $\{V_k\}$ is the total volume associated with each image, $\{\hat{V}_k\}$ the estimated one, and $\mathrm{AE}(k)$ is the absolute error of the $k$-th image.

\begin{align}
        \begin{split}
            \mathrm{RMSE}(\{I_k\}) &=\sqrt{\frac{1}{K} \underset{k=1}{\overset{K}{\sum}} (V_k - \hat{V}_k)^2} \\
            &= \frac{1}{\sqrt{K}} \sqrt{\underset{k=1}{\overset{K}{\sum}} |V_k - \hat{V}_k|^2} \\
            &=\frac{1}{\sqrt{K}} \sqrt{\underset{k=1}{\overset{K}{\sum}} [\mathrm{AE}(k)]^2} \\
            &\propto \frac{1}{K} \underset{k=1}{\overset{K}{\sum}} \mathrm{AE}(k) = \mathrm{MAE}(\{I_k\})\\
        \end{split}
\label{eq:proportional}
\end{align}

Nonetheless, we extend Table~1 of the main paper with Table~\ref{tab:rmse_results}, showing the RMSE for the proposed models.
\section{Qualitative Evaluation: MAE vs PP-MAE}\label{sec:mae_vs_ppmae_qualitative}

In this section, we complement the analysis presented in Fig.~2 of the main paper by providing qualitative examples of instances where the MAE and PP-MAE exhibit notable misalignment, deviating significantly from the primary trend observed in the graph.

\begin{figure*}[ht!]
    \centering
    \begin{subfigure}{\linewidth}
    \centering
        \begin{subfigure}{0.37\linewidth}
            \includegraphics[width=\linewidth]{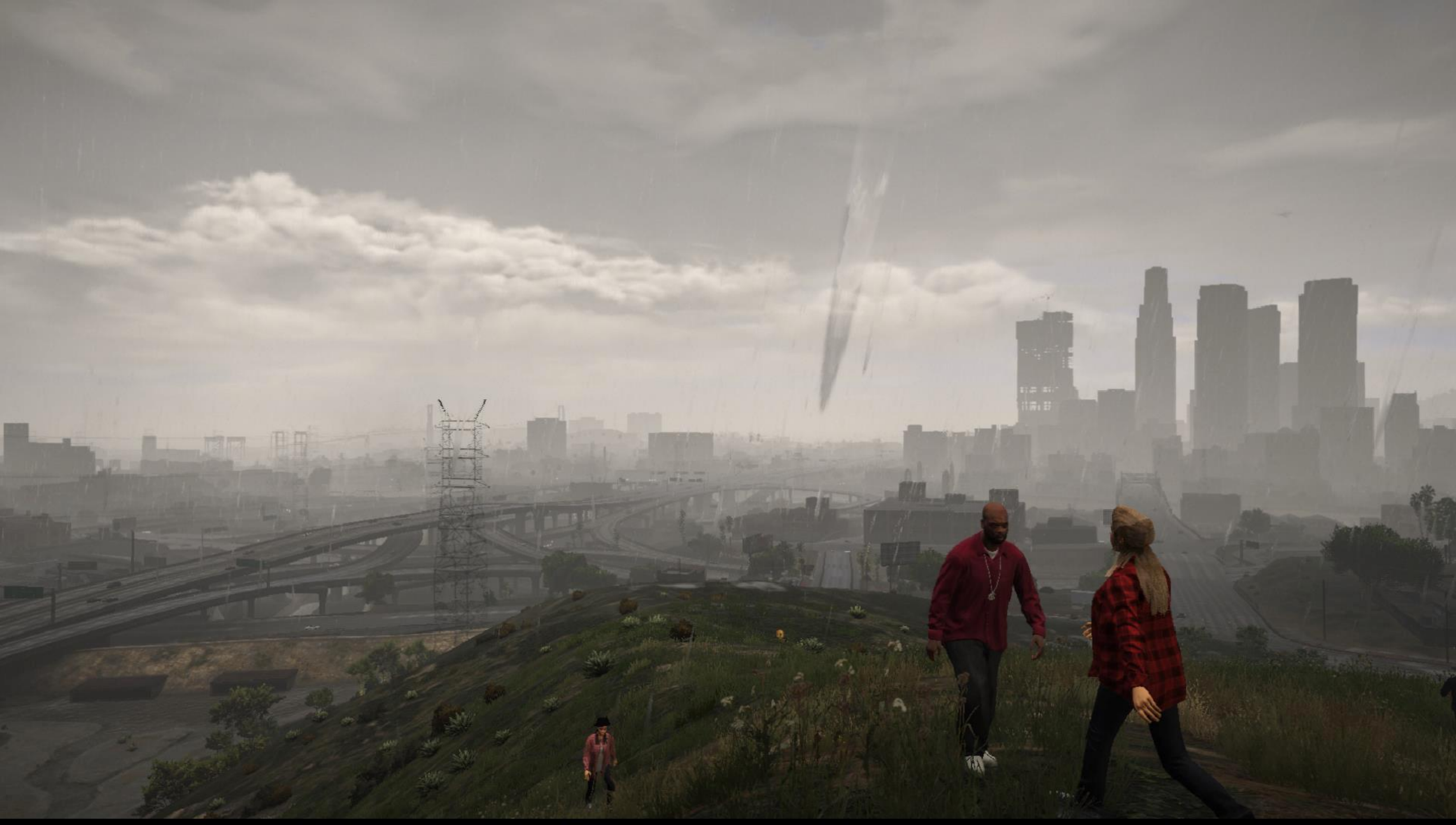}
        \end{subfigure}\hspace{.5em}
        \begin{subfigure}{0.37\linewidth}
            \includegraphics[width=\linewidth]{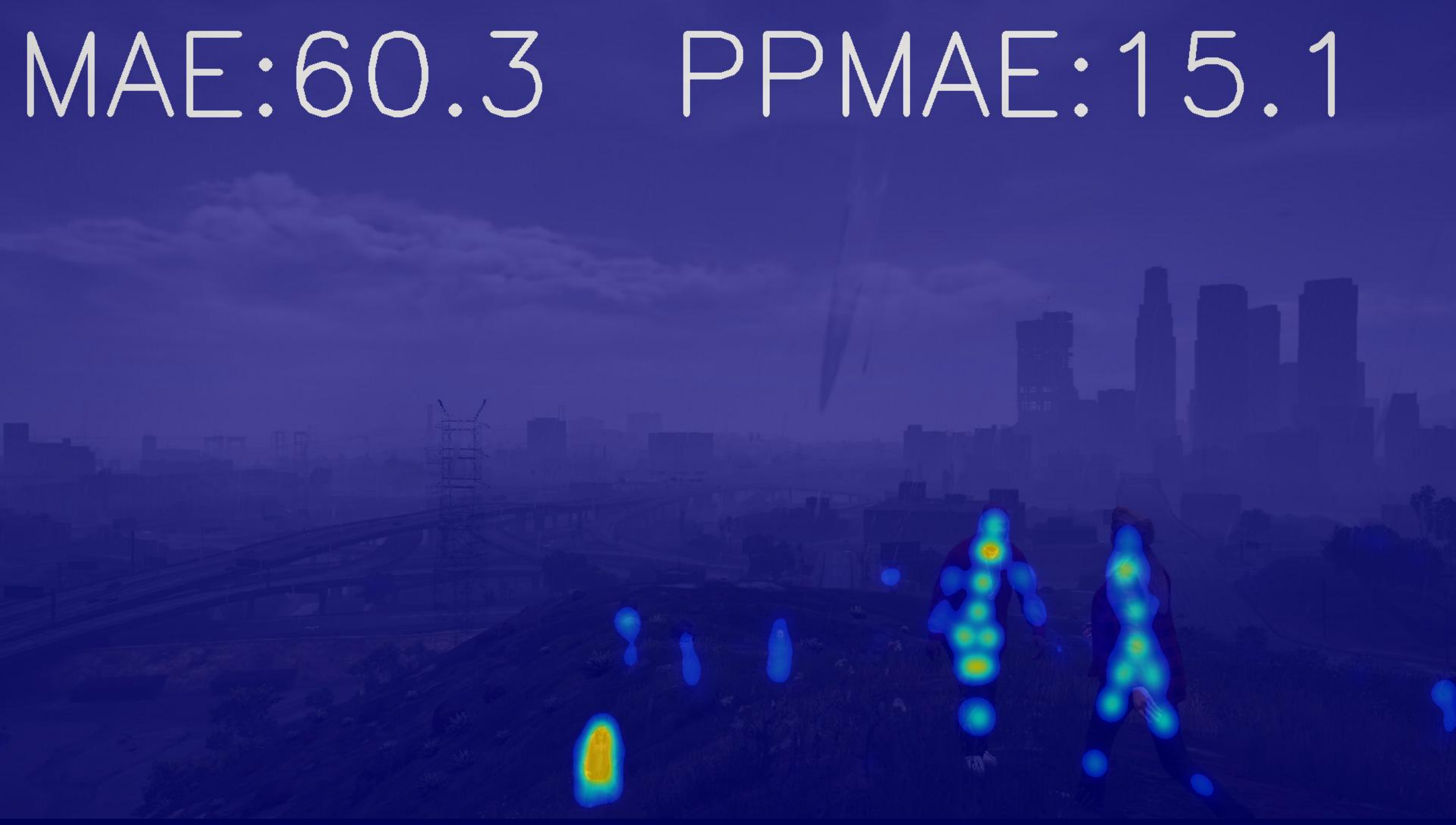}
        \end{subfigure}
        \caption{} 
        \label{subfig:a}
    \end{subfigure}

    \begin{subfigure}{\linewidth}
    \centering
        \begin{subfigure}{0.37\linewidth}
            \includegraphics[width=\linewidth]{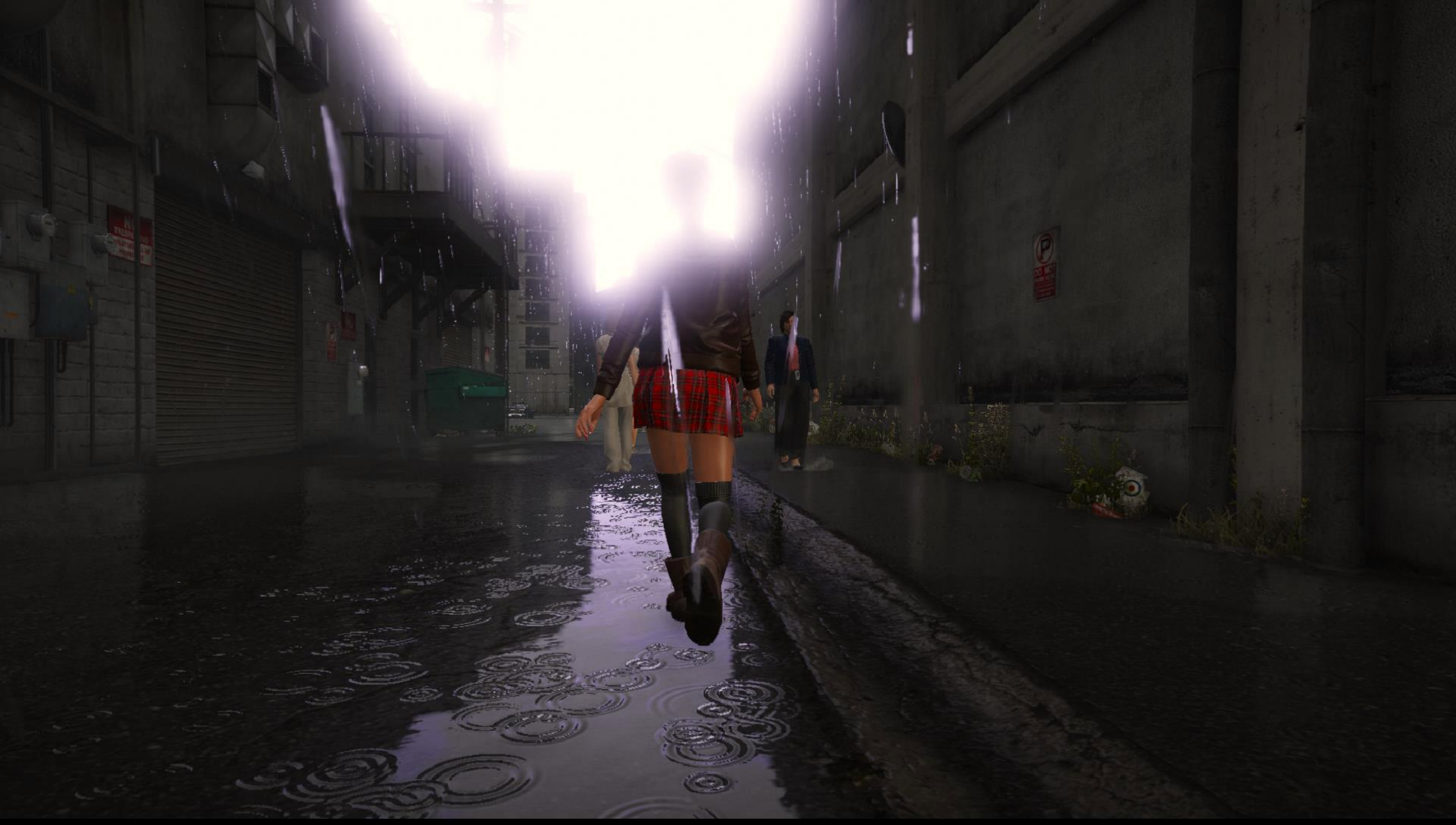}
        \end{subfigure}\hspace{.5em}
        \begin{subfigure}{0.37\linewidth}
            \includegraphics[width=\linewidth]{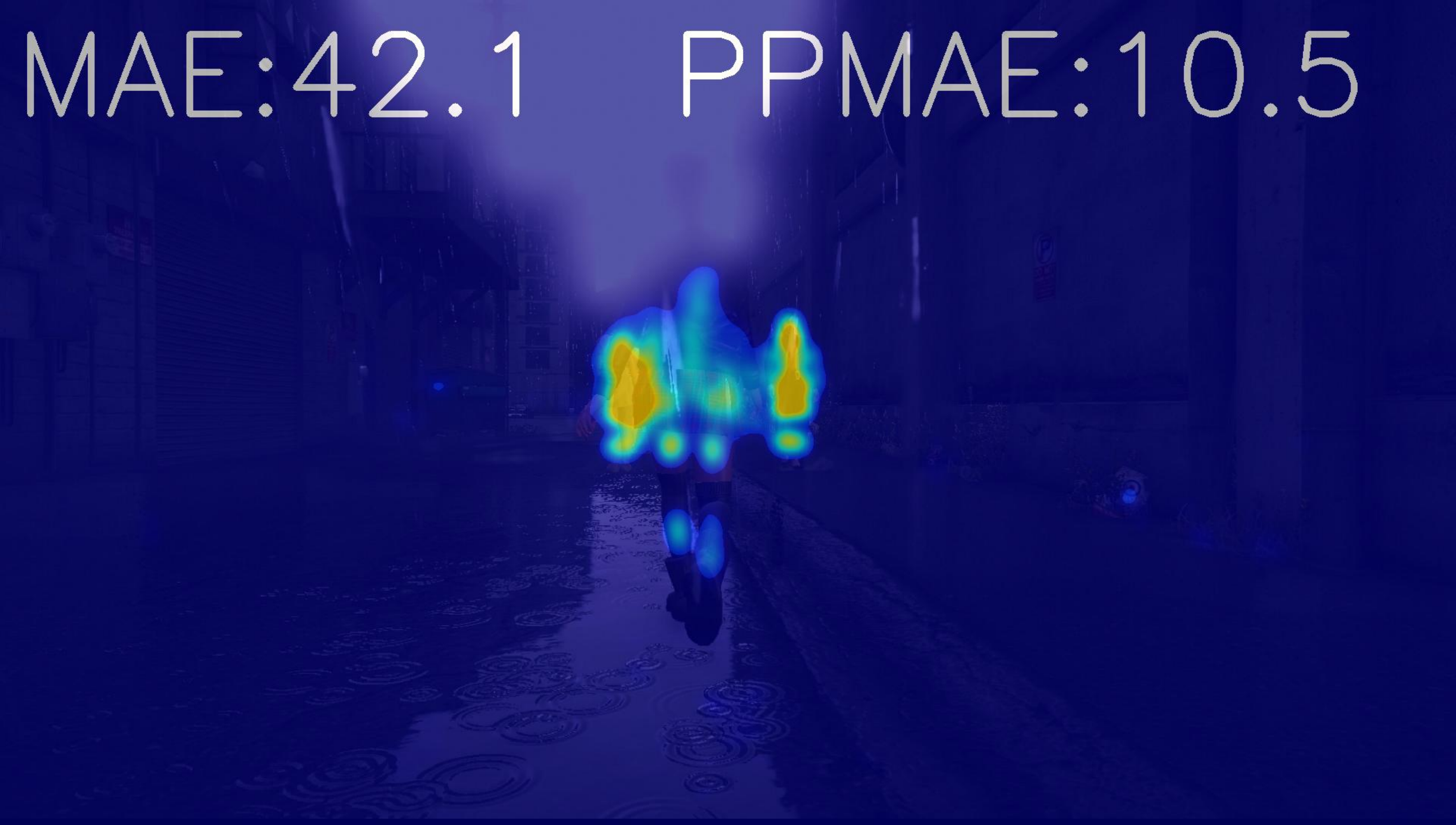}
        \end{subfigure}
        \caption{} 
        \label{subfig:b}
    \end{subfigure}

    \begin{subfigure}{\linewidth}
    \centering
        \begin{subfigure}{0.37\linewidth}
            \includegraphics[width=\linewidth]{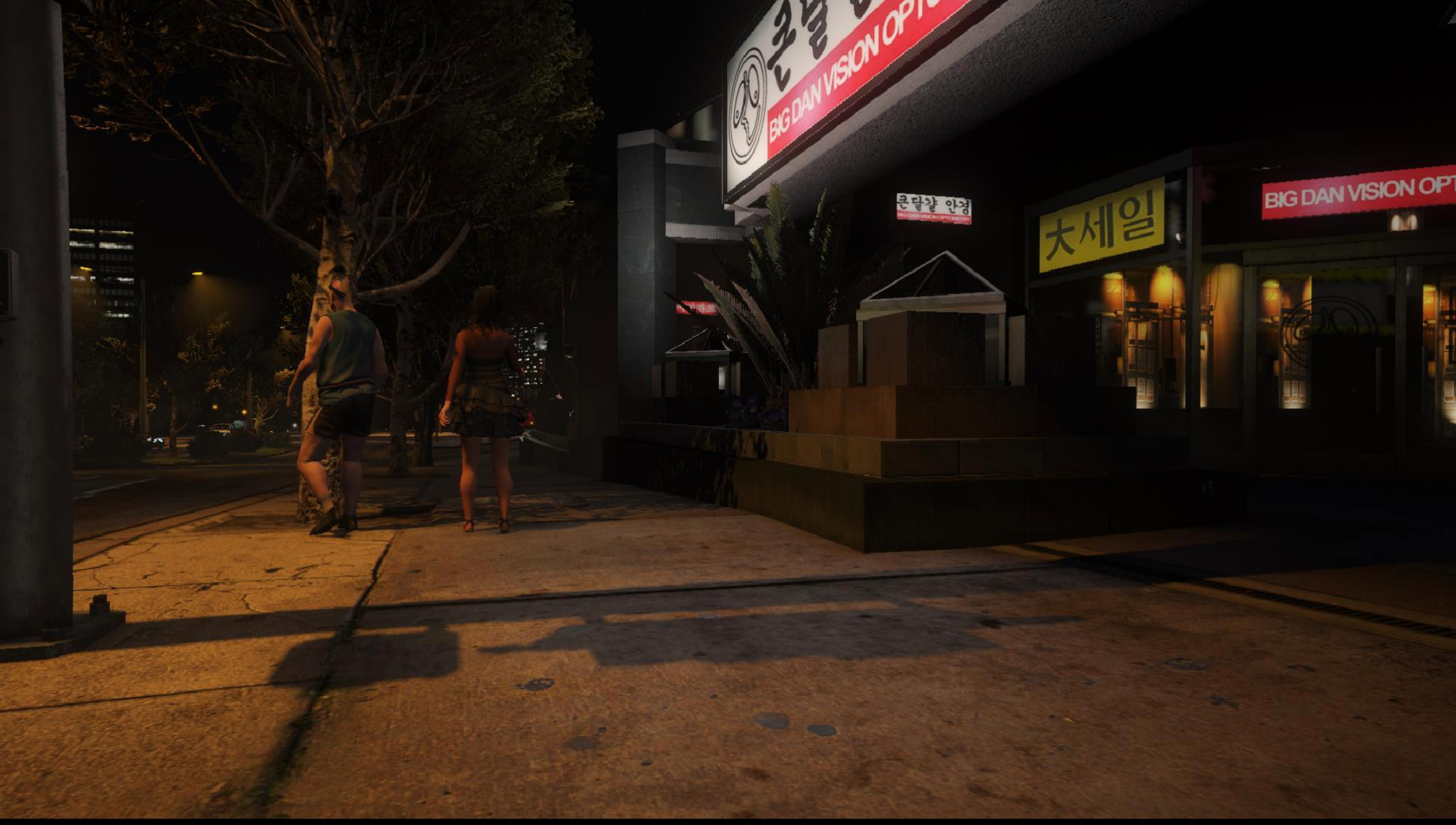}
        \end{subfigure}\hspace{.5em}
        \begin{subfigure}{0.37\linewidth}
            \includegraphics[width=\linewidth]{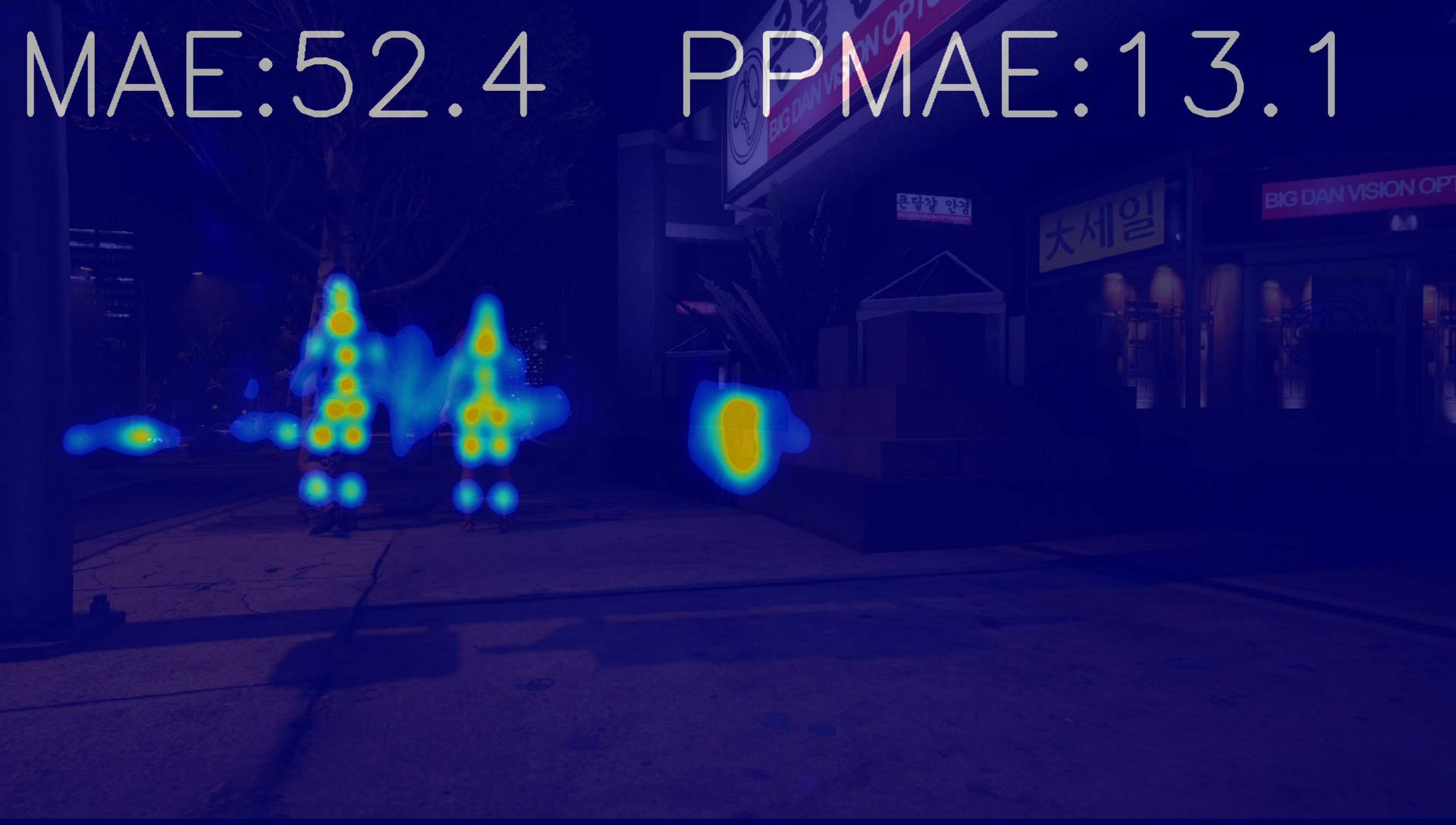}
        \end{subfigure}
        \caption{} 
        \label{subfig:c}
    \end{subfigure}

    \begin{subfigure}{\linewidth}
    \centering
    \begin{subfigure}{0.37\linewidth}
            \includegraphics[width=\linewidth]{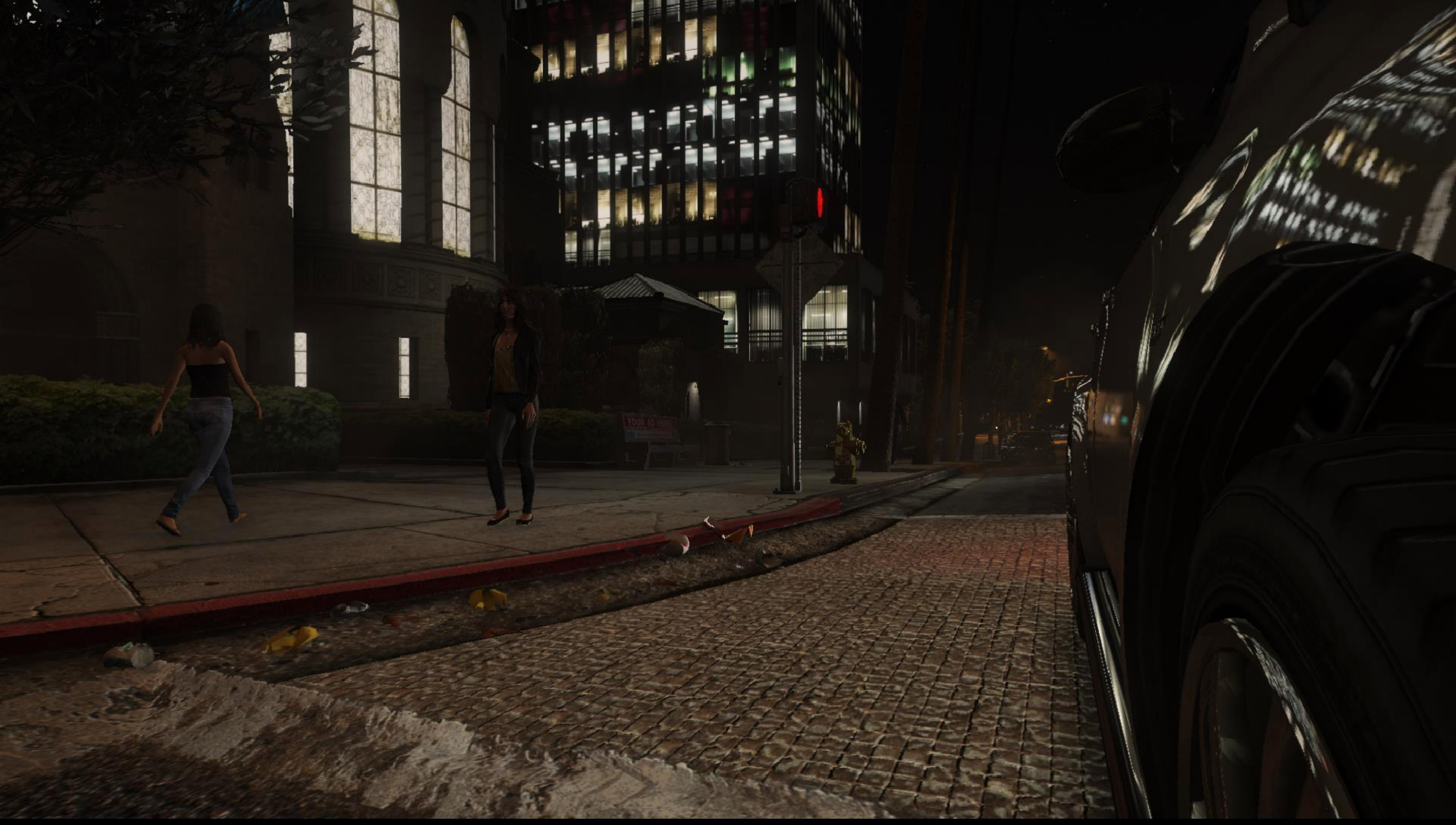}
        \end{subfigure}\hspace{.5em}
        \begin{subfigure}{0.37\linewidth}
            \includegraphics[width=\linewidth]{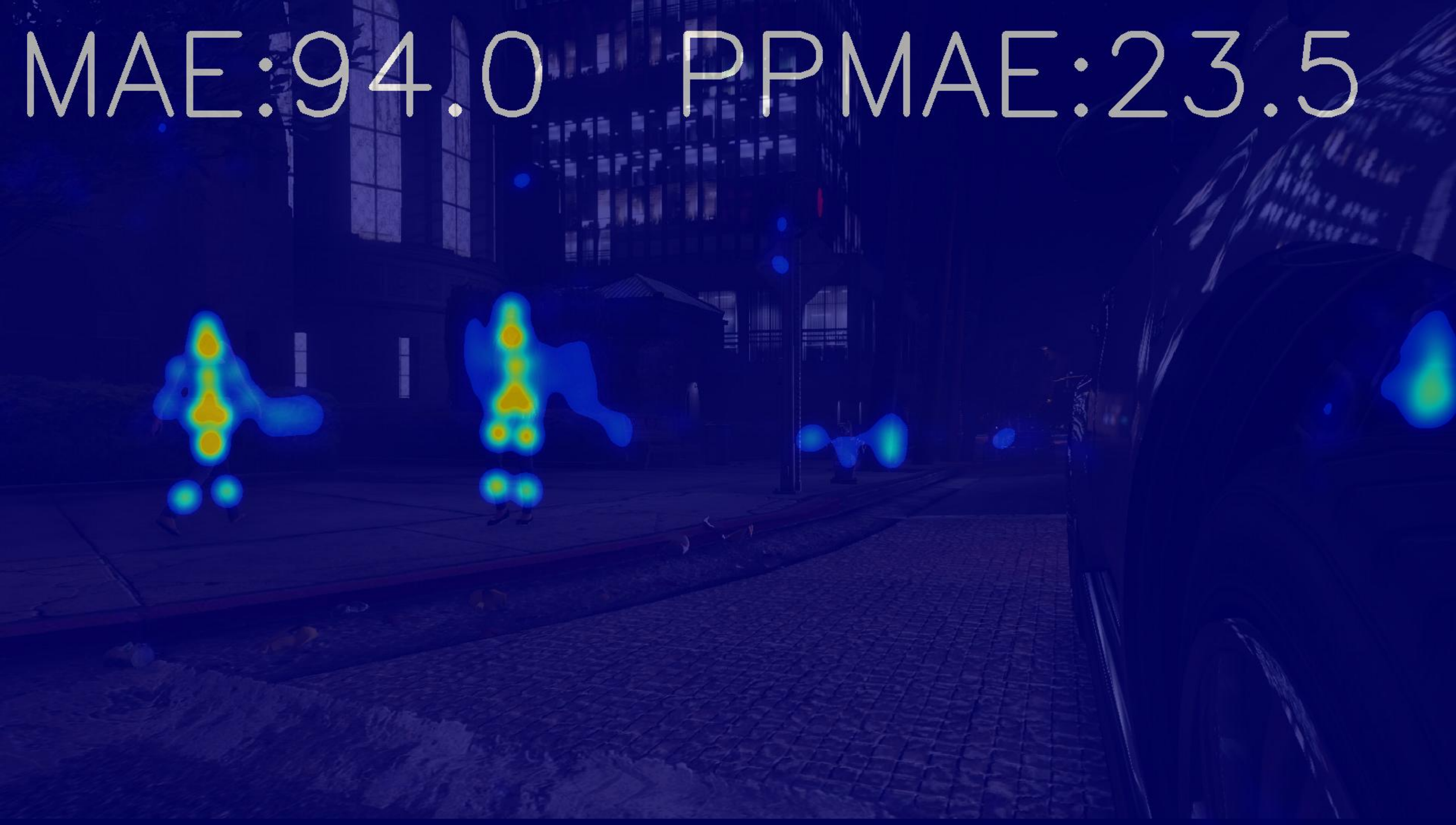}
        \end{subfigure}
        \caption{} 
        \label{subfig:d}
    \end{subfigure}

    \begin{subfigure}{\linewidth}
    \centering
        \begin{subfigure}{0.37\linewidth}
            \includegraphics[width=\linewidth]{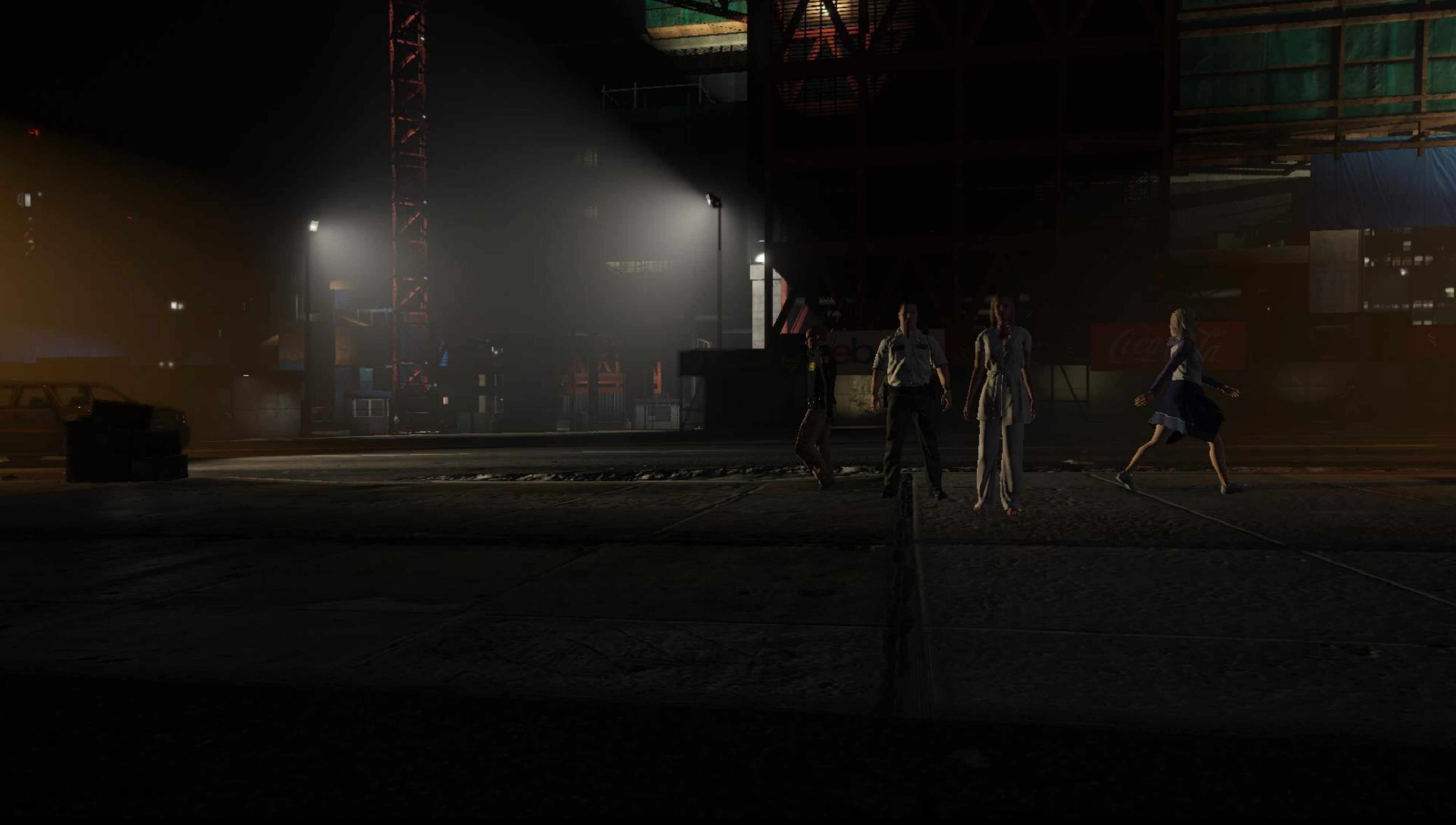}
        \end{subfigure}\hspace{.5em}
        \begin{subfigure}{0.37\linewidth}
            \includegraphics[width=\linewidth]{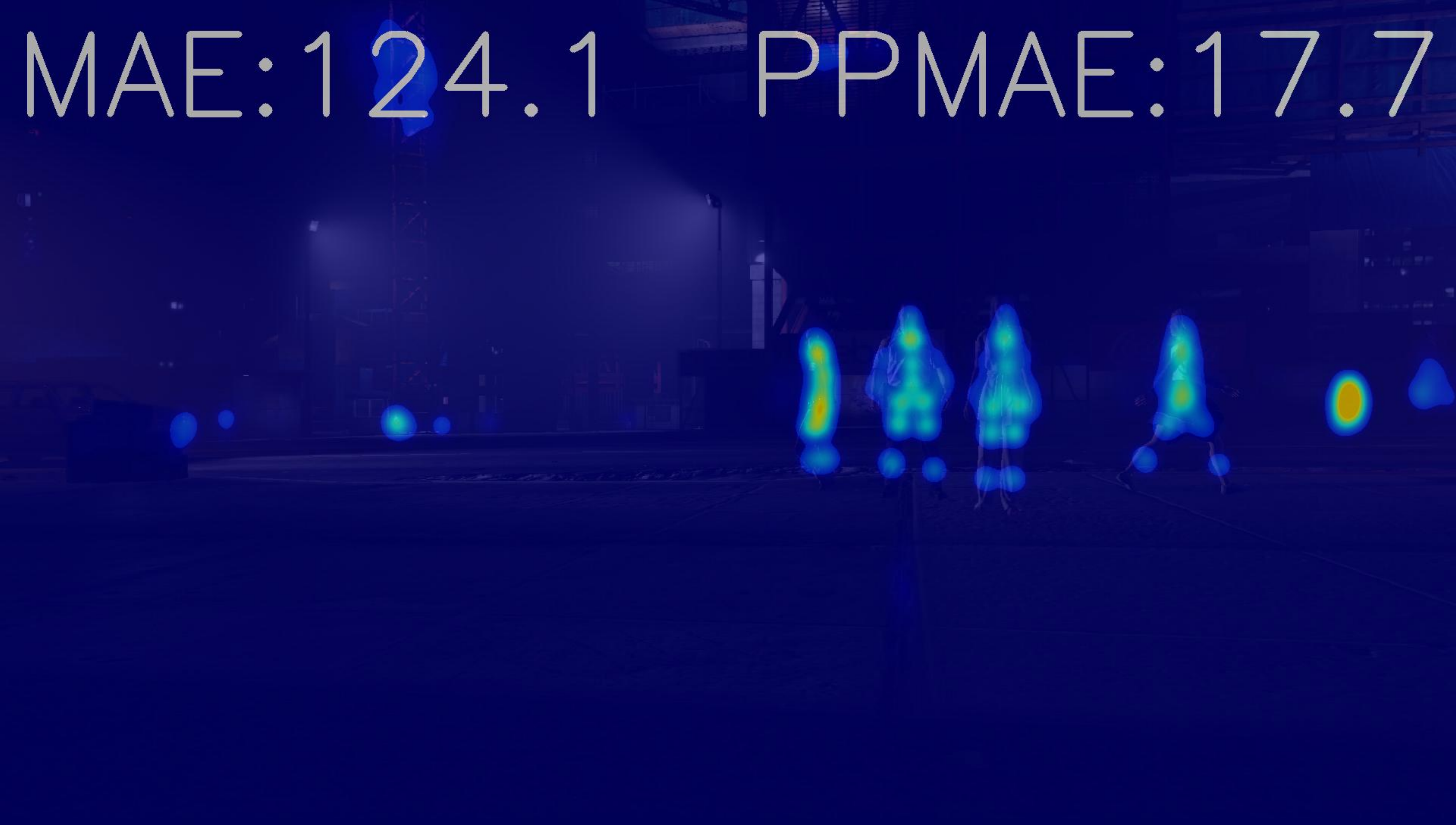}
        \end{subfigure}
        \caption{} 
        \label{subfig:e}
    \end{subfigure}

\caption{Qualitative evaluation including images corresponding to the points in the scatter plot in Fig.~2 of the main paper for which MAE and PP-MAE deviate from the primary trend.}\label{fig:sup_mae_vs_ppmae}
\end{figure*}

In Fig.~\title{}\ref{fig:sup_mae_vs_ppmae}, we present qualitative results illustrating that the alignment between these two metrics significantly deteriorates under rainy and dark environmental conditions. Specifically, in the first scenario, image distortion caused by raindrops leads STEERER-V to incorrectly infer volumes at a distance (as shown in Fig.\title{}~\ref{subfig:a}), while the glare from lightnings complicates the model's ability to detect and assess human figures (refer to Fig.\title{}~\ref{subfig:b}). In the context of darkness, the model can confuse environmental objects with humans, such as mistakenly identifying a tree situated between two individuals as a person (illustrated in Fig.\title{}~\ref{subfig:c}), along with other inaccuracies demonstrated in Fig.\title{}~\ref{subfig:d} and Fig.\title{}~\ref{subfig:e}.

\section{Other tasks with \ourdataset}\label{sec:other_tasks}

\begin{table}[ht!]

\centering
\resizebox{0.9\linewidth}{!}{%
\begin{tabular}{l|ccc}
\toprule
\textbf{Crowd Counting} & \textbf{MAE} & \textbf{RMSE} & \\
\midrule
Bayesian+~\cite{ma2019bayesian} & 3.50 & 5.87 \\
P2P~\cite{song2021rethinking} & 8.38 & 11.7 \\
MAN~\cite{lin2022boosting} & 3.54 & 5.88 \\
STEERER~\cite{hani2023steerer} & 5.56 & 6.94 \\

\specialrule{1.2pt}{2pt}{2pt}

\textbf{HMR} & \textbf{MPJPE} & \textbf{PA-MPJPE} & \textbf{PVE} \\
\midrule
CLIFF~\cite{li2022cliff} & 807.6 & 165.9 & 940.8 \\
BEDLAM-CLIFF~\cite{black2023bedlam} & 794.5 & 165.7 & 991.0\\
ReFit~\cite{wang2023refit} & 397.2 & 310.2 & 416.3 \\
\bottomrule
\end{tabular}}
\vspace{0.5em}
\caption{Results of Crowd Counting and Human Mesh Recovery on ANTHROPOS-V. MPJPE, PA-MPJPE, and PVE are measured in millimeters.}\label{tab:other_tasks}
\vspace{-10pt}
\end{table}

We evidence that \ourdataset\ can further serve as a benchmark for Crowd Counting and Human Mesh Recovery (HMR), as we report in Table~\ref{tab:other_tasks}.

The low performance of HMR methods stems from the increased complexity in the lighting and weather conditions, the number of individuals in the scene, and the large number of occlusions that invalidate the person-detection step leading to inaccurate predictions. CLIFF and BEDLAM-CLIFF particularly struggle to estimate the global scale and orientation, with an MPJPE value of $807.6$ mm and $794.5$ mm, respectively; the error on the prediction drastically reduces to $~165$ mm after the Procrustes alignment.
For what concerns Crowd Counting, Bayesian+ exhibits the best performance, yielding an average error of $3.5$ individuals per frame and surpassing more recent methods such as STEERER.
\begin{figure*}[!htbp]
    \centering
    \begin{subfigure}{\linewidth}
    \centering
        \begin{subfigure}{0.37\linewidth}
            \includegraphics[width=\linewidth]{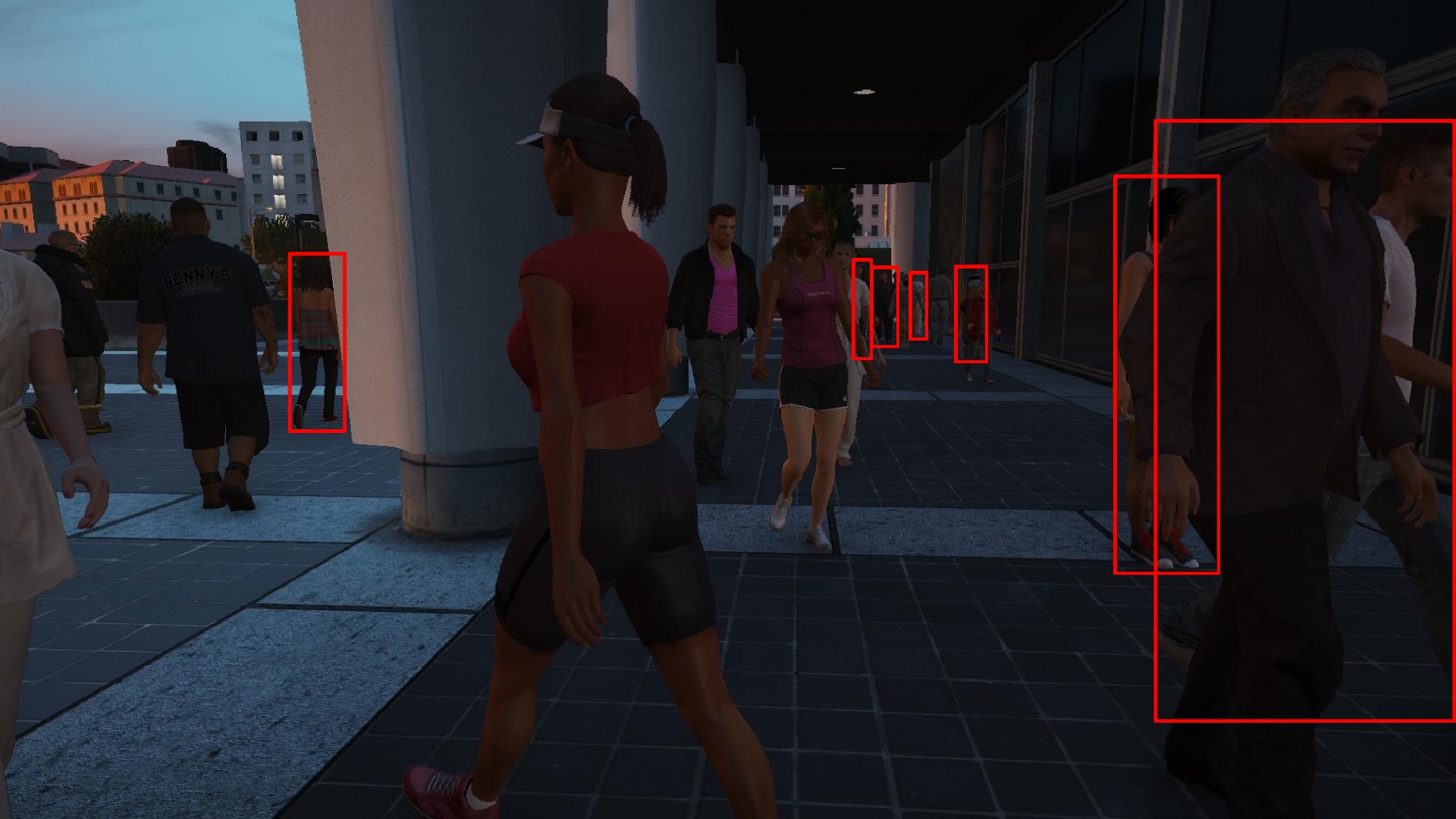}
        \end{subfigure}\hspace{.5em}
        \begin{subfigure}{0.37\linewidth}
            \includegraphics[width=\linewidth]{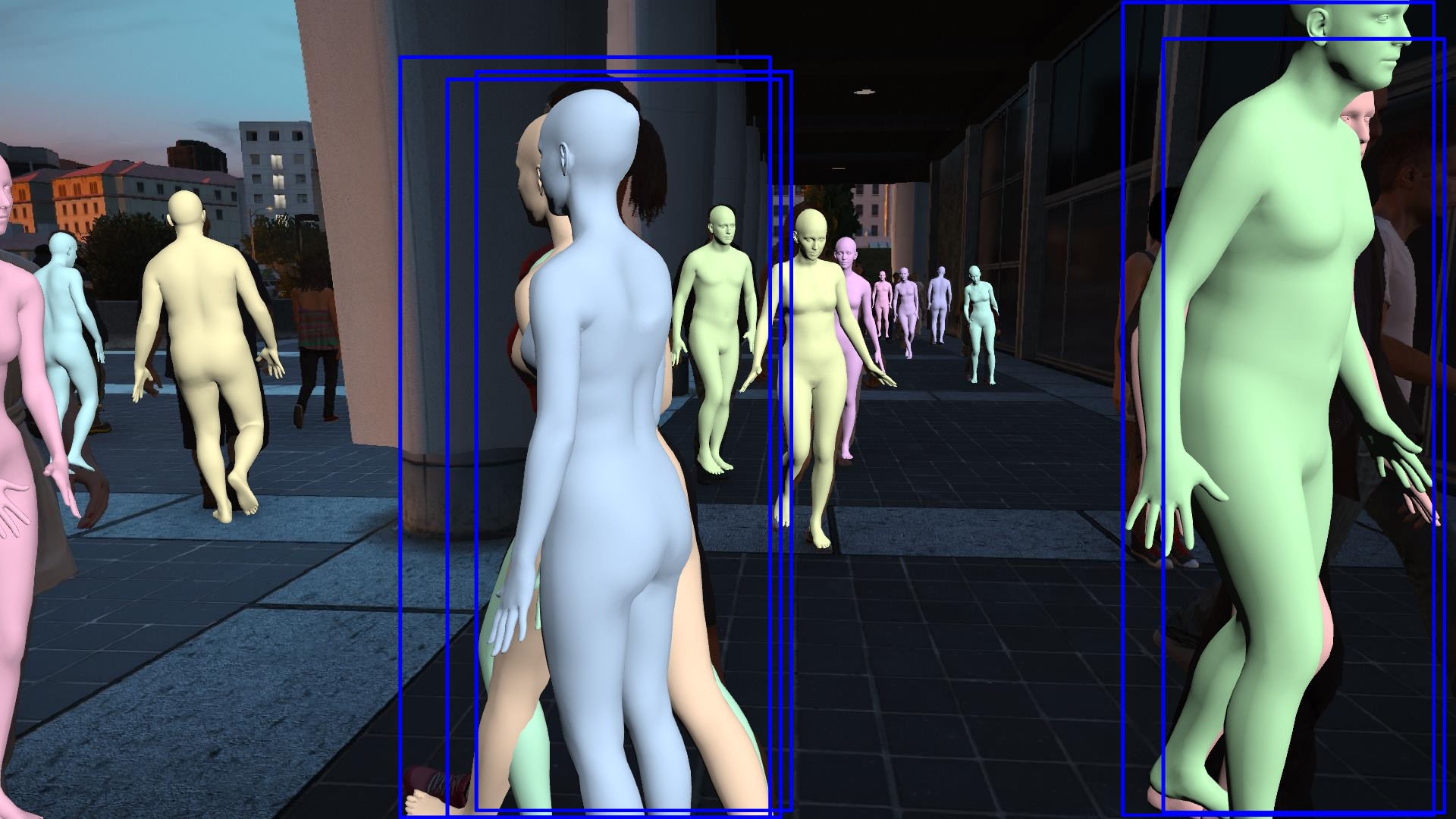}
        \end{subfigure}
        \caption{}\label{subfig:hmr_result_day}
    \end{subfigure}
    \begin{subfigure}{\linewidth}
    \vspace{0.5em}
    \centering
        \begin{subfigure}{0.37\linewidth}
            \includegraphics[width=\linewidth]{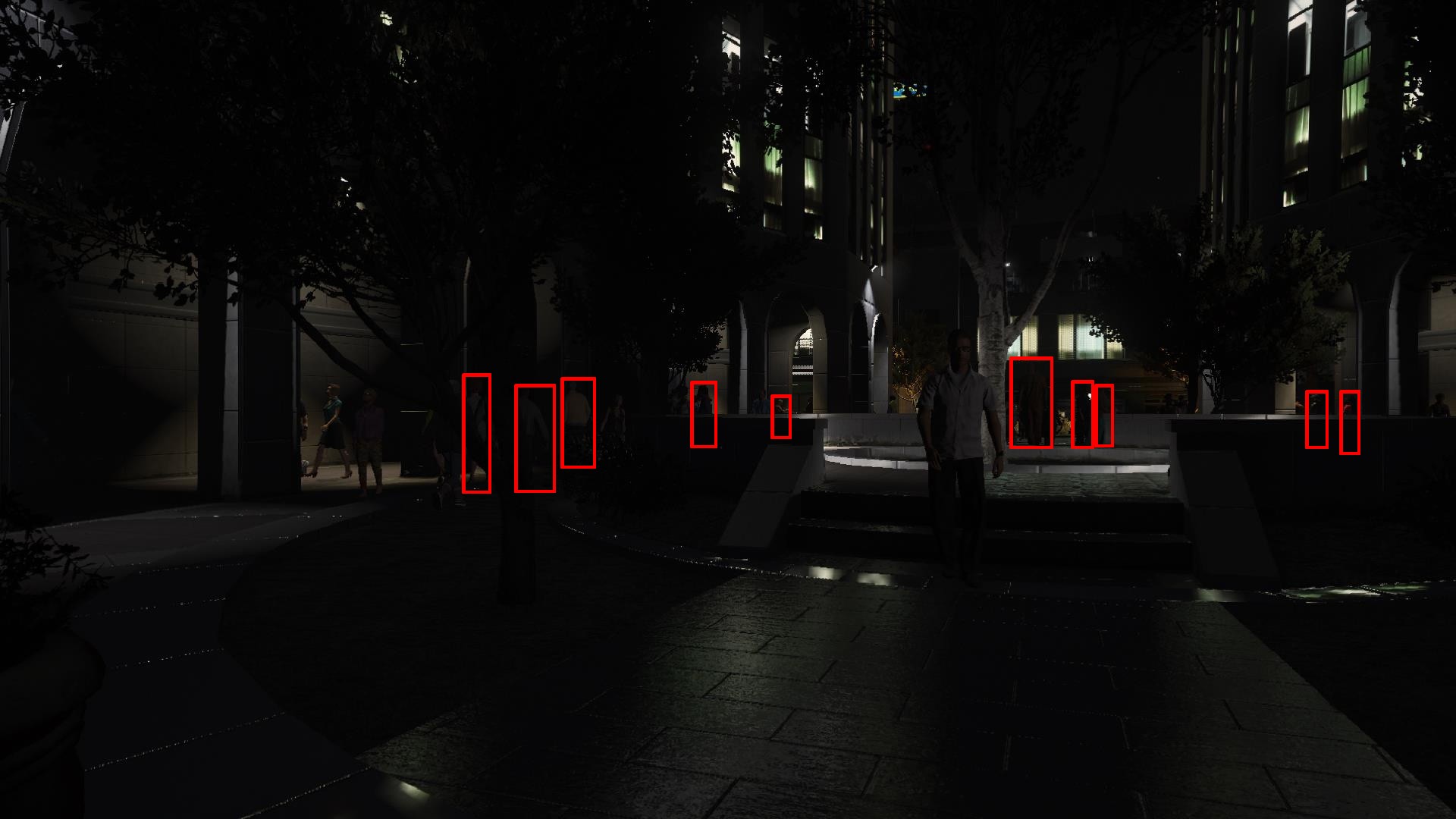}
        \end{subfigure}\hspace{.5em}
        \begin{subfigure}{0.37\linewidth}
            \includegraphics[width=\linewidth]{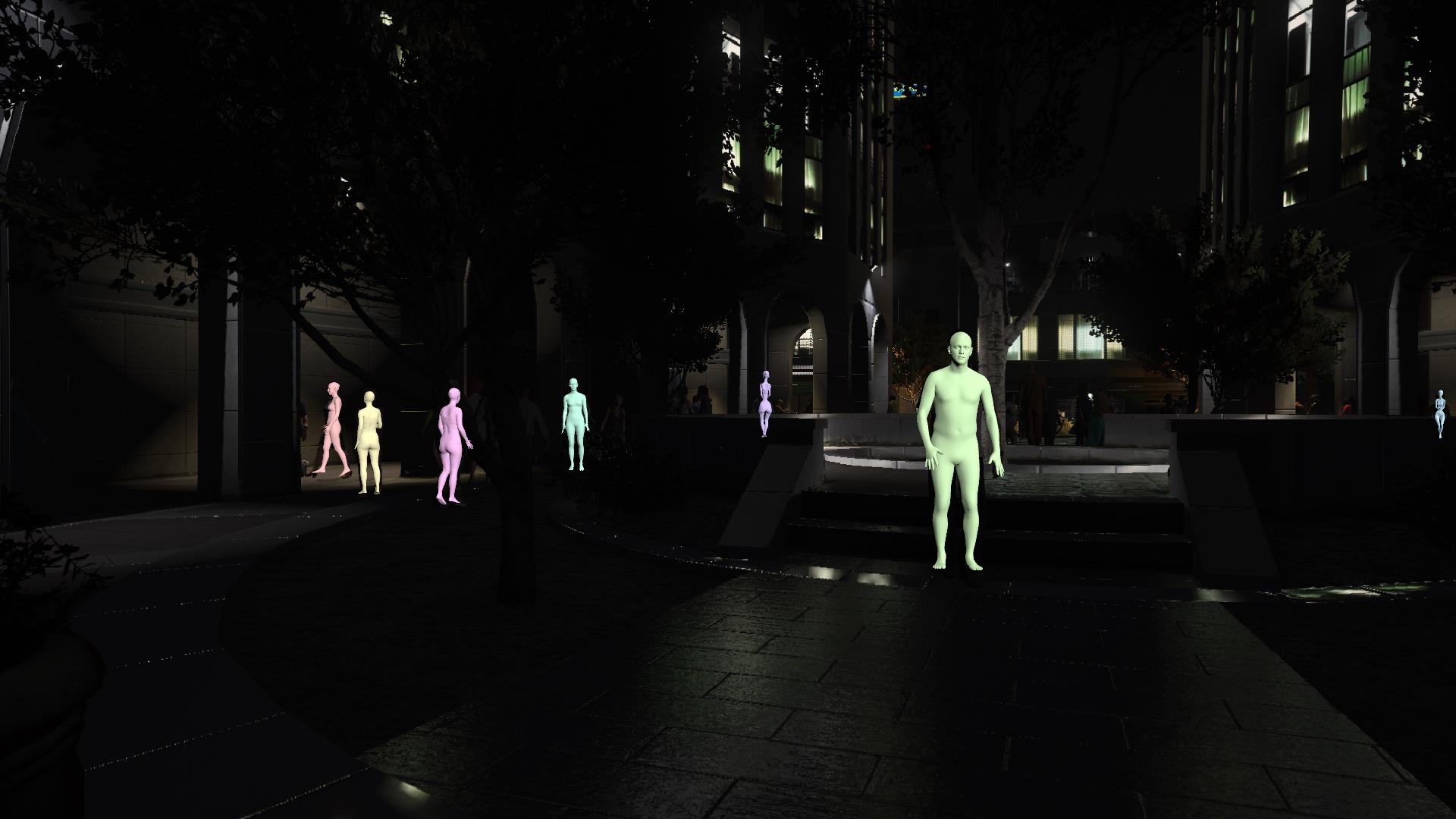}
        \end{subfigure}
        \caption{}\label{subfig:hmr_result_night}
    \end{subfigure}
    \caption{Qualitative results of BEDLAM-CLIFF~\cite{black2023bedlam} on \ourdataset{} when provided with the predicted bounding boxes of YOLOv7~\cite{wang2023yolov7} (we omit some of them for clarity). We highlight in red some of the instances that have not been detected and in blue those that have been detected multiple times.}\label{fig:hmr_results}
\end{figure*}

\section{Further notes on the baselines}
\label{sec:baselines}

In this section, we add some notes about the results of the HD+HMR baselines (Sec.~\ref{subsec:hmr_cve_notes}), and the architectural adaptation of the Crowd Counting models, which we modify for the CVE task (Sec.~\ref{subsec:arch_adapt_cc_models}).

\subsection{About the low performance of HD+HMR baselines for CVE}\label{subsec:hmr_cve_notes}

Here we focus on BEDLAM-CLIFF~\cite{black2023bedlam}.

As evidenced in Fig.~\ref{subfig:hmr_result_day}, the performance of the human detection model has a critical impact on the overall CVE results of the HD+HMR models. One of the failure cases originates from either a missing detection, as for the woman on the left of the pillar, or multiple predicted bounding boxes of the same instance, as for the subjects in the foreground. Also, even when the error deriving from the human detection step is driven to zero, like in the oracular experiment described in Sec.~5.1 of the main paper, BEDLAM-CLIFF underperforms when compared with STEERER-V. Indeed, occlusions, color contrast, and extreme light conditions harm the body shape regression, which in turn increases the volume estimation error, as Fig.~\ref{subfig:hmr_result_night} empirically confirms; for example, the woman on the left of the image is assigned with an excessively skinny SMPL mesh, and the people partially occluded by the central round terrace are approximated with an amorphous mesh.

Finally, the HD+HMR baselines yield more parameters than the baselines adapted from Crowd Counting, as shown in Table~\ref{tab:imp_details_baselines}.

\begin{table}[!htbp]
\centering
\resizebox{0.7\linewidth}{!}{%
\begin{tabular}{l|cc}
\toprule
\textbf{Model}     & \textbf{\#Params} & \textbf{LR} \\ \midrule
\textcolor{gray}{YOLOv7} & \textcolor{gray}{165M} & \textcolor{gray}{$1 \times 10^{-5}$}\\
\midrule
\midrule
CLIFF & 247M & $5 \times 10^{-5}$\\
BEDLAM-CLIFF & 247M & $5 \times 10^{-5}$\\
ReFit & 240M & $1 \times 10^{-4}$\\
\midrule
MAN       & 40.4M &  $1\times10^{-5}$ \\ 
Bayesian+ & 21.5M & $1\times10^{-5}$ \\ 
P2P-net   & 21.6M & $1\times10^{-5}$ \\ 
STEERER   & 64.6M & $5\times10^{-7}$ \\ \bottomrule
\end{tabular}}
\vspace{-0.7em}
\caption{Details on the baselines employed for CVE. The number of parameters of the HD+HMR baselines includes the one of YOLOv7~\cite{wang2023yolov7}, for which we report the parameters in gray on top of the table.}
\label{tab:imp_details_baselines}
\end{table}

\begin{figure*}[!htbp]
    \centering
    \includegraphics[width=0.9\textwidth]{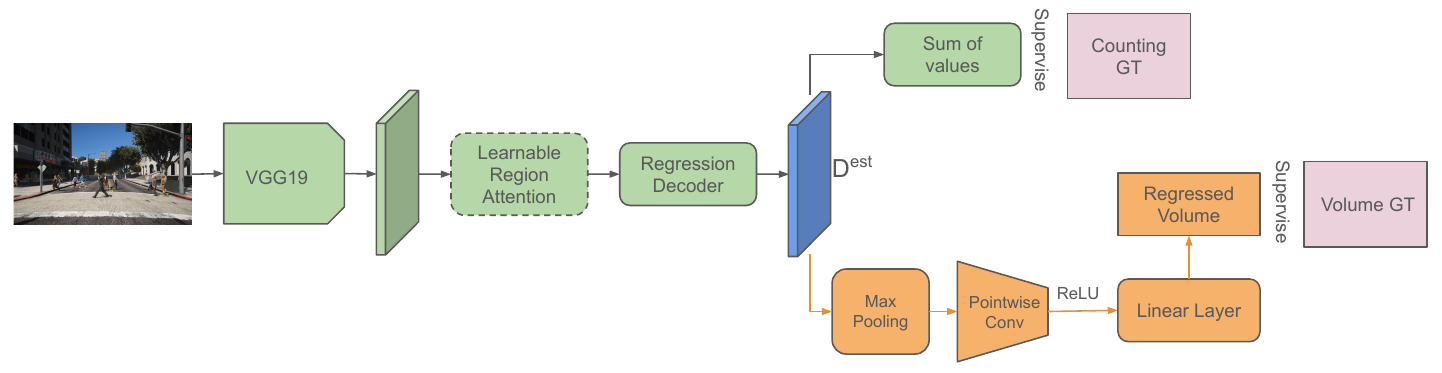}
    \vspace{-0.5em}
    \caption{Bayesian+ and MAN modified architectures. The Transformer Encoder blocks with \textit{Learnable Region Attention} are used only by MAN. The green layers are kept as they are in the original architectures. Orange layers are part of the additional branch tailored on the CVE task. The estimated density map ($D^{est}$) is reduced with summation to a single value and counting losses are calculated on it. $D^{est}$ is also used as input for our additional volume-related branch to regress the volume occupied in a frame. Finally, we calculate the L1 loss between regressed and ground truth volumes.}
    \label{fig:man_bayesian_modif}
\end{figure*}

\subsection{Details on the architectural adaptation of Crowd Counting baselines}\label{subsec:arch_adapt_cc_models}

We train all models on a single NVIDIA A100 GPU until convergence. Both the original codebases and the edited code leverage the PyTorch framework.

Table~\ref{tab:imp_details_baselines} provides further implementation details. 

\paragraph{Bayesian+ and MAN:}

These models have nearly the same base architecture, so we modify them in the same way. These architectures are described by the green blocks in Fig.~\ref{fig:man_bayesian_modif}, with only MAN employing the Transformer Encoder with the Learnable Region Attention block.
Bayesian+ and MAN are both Crowd Counting architectures. Hence, to adapt them to the CVE task, we define an additional branch besides the one performing counting. 
The orange blocks in Fig.~\ref{fig:man_bayesian_modif} illustrate the novel branch. Since these models are trained on $512\times512$ image crops, Max Pooling is employed for computing volume on larger-sized images, while Pointwise Convolution compresses tensors to a single dimension.\newline
Furthermore, we alter the pre-processing pipeline to compute both counting and volume-related ground truths on which both models are supervised, i.e., the total number of persons in the frame and the total volume occupied by them, respectively.\newline
Notice that the counting branch is necessary because we use its output, which is the estimated density map, as input for our additional volume regressive branch.
For both these Bayesian-based models, the loss we use for the volume branch is the L1 loss between the regressed and the ground truth volumes.
We also keep the counting losses of Bayesian+ and MAN as described in their papers.

\paragraph{P2P-Net:}

To adapt P2P-Net for the CVE task, we expand
the model's capabilities to predict $x$ and $y$ coordinates for each identified head with a $v$ label indicating the volume of the corresponding person. To encourage accurate predictions for the volume ($v$) rather than solely emphasizing $x$ and $y$ predictions, we introduce an additional loss component. This supplementary component is an L1 loss computed between predicted and ground truth volumes. To tune the influence of this loss, we apply a weighting coefficient $\lambda$, determined through experimentation to be optimal at the value of 1e-4.

\section{Qualitative Evaluation: real-world images}\label{sec:add_qualitatives}

\begin{figure*}[!htbp]
\centering
\begin{tabular}{cc}
    \includegraphics[width=.40\linewidth]{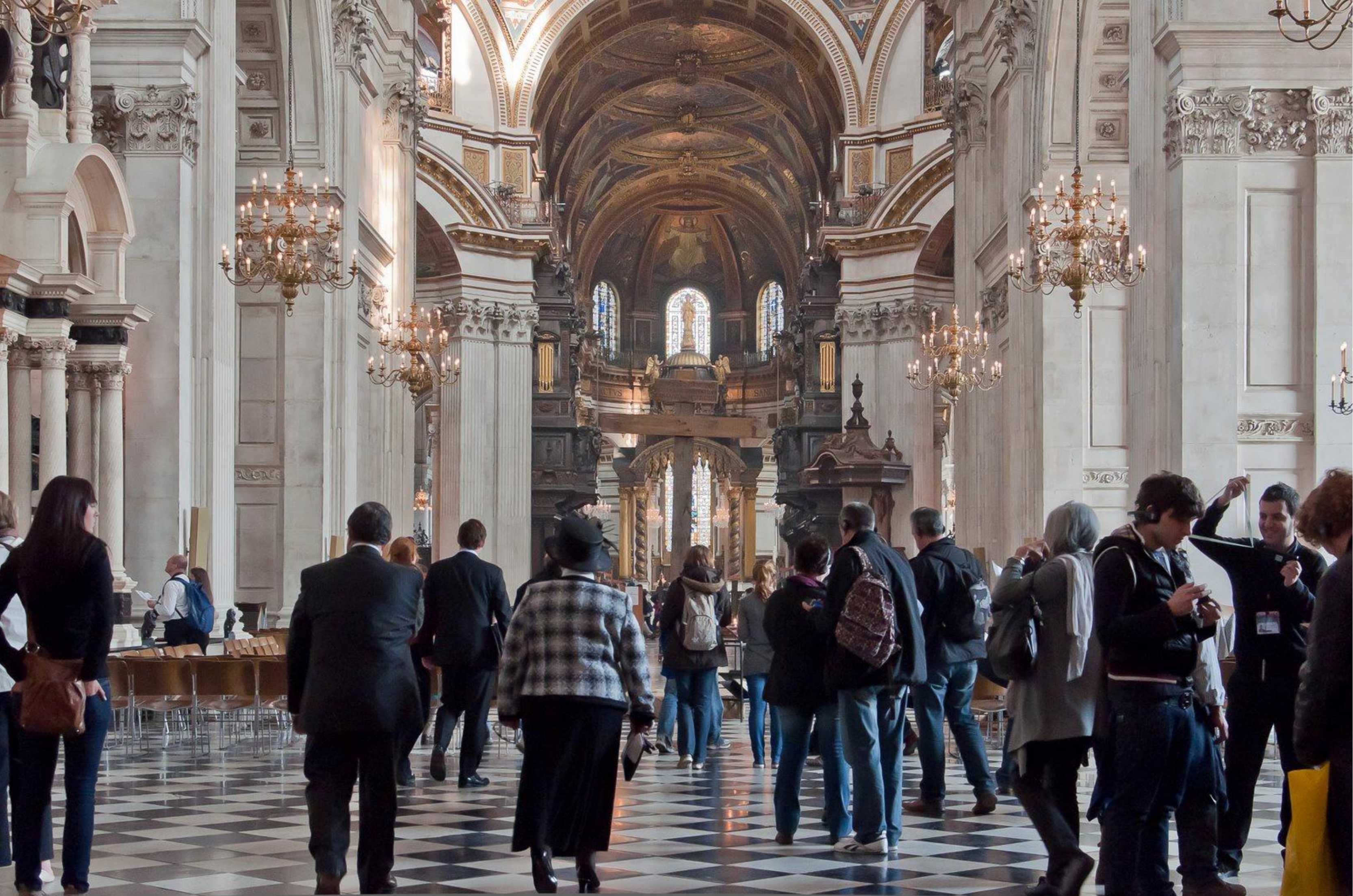} &
    \includegraphics[width=.40\linewidth]{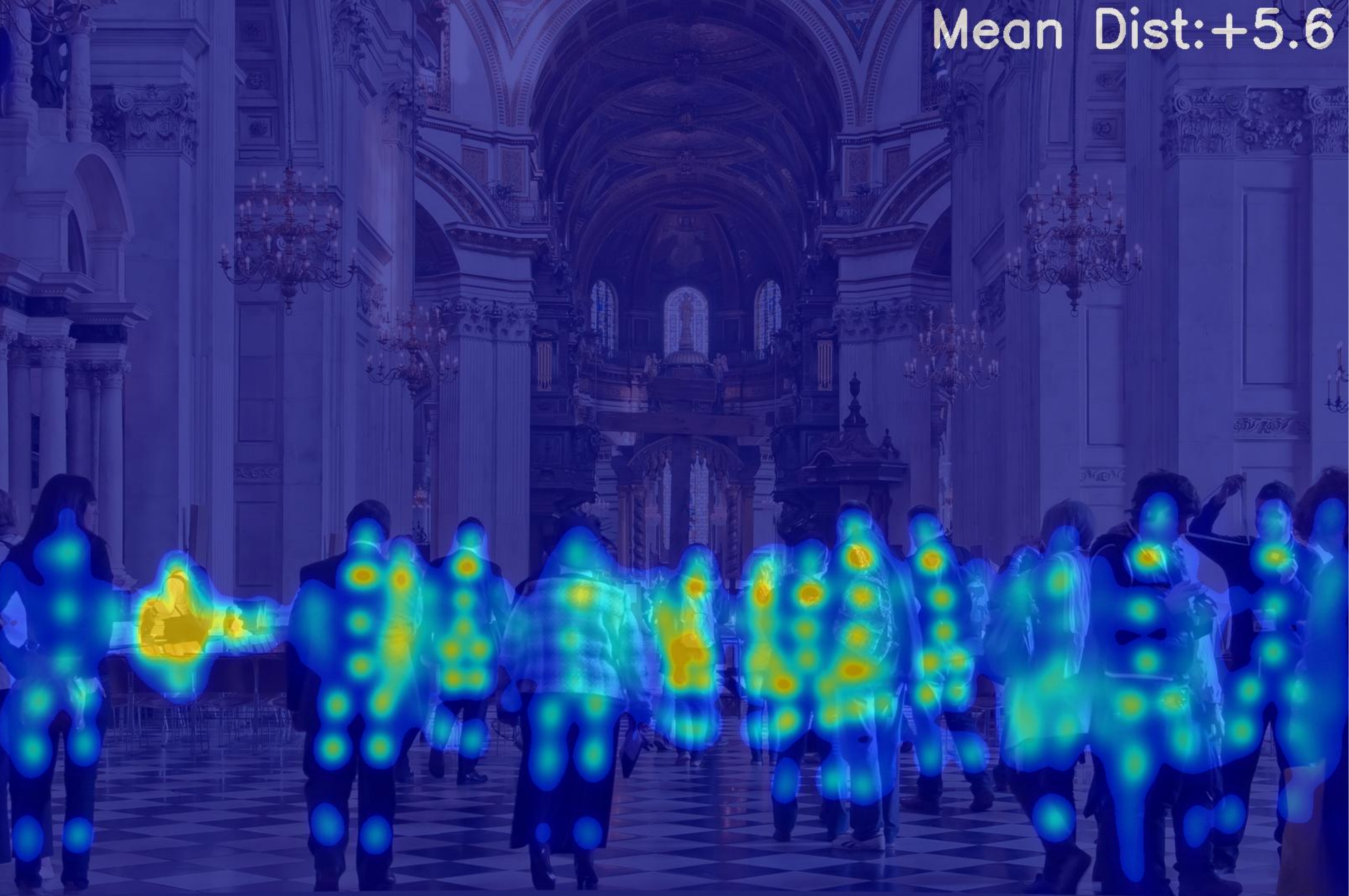} 
    
    \\

    \includegraphics[width=.40\linewidth]{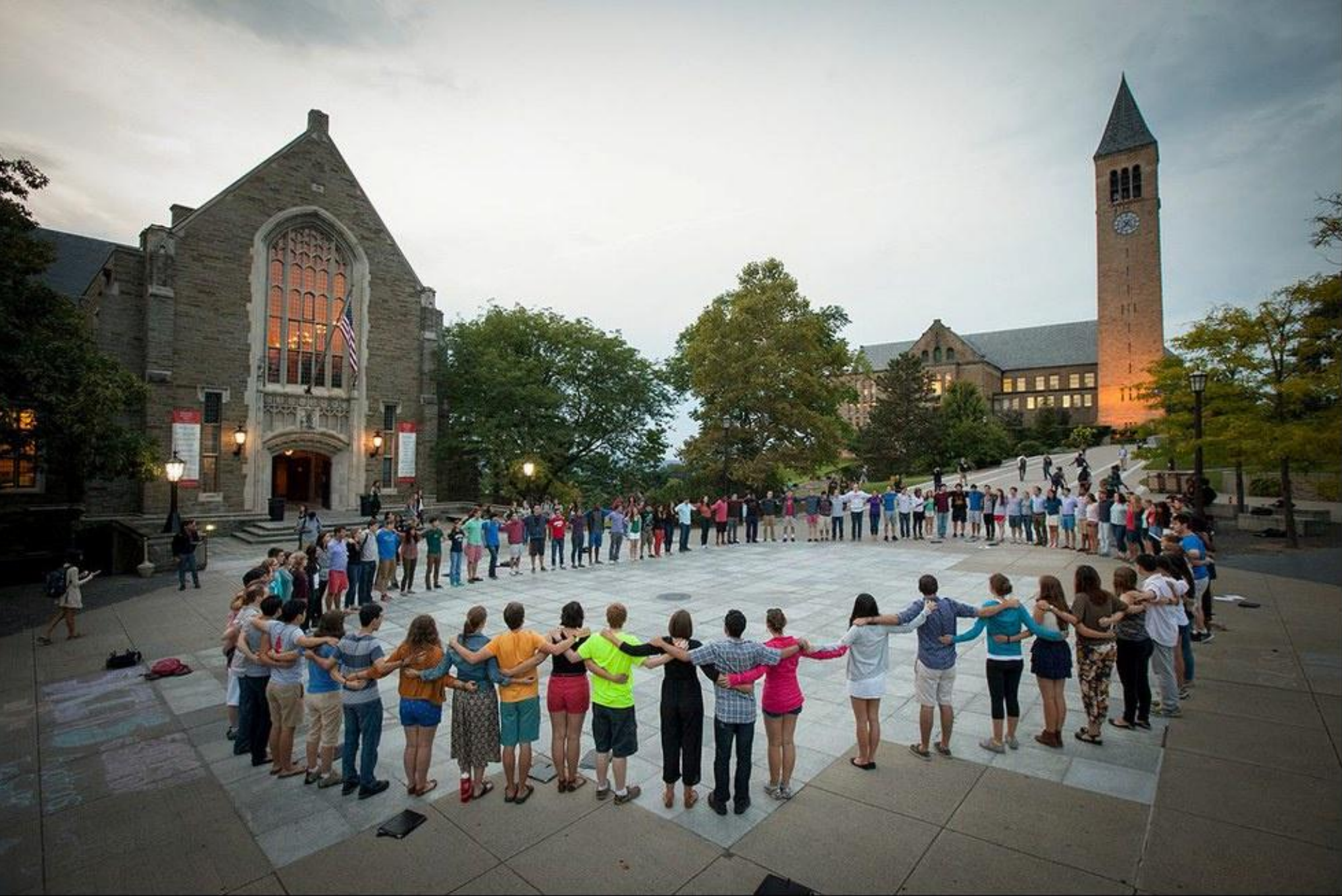} &
    \includegraphics[width=.40\linewidth]{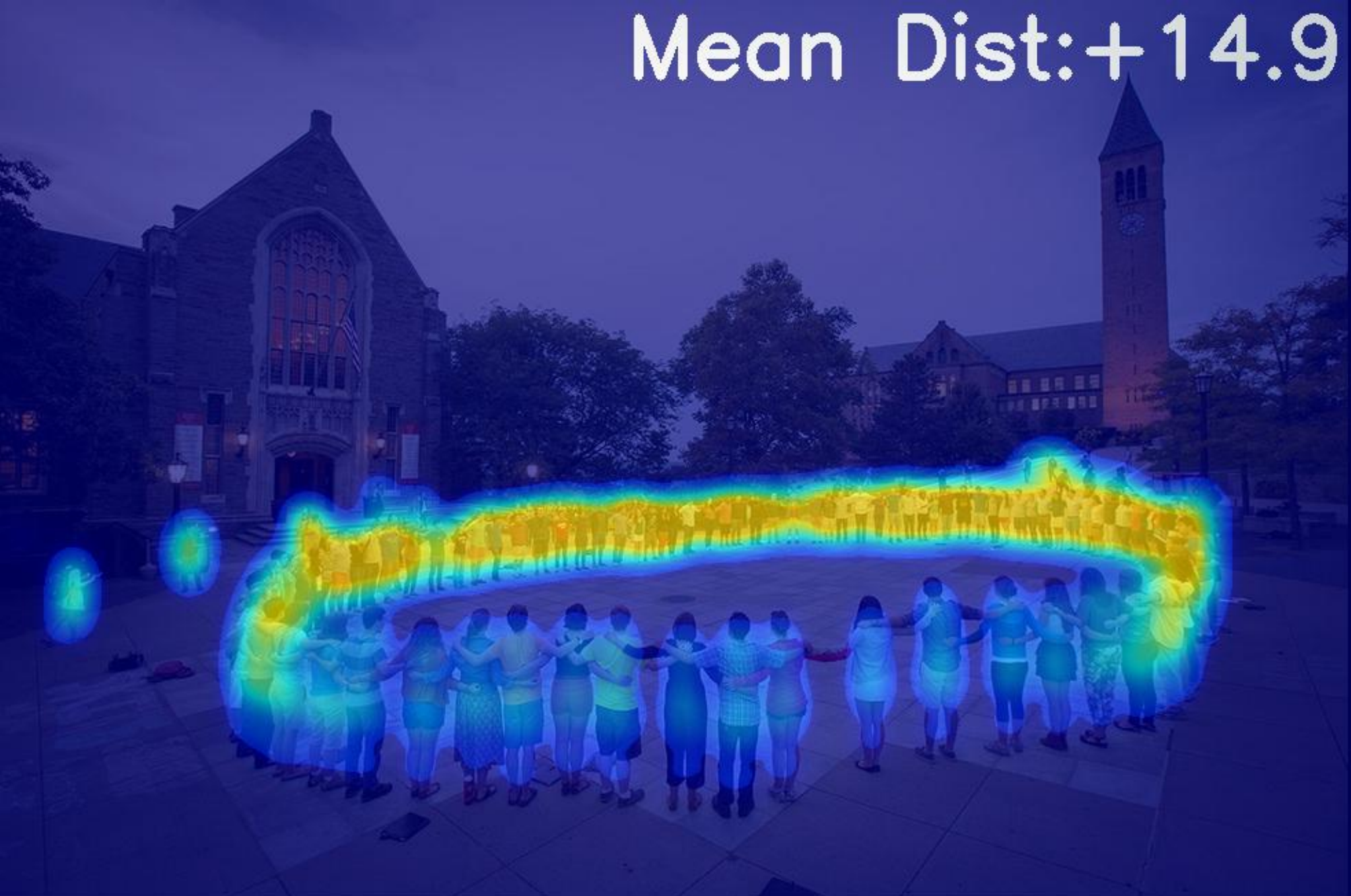}
    \\

    \includegraphics[width=.40\linewidth]{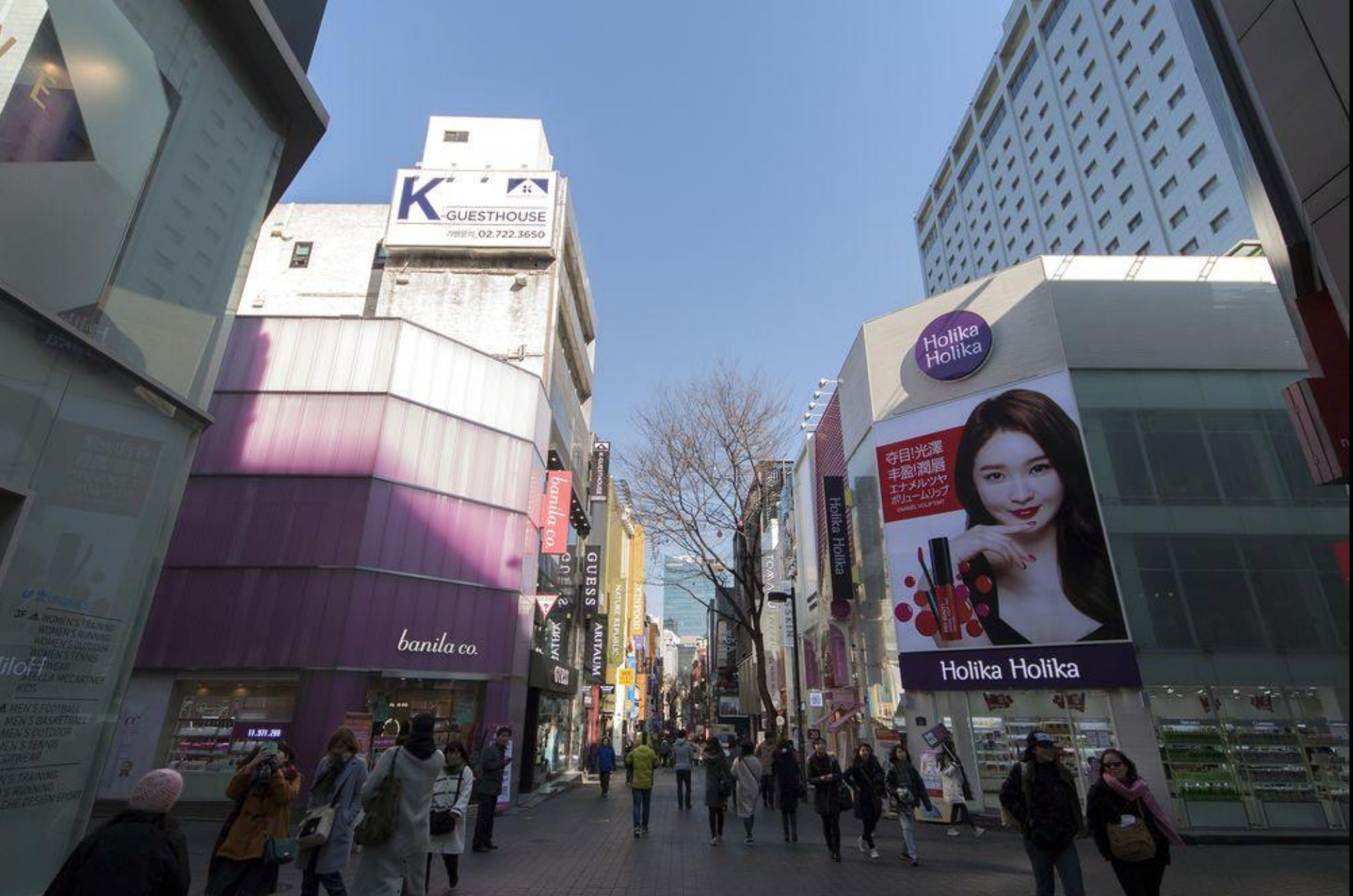} &
    \includegraphics[width=.40\linewidth]{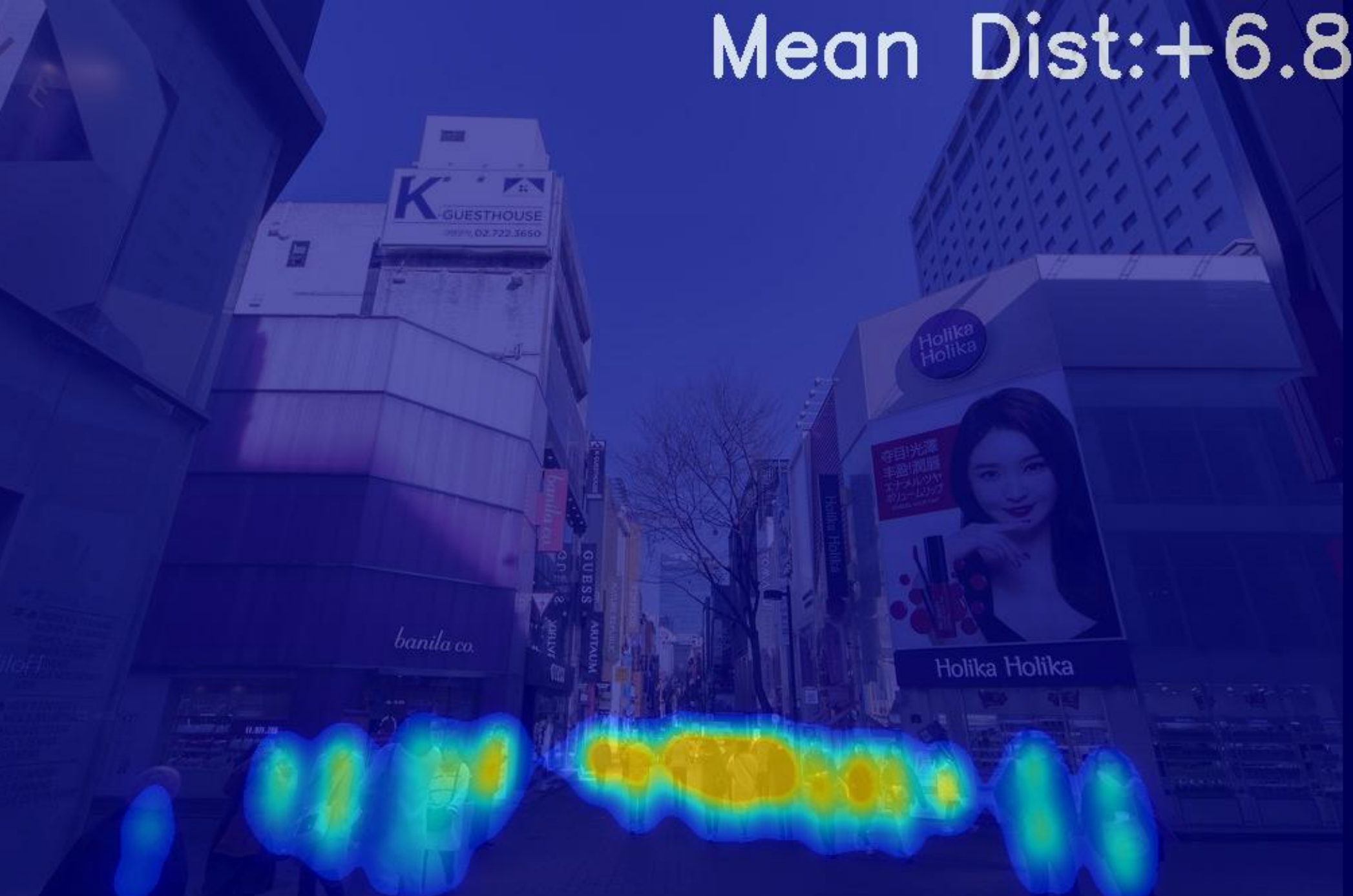} \\

\end{tabular}
  \caption{Zero-shot results of STEERER-V on CrowdHuman~\cite{shao2018crowdhuman} images when predicting the volume of adults. On each predicted volume map we superimpose the difference between the average real-world per-person volume and the predicted per-person one.}
  \label{fig:supp_qual_results_1}
\end{figure*}

\begin{figure*}[!htbp]
  \centering
  \includegraphics[width=.40\linewidth]{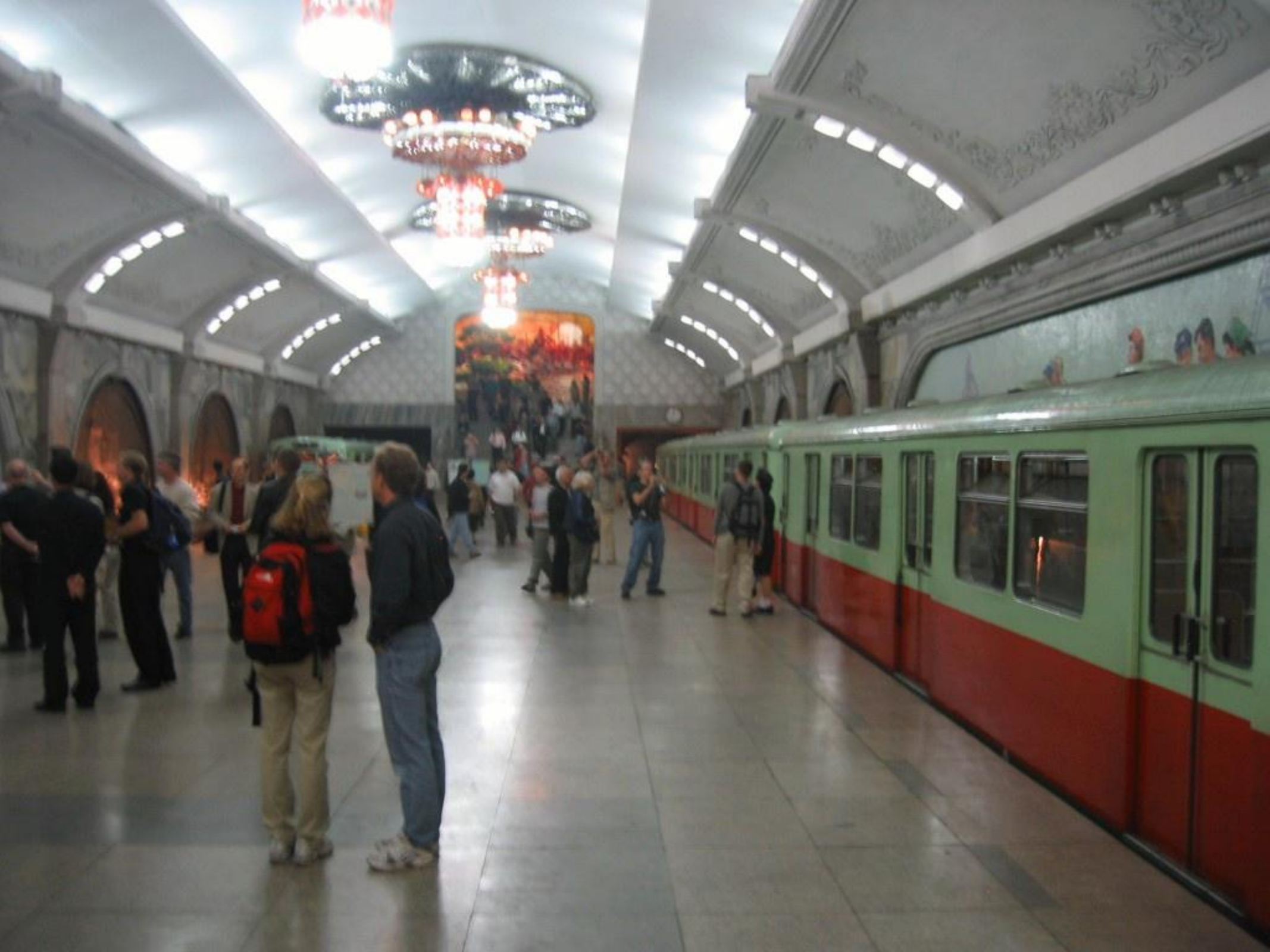}
  \includegraphics[width=.40\linewidth]{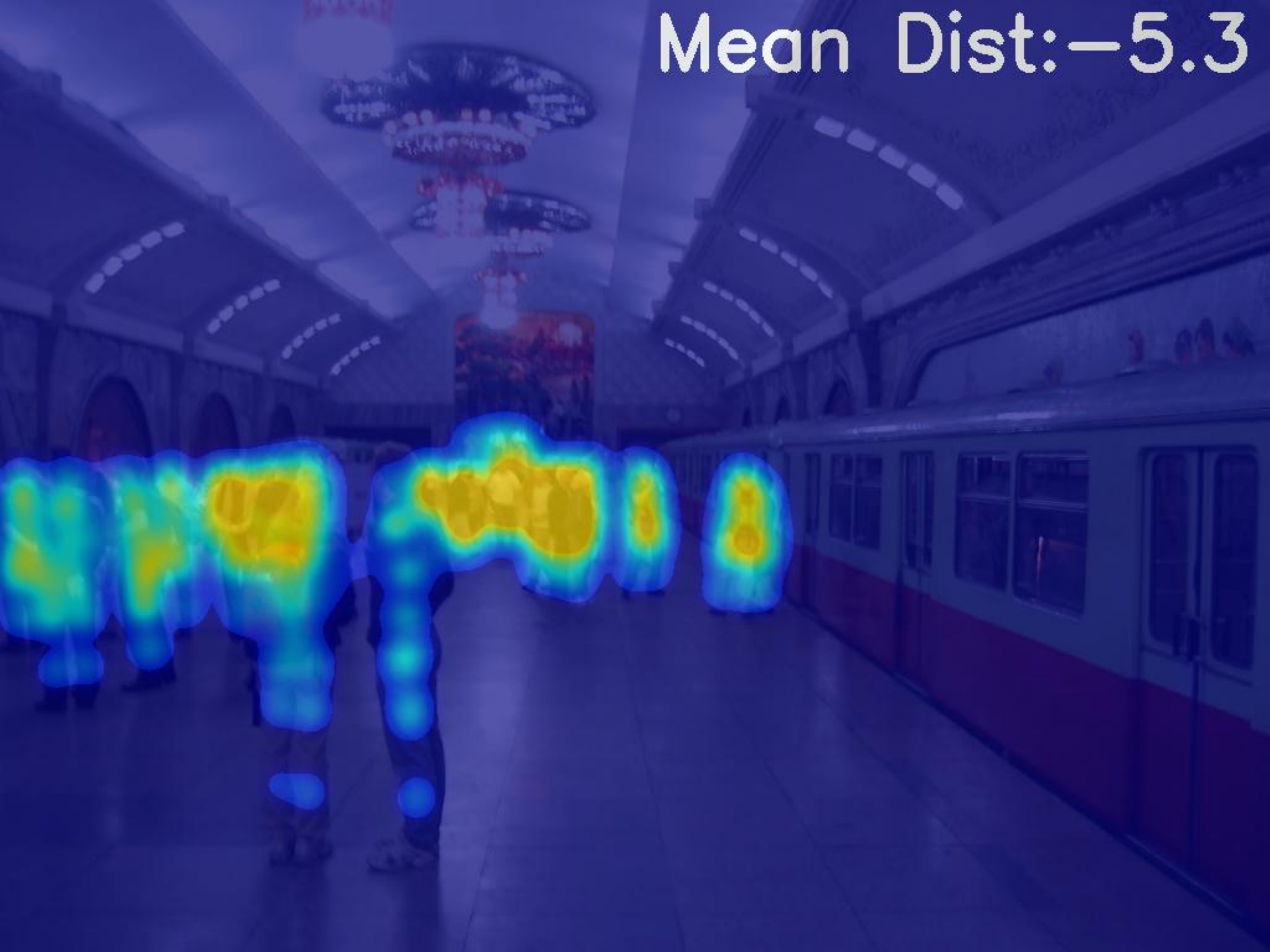} \\
  \caption{{Zero-shot result of STEERER-V on a CrowdHuman~\cite{shao2018crowdhuman} image. STEERER-V underestimates the total volume due to the low quality of the image, the domain gap, and the severe occlusions. On the predicted volume map we superimpose the difference between the average real-world per-person volume and the predicted per-person one.}}
  \label{fig:supp_qual_results_3}
\end{figure*}

\begin{figure*}[!htbp]
\centering
\begin{tabular}{cc}
\centering

    \includegraphics[width=.40\linewidth]{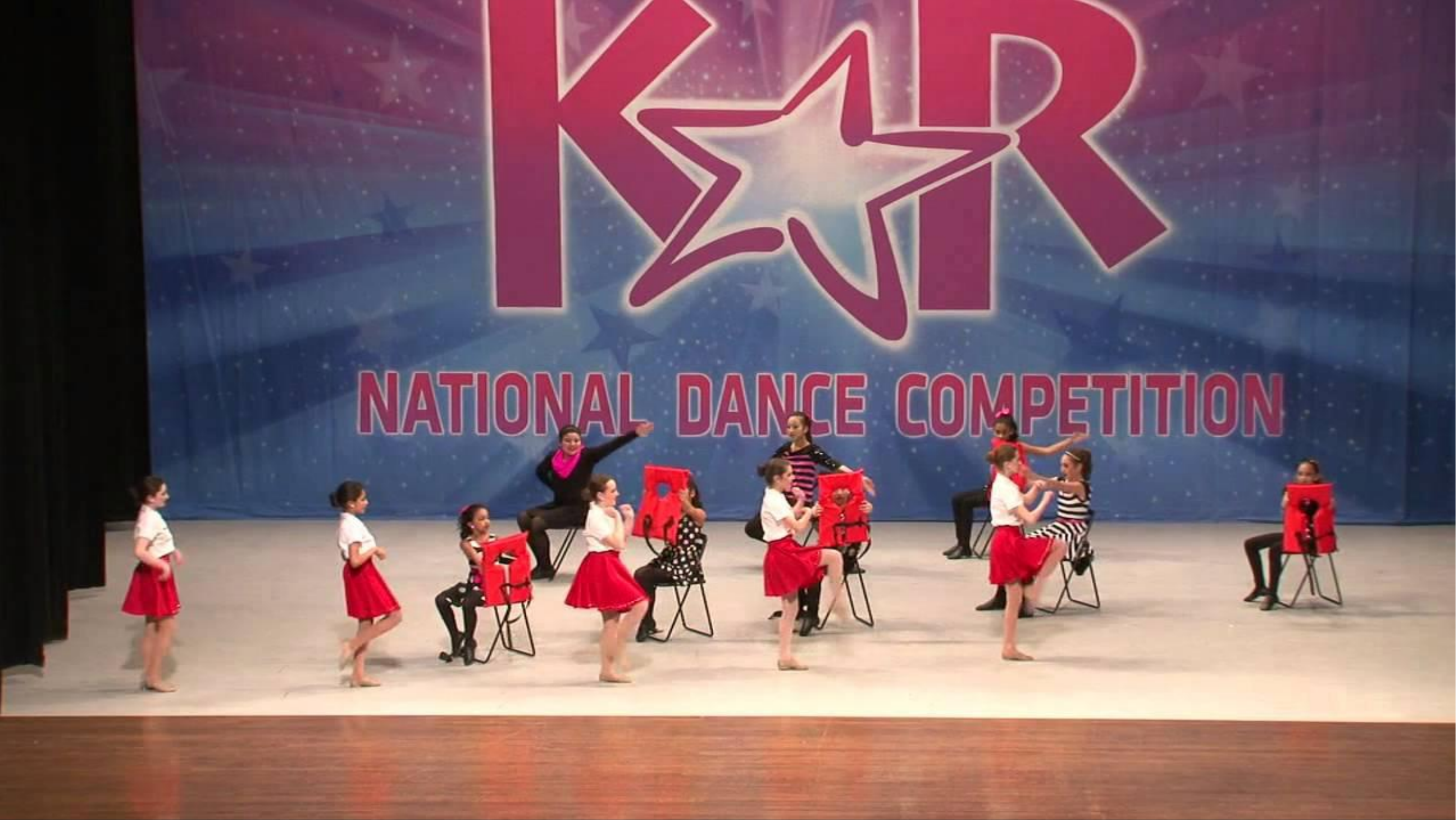} &
    \includegraphics[width=.40\linewidth]{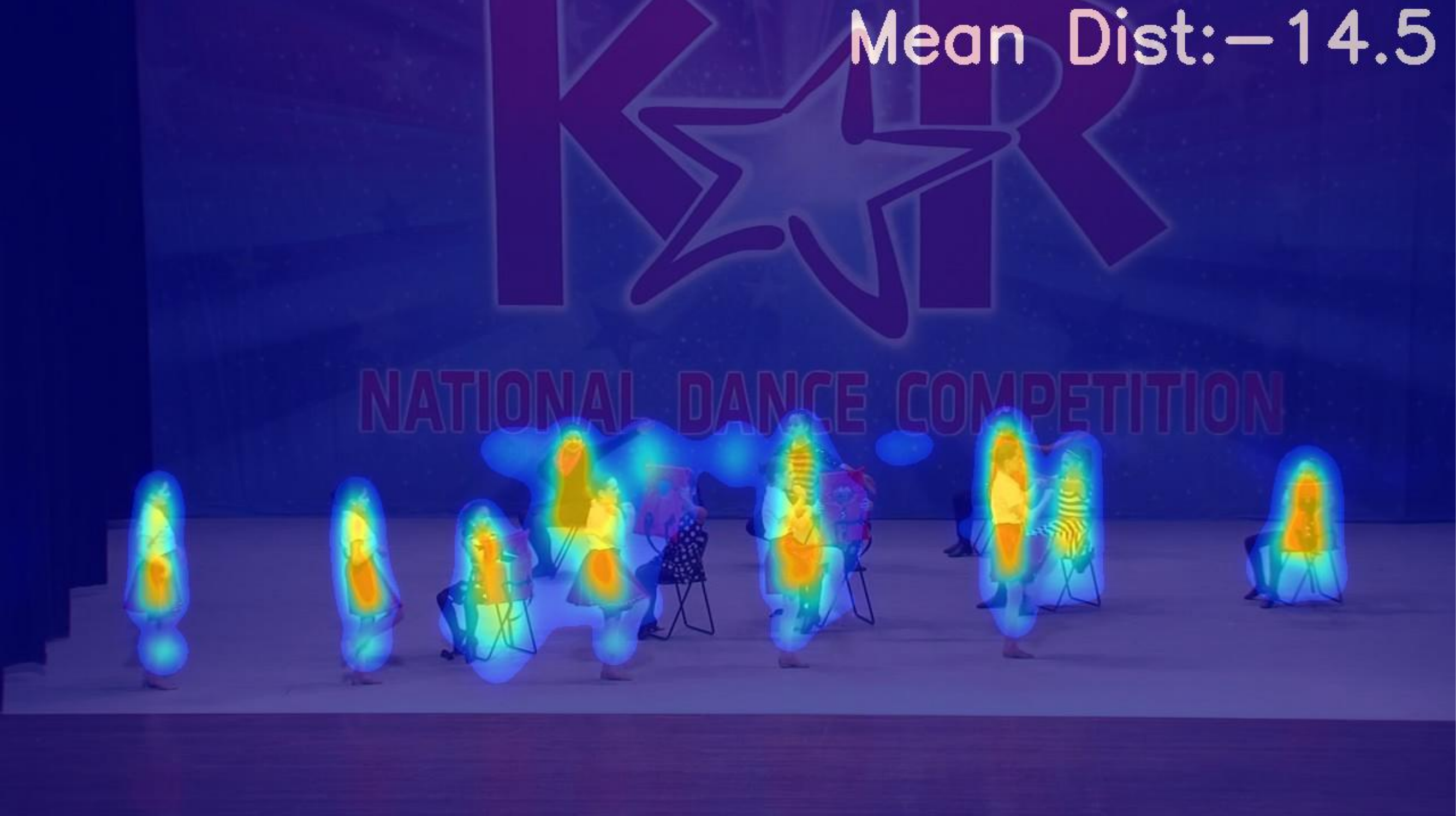}
    \\
  
    \includegraphics[width=.40\linewidth]{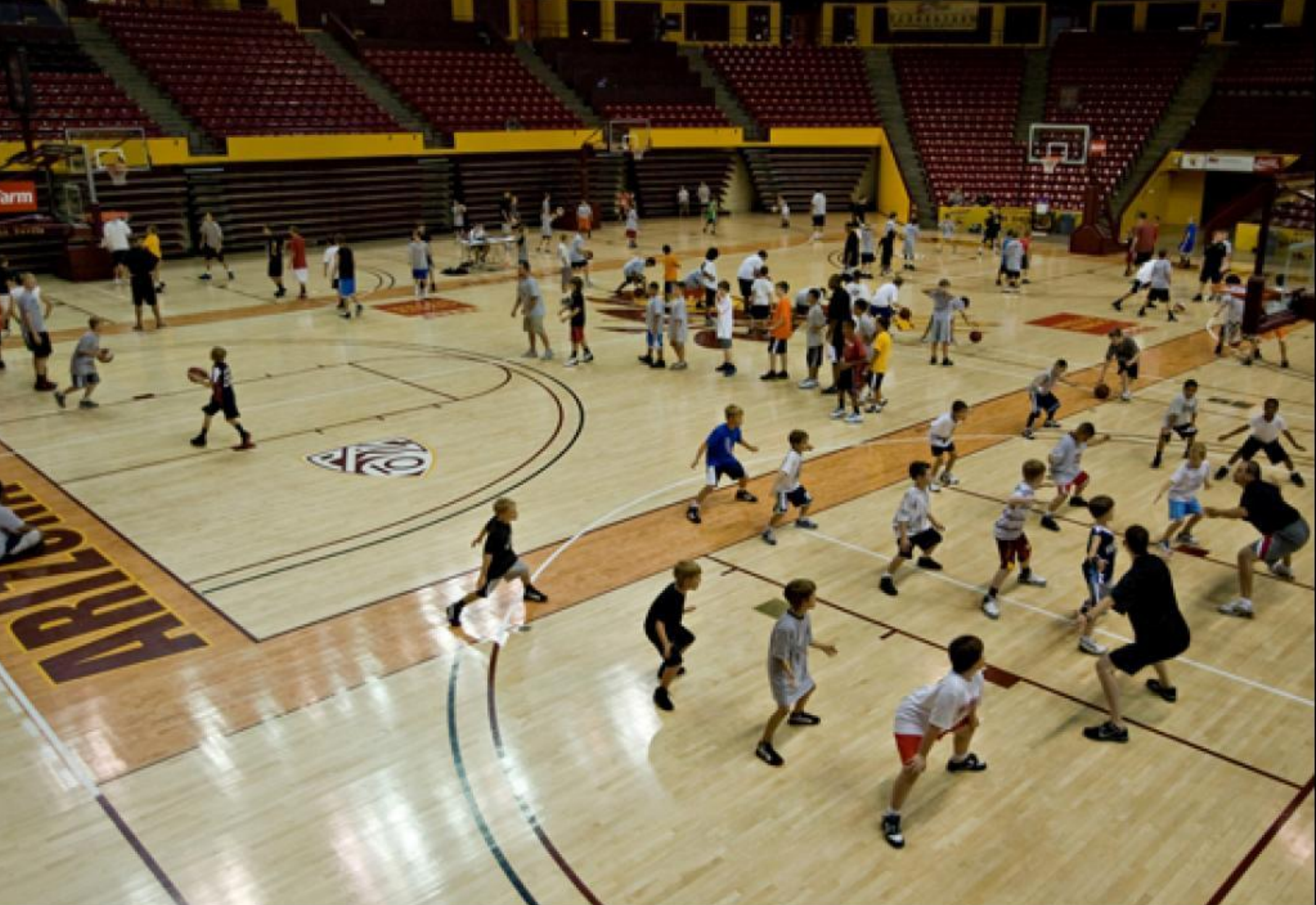} &
    \includegraphics[width=.40\linewidth]{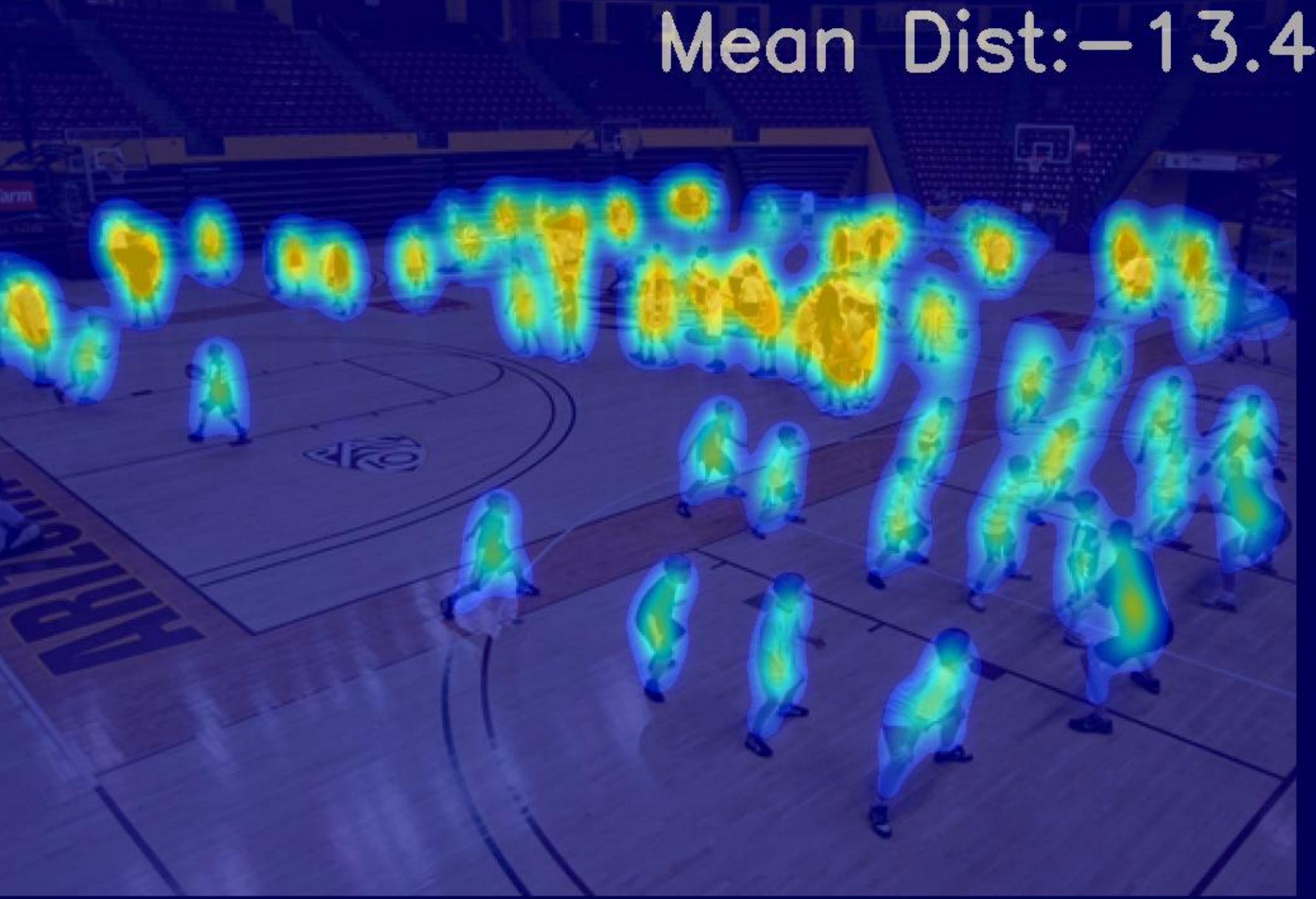} 
    \\
  
    \includegraphics[width=.40\linewidth]{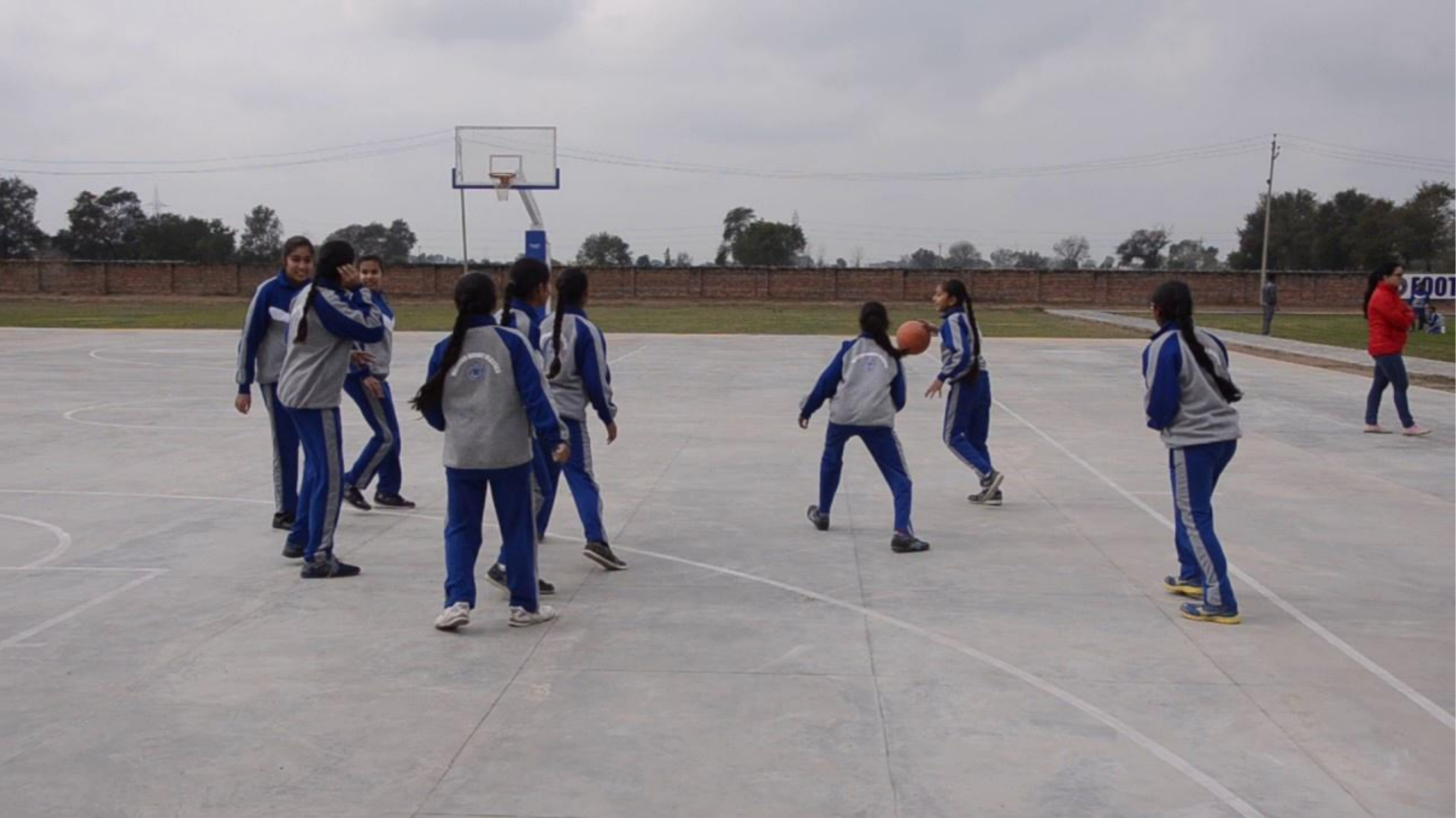} &
    \includegraphics[width=.40\linewidth]{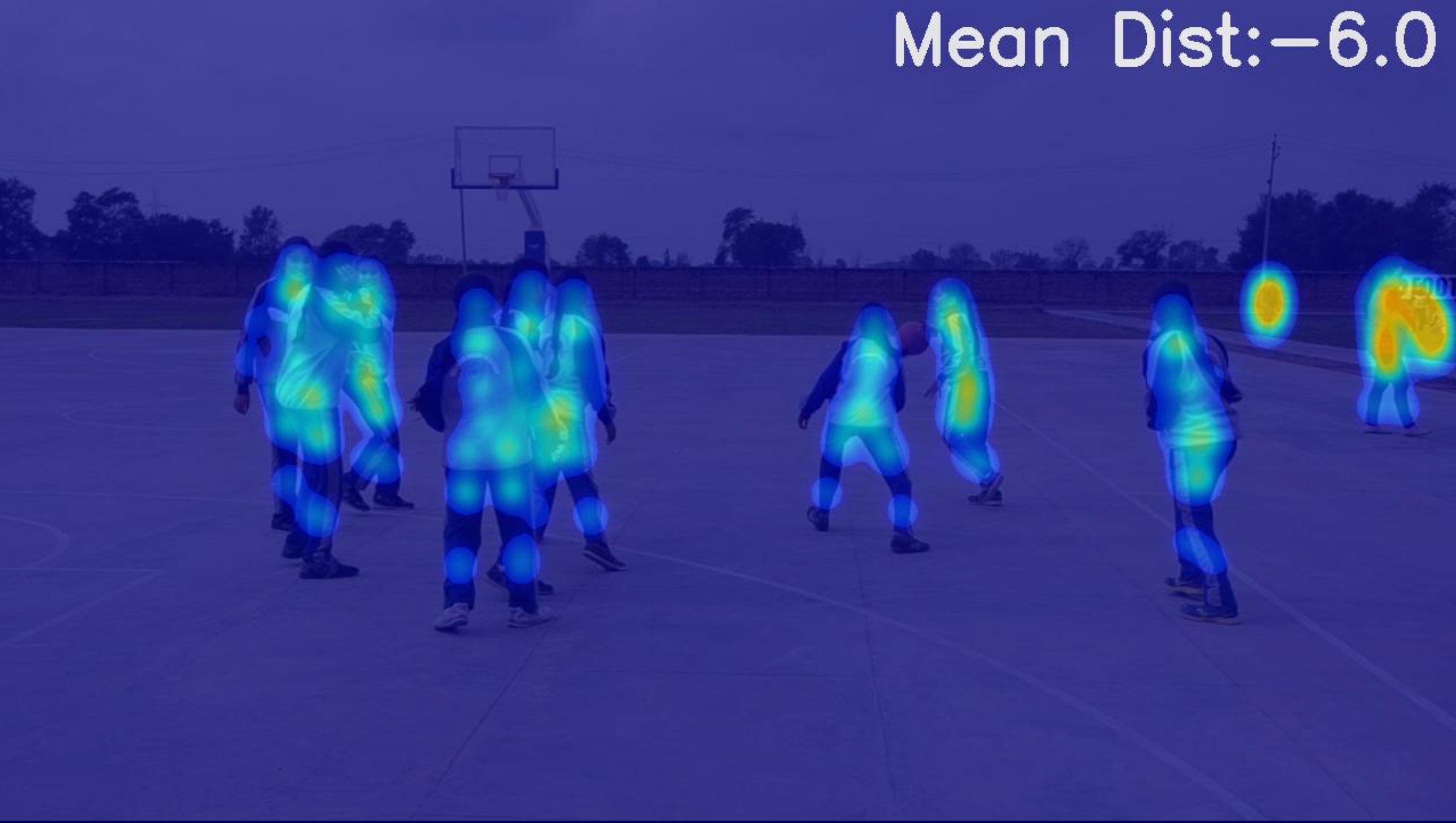}
    \\
    
\end{tabular}
  \caption{{Zero-shot results of STEERER-V on CrowdHuman~\cite{shao2018crowdhuman} images when predicting the volume of builds that have not been seen at train time, e.g., kids. On each predicted volume map we superimpose the difference between the average real-world per-person volume and the predicted per-person one.}}
  \label{fig:supp_qual_results_2}
\end{figure*}

We present some examples of STEERER-V's zero-shot predictions on the real-world images of CrowdHuman~\cite{shao2018crowdhuman} dataset. We remark that this dataset lacks ground-truth volume annotations, which are roughly approximated by imputing the average real-world volume to each individual in the images, leveraging the statistics from \cite{silverman2022exact} (cf. Sec~5.3 of the main paper). It is worth noting that such an approximation strongly assumes that people are all similar in size to the average adult and that genders are equally represented in crowds. \\
As Fig.~\ref{fig:supp_qual_results_1} shows, STEERER-V reasonably estimates the volume of adults in indoor environments, large crowds, and with severe occlusions, diverging from the mean to take into account diverse builds, e.g., the men in the foreground of the first row, and uneven balance of genders, e.g., the crowds displayed in the second and third row. Our model fairly performs even with crowds of kids, totally absent from the GTA-V game. The image in the second row of Fig.~\ref{fig:supp_qual_results_2} stresses how the model assigns greater volume to the adults in the right foreground than to the surrounding children.\\
STEERER-V struggles with low-quality images, such as in the example of Fig.~\ref{fig:supp_qual_results_3}.


\newpage
\section{Qualitative Evaluation: additional results on \ourdataset{}}\label{sec:additional_results_steerer_v}

In Table~\ref{tab:sup_qual_results}, we present additional qualitative comparisons between the baseline model, STEERER, and our proposed model, STEERER-V. The comparison demonstrates that when faces are clearly visible and occlusions are minimal, the performance of both models is similar, as shown in the first and second rows of the table. However, in scenarios where occlusions occur, either due to other pedestrians or environmental elements, STEERER-V outperforms STEERER significantly, as evidenced from the third to the sixth row. In these images, it is evident that STEERER fails to attribute any volume to several individuals. Moreover, in the third row, we show that both models have learned that, from bird's-eye-view camera angles, environmental elements like trees can hide persons. Nevertheless, STEERER-V demonstrates superior robustness by not erroneously assigning any volume to the space obscured by the upper part of trees, highlighting its enhanced capability in handling such occlusions.

\begin{table*}[!htbp]
  \centering
  \begin{tabular}{ccc}
    {\footnotesize ANTHROPOS-V} &
    {\footnotesize STEERER } &
    {\footnotesize STEERER-V } \\
    
      \includegraphics[width=0.32\linewidth]{images/supplementary/other_qualitatives/2_all_1-1.pdf} &

      \includegraphics[width=0.32\linewidth]{images/supplementary/other_qualitatives/2_all_4-1.pdf} &

      \includegraphics[width=0.32\linewidth]{images/supplementary/other_qualitatives/2_all_3-1.pdf} \\
            
      \includegraphics[width=0.32\linewidth]{images/supplementary/other_qualitatives/4_all_1-1.pdf} &

      \includegraphics[width=0.32\linewidth]{images/supplementary/other_qualitatives/4_all_4-1.pdf} &

      \includegraphics[width=0.32\linewidth]{images/supplementary/other_qualitatives/4_all_3-1.pdf} \\
        
      \includegraphics[width=0.32\linewidth]{images/supplementary/other_qualitatives/3_all_1-1.pdf} &

      \includegraphics[width=0.32\linewidth]{images/supplementary/other_qualitatives/3_all_4-1.pdf} &

      \includegraphics[width=0.32\linewidth]{images/supplementary/other_qualitatives/3_all_3-1.pdf} \\
        
      \includegraphics[width=0.32\linewidth]{images/supplementary/other_qualitatives/11_all_1-1.pdf} &

      \includegraphics[width=0.32\linewidth]{images/supplementary/other_qualitatives/11_all_4-1.pdf} &

      \includegraphics[width=0.32\linewidth]{images/supplementary/other_qualitatives/11_all_3-1.pdf} \\

      \includegraphics[width=0.32\linewidth]{images/supplementary/other_qualitatives/15_all_1-1.pdf} &

      \includegraphics[width=0.32\linewidth]{images/supplementary/other_qualitatives/15_all_4-1.pdf} &

      \includegraphics[width=0.32\linewidth]{images/supplementary/other_qualitatives/15_all_3-1.pdf} \\
      
     \includegraphics[width=0.32\linewidth]{images/supplementary/other_qualitatives/19_all_1-1.pdf} &

      \includegraphics[width=0.32\linewidth]{images/supplementary/other_qualitatives/19_all_4-1.pdf} &

      \includegraphics[width=0.32\linewidth]{images/supplementary/other_qualitatives/19_all_3-1.pdf} \\

\end{tabular}
\vspace{0.5em}
\caption{{Visualization results of STEERER and STEERER-V on ANTHROPOS-V crowded images. STEERER's density map highlights volume on head positions, while STEERER-V's density map emphasizes the volume spread on the whole body.}}
\label{tab:sup_qual_results}
\end{table*}

\newpage
\onecolumn
\section{Examples of images of \ourdataset{}}\label{sec:anthropos_qualitatives}
\begin{figure}[!ht]
    \centering
    {\renewcommand{\arraystretch}{0.5}
    \begin{tabular}{cc}
    
        ANTHROPOS-V & GTA-V \\
        
        \includegraphics[width=0.40\linewidth]{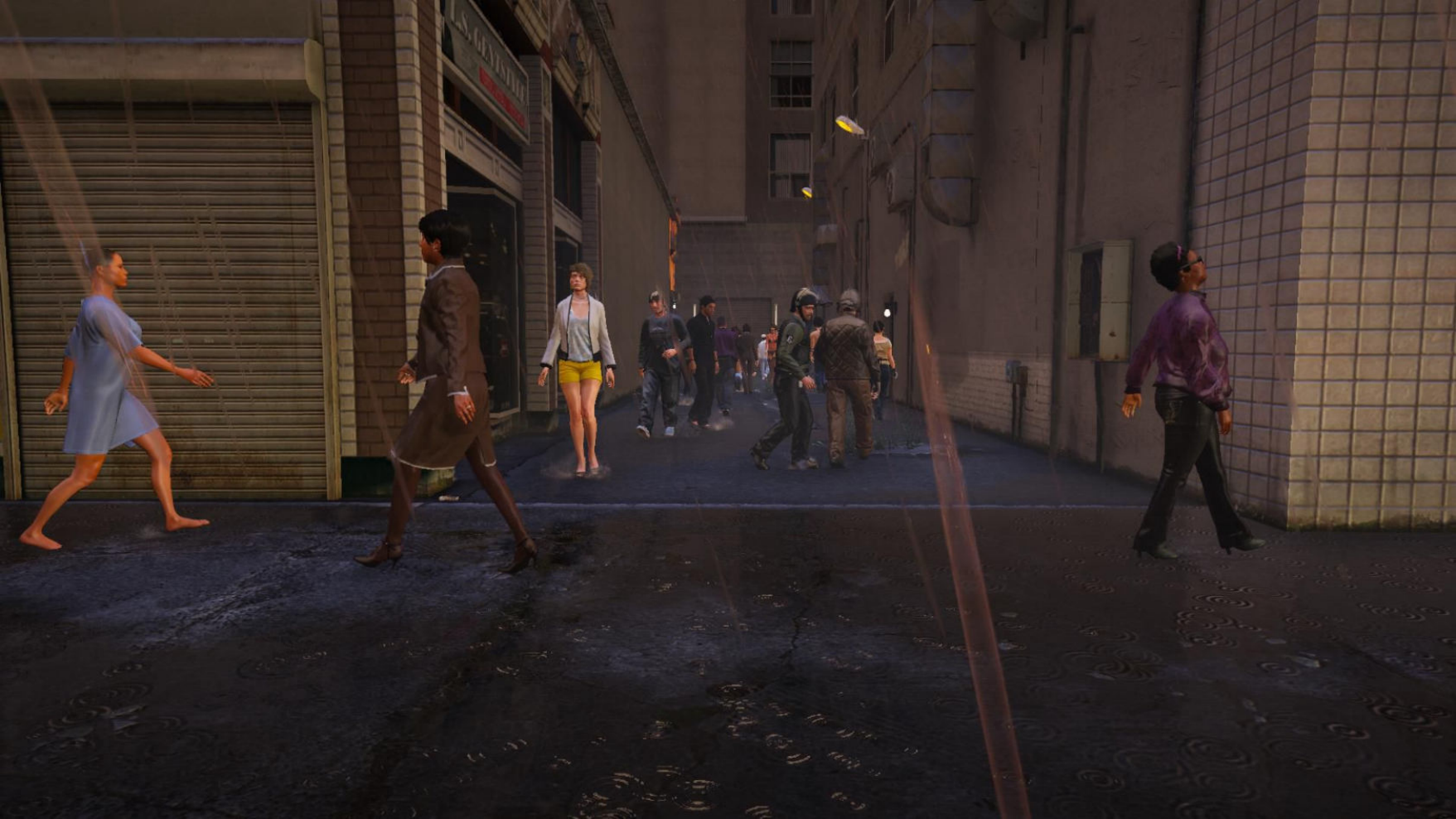} &
        \includegraphics[width=0.40\linewidth]{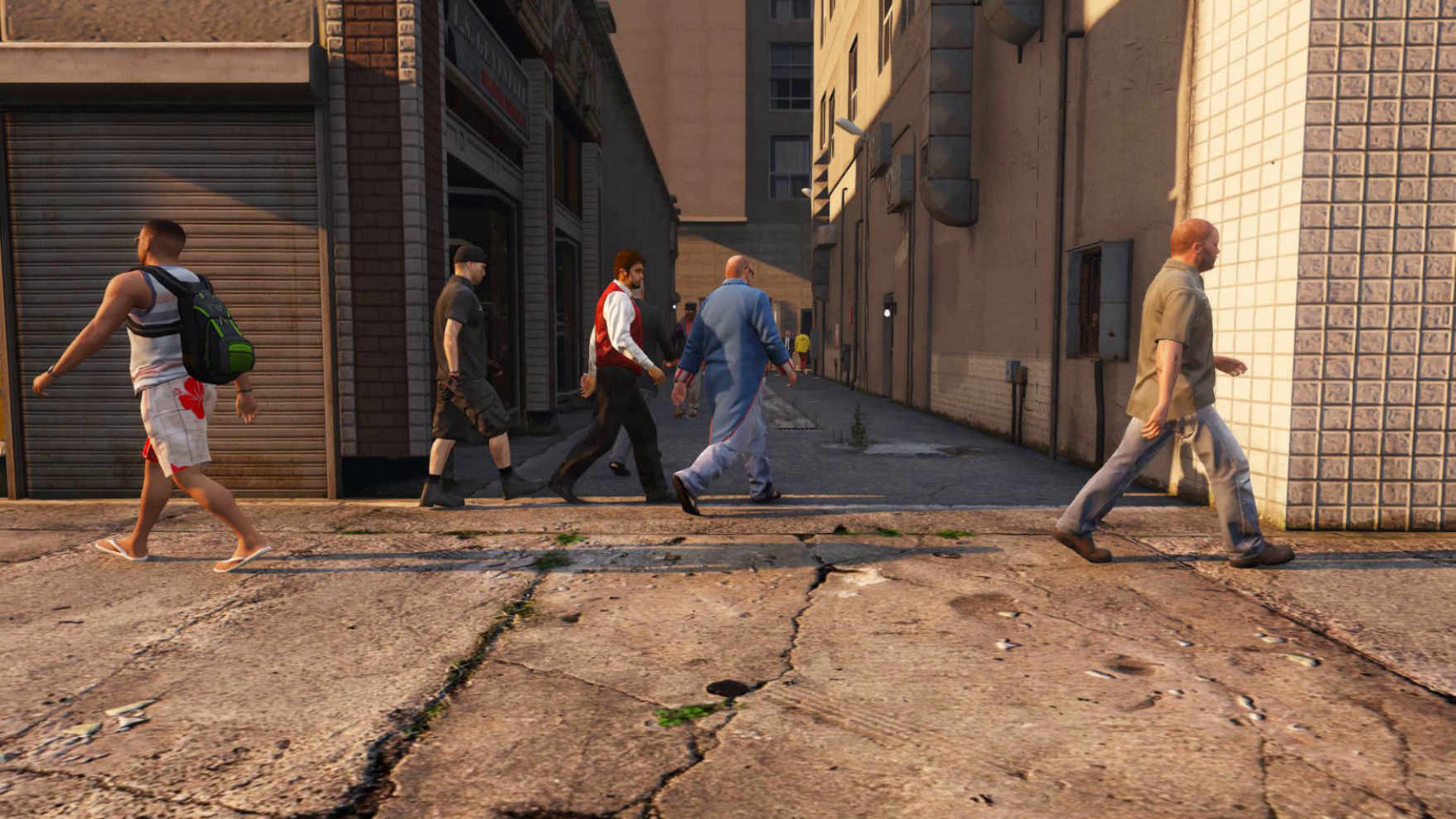} \\
    
        \includegraphics[width=0.40\linewidth]{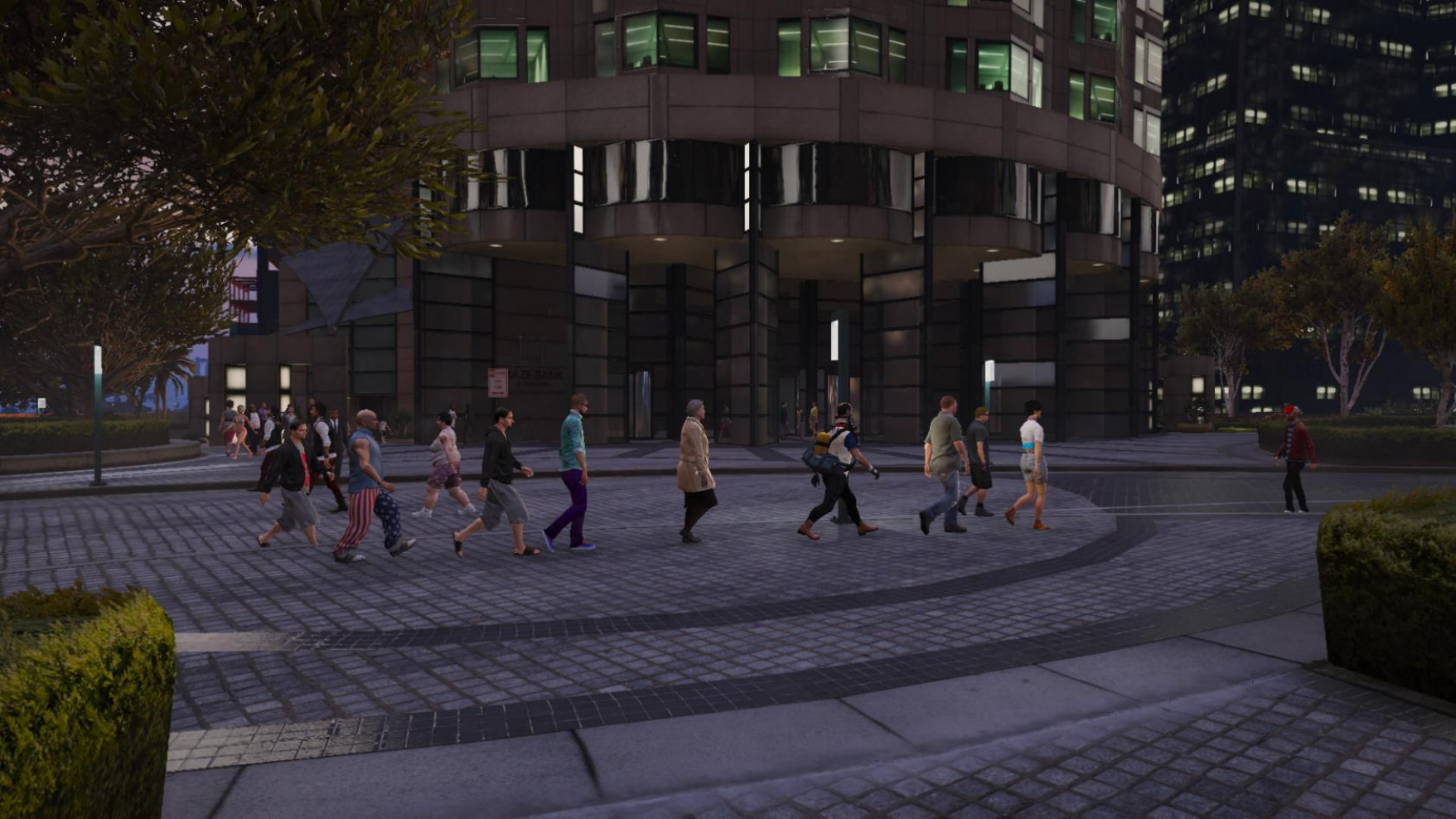} &
        \includegraphics[width=0.40\linewidth]{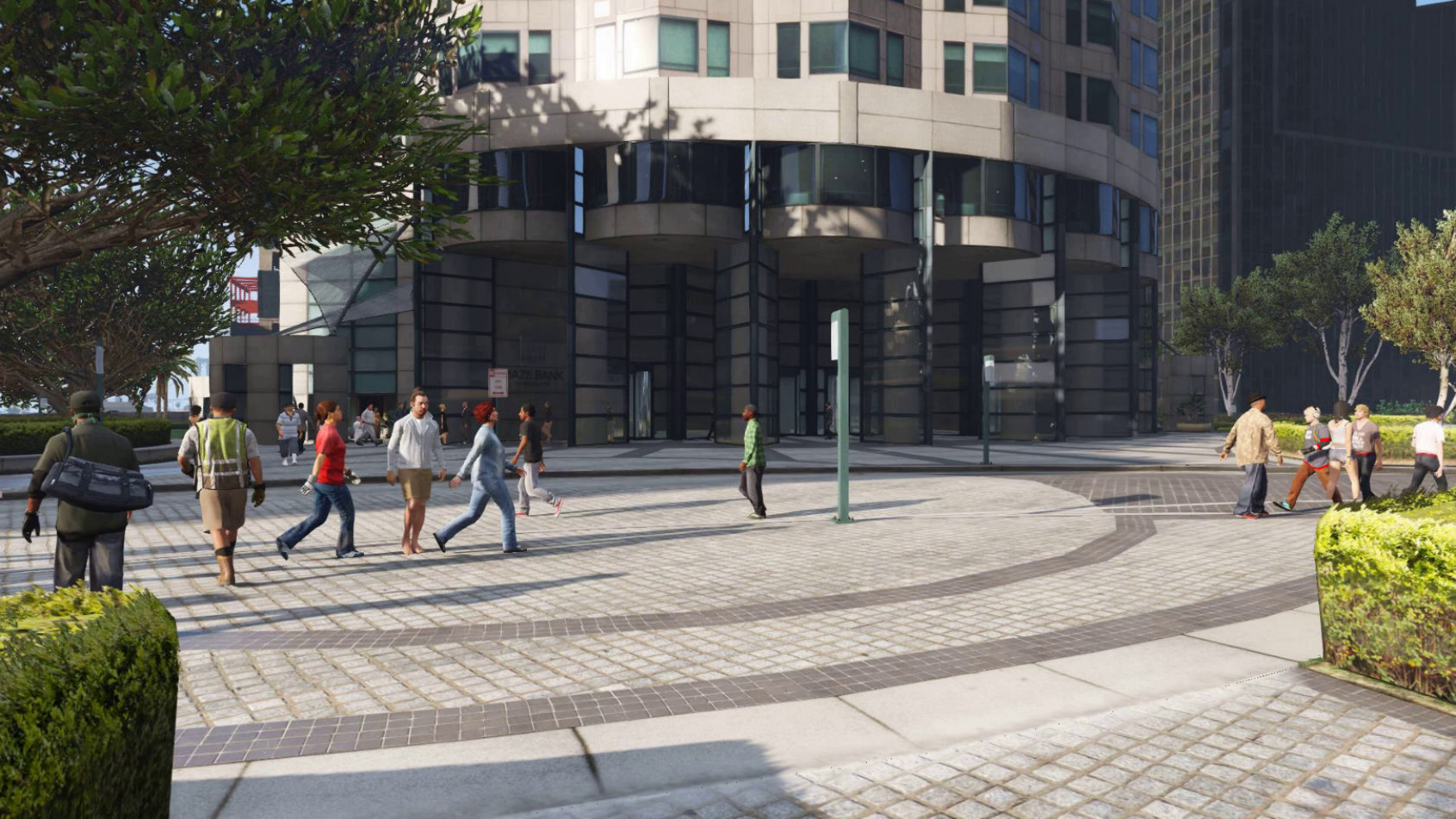} \\
    
        \includegraphics[width=0.40\linewidth]{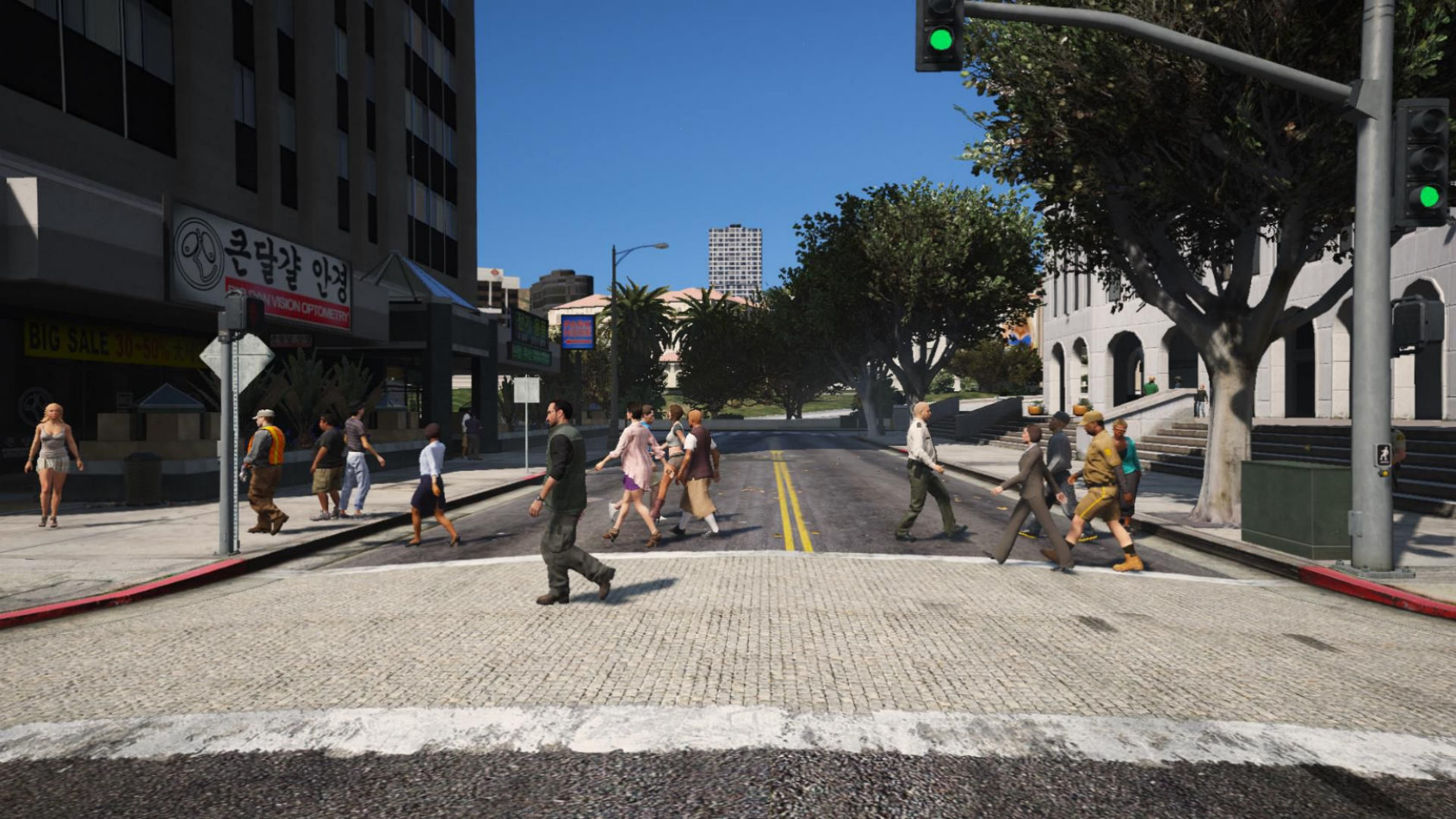} &
        \includegraphics[width=0.40\linewidth]{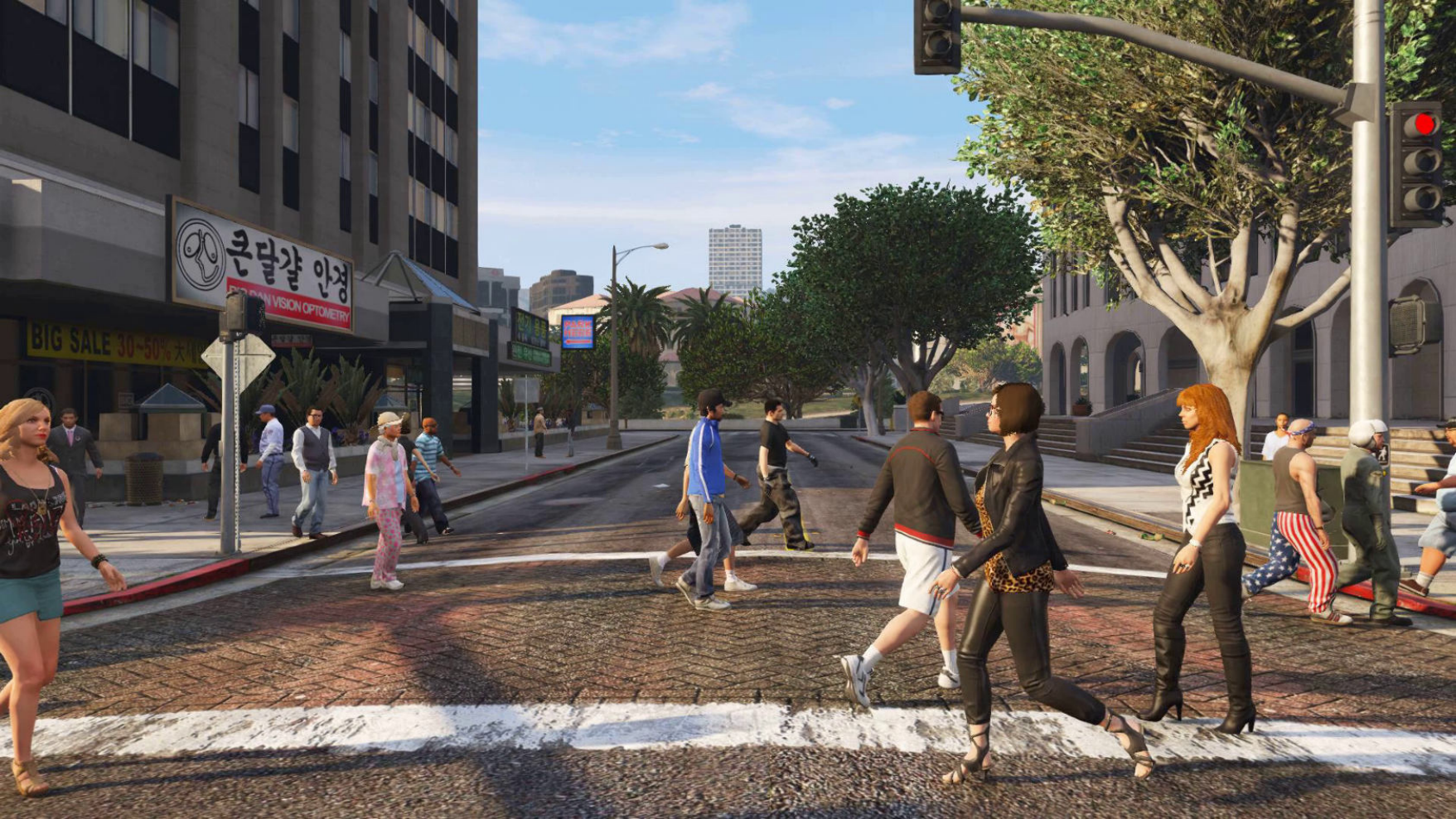} \\
            
        \includegraphics[width=0.40\linewidth]{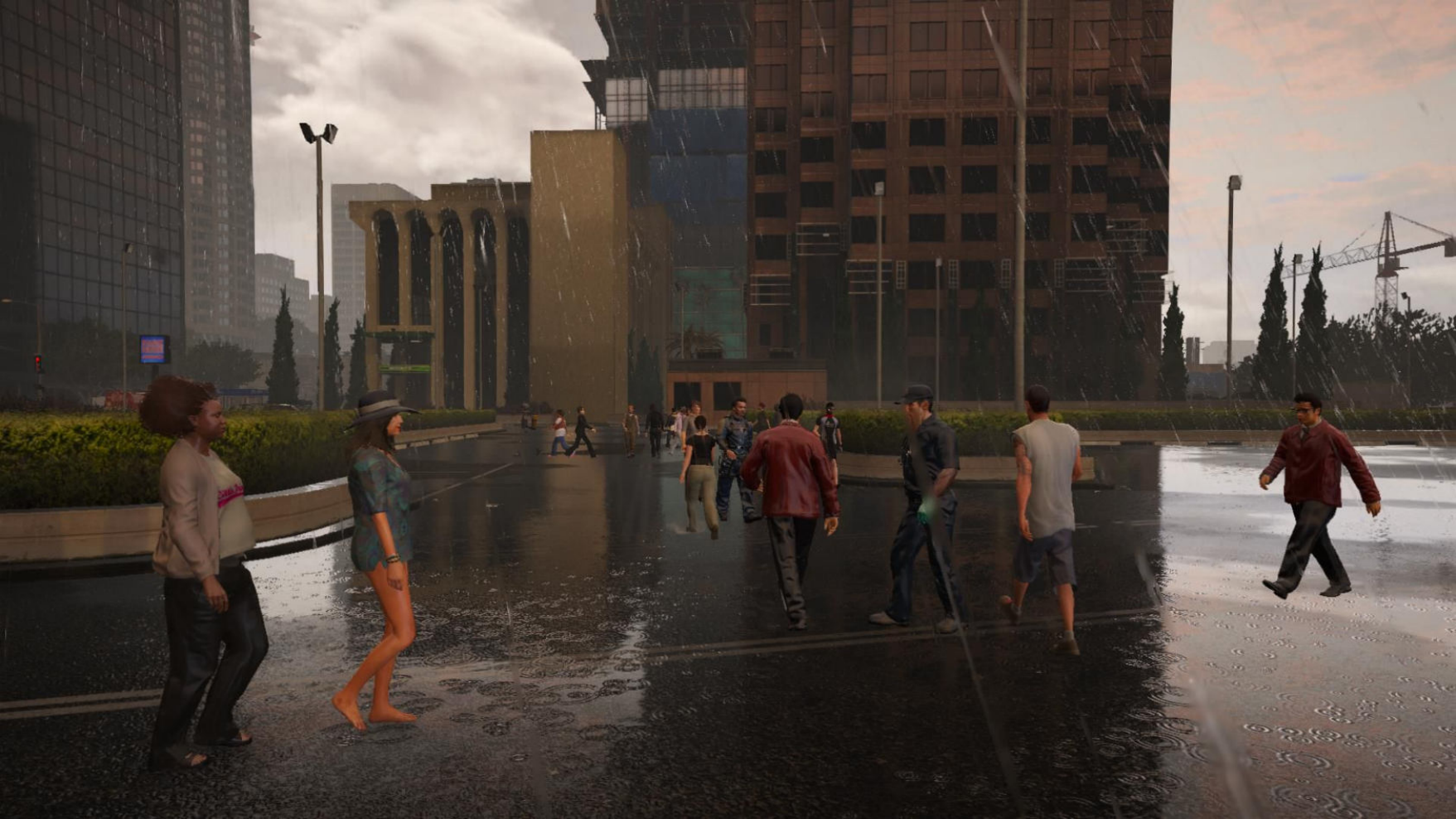} &
        \includegraphics[width=0.40\linewidth]{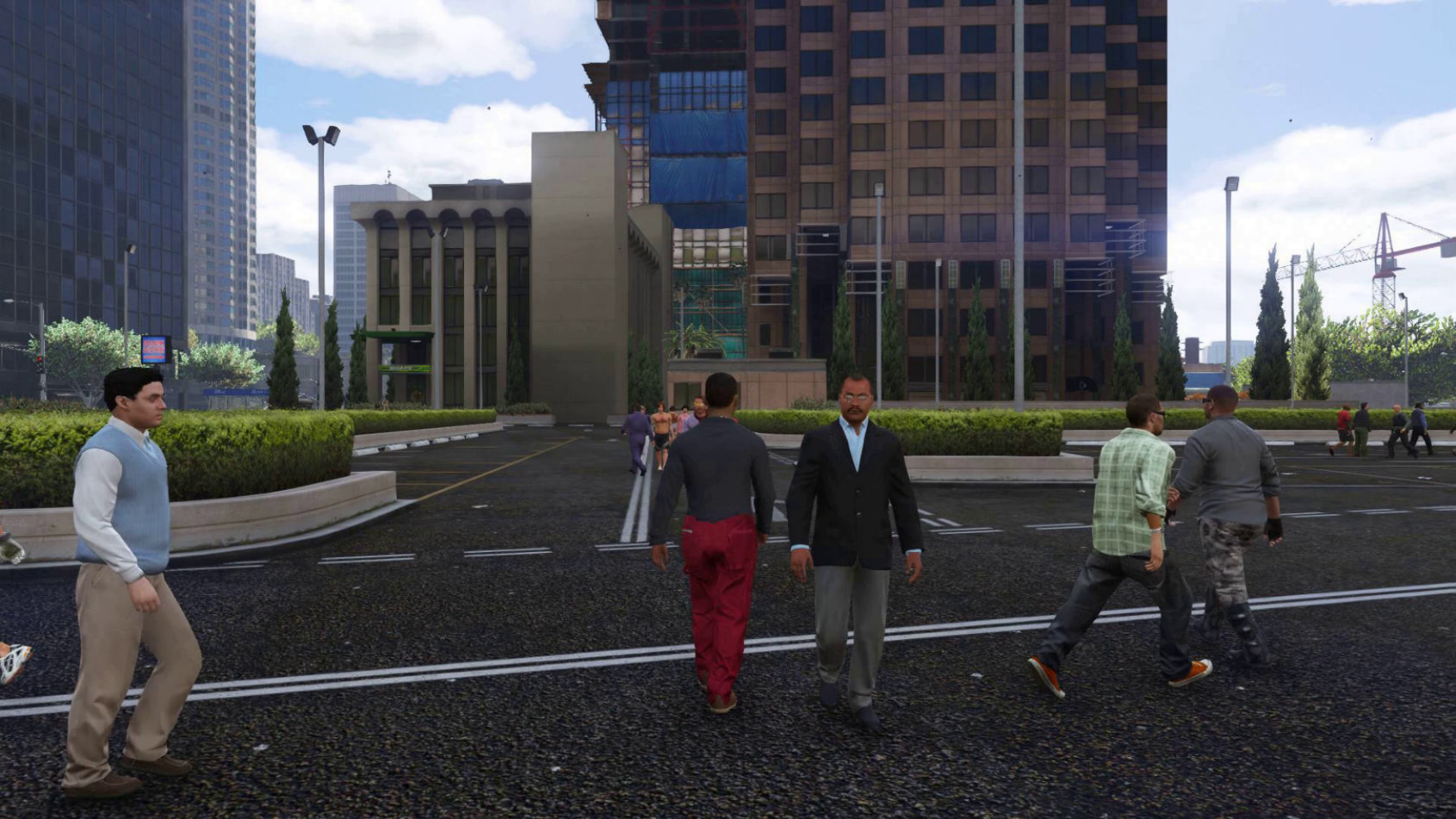} \\
    
        \includegraphics[width=0.40\linewidth]{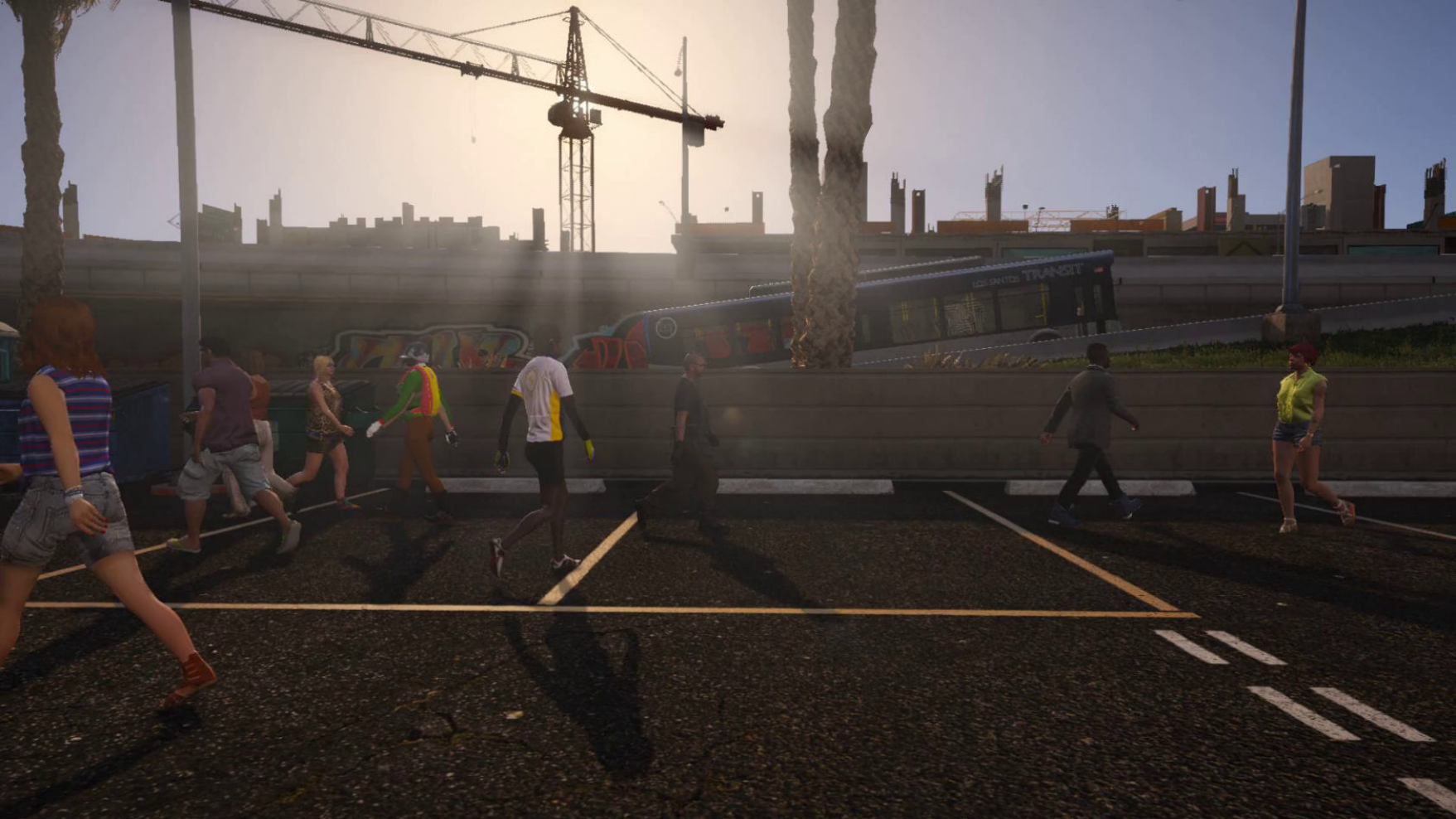} &
        \includegraphics[width=0.40\linewidth]{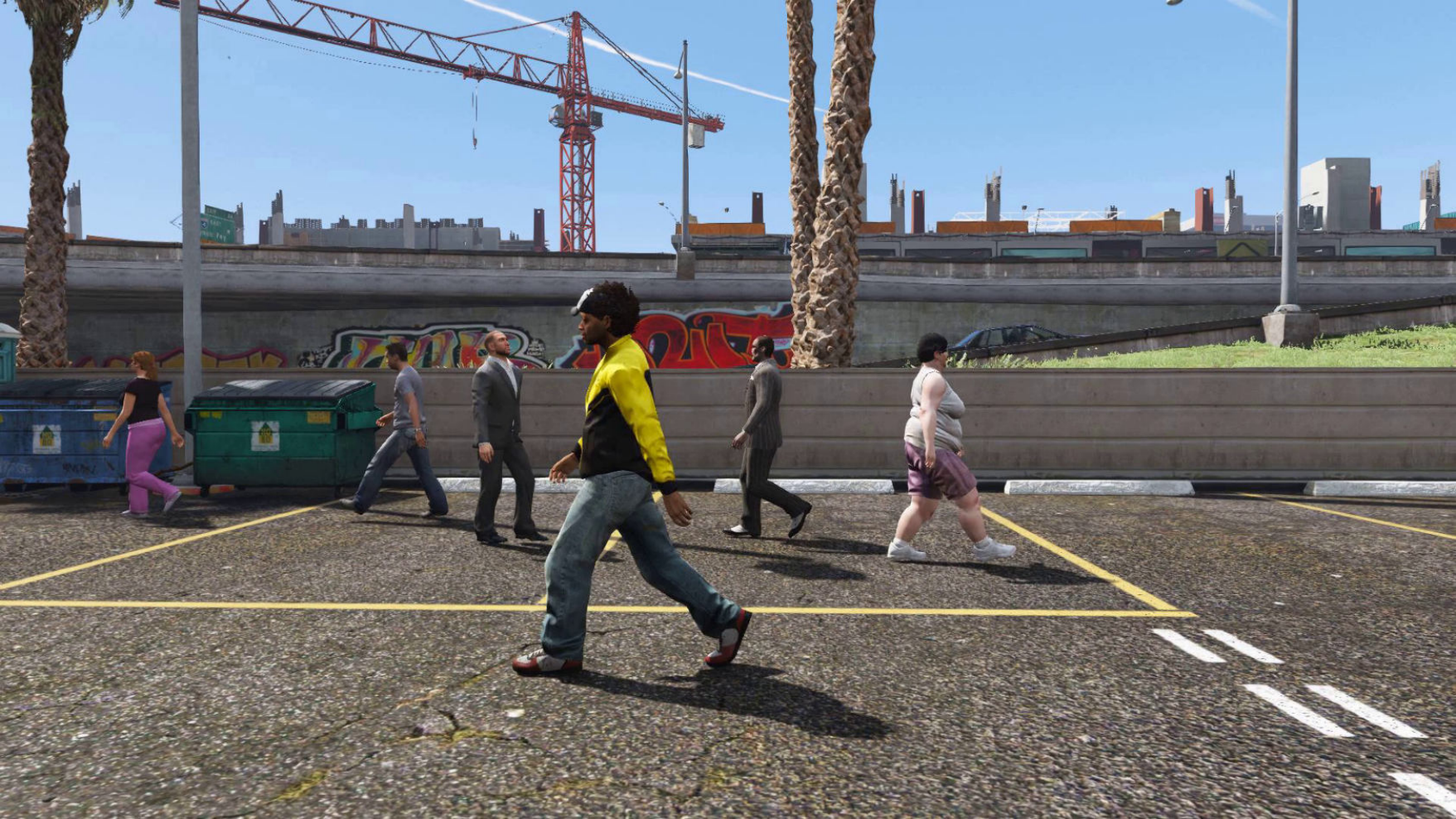} \\
    \end{tabular}
    }

\caption{Examples of frames from \ourdataset{} (left column) and original GTA-V footages (right column), as synthesized via \cite{fabbri2021motsynth}. Images on the left present a wider variety of people's heights.}\label{fig:qualitative_scaling}
\end{figure}
\twocolumn


\begin{figure*}[ht]
    \centering
    {\renewcommand{\arraystretch}{0.5}
    \begin{tabular}{cc}
        ANTHROPOS-V & GTA-V \\

        \includegraphics[width=0.40\linewidth]{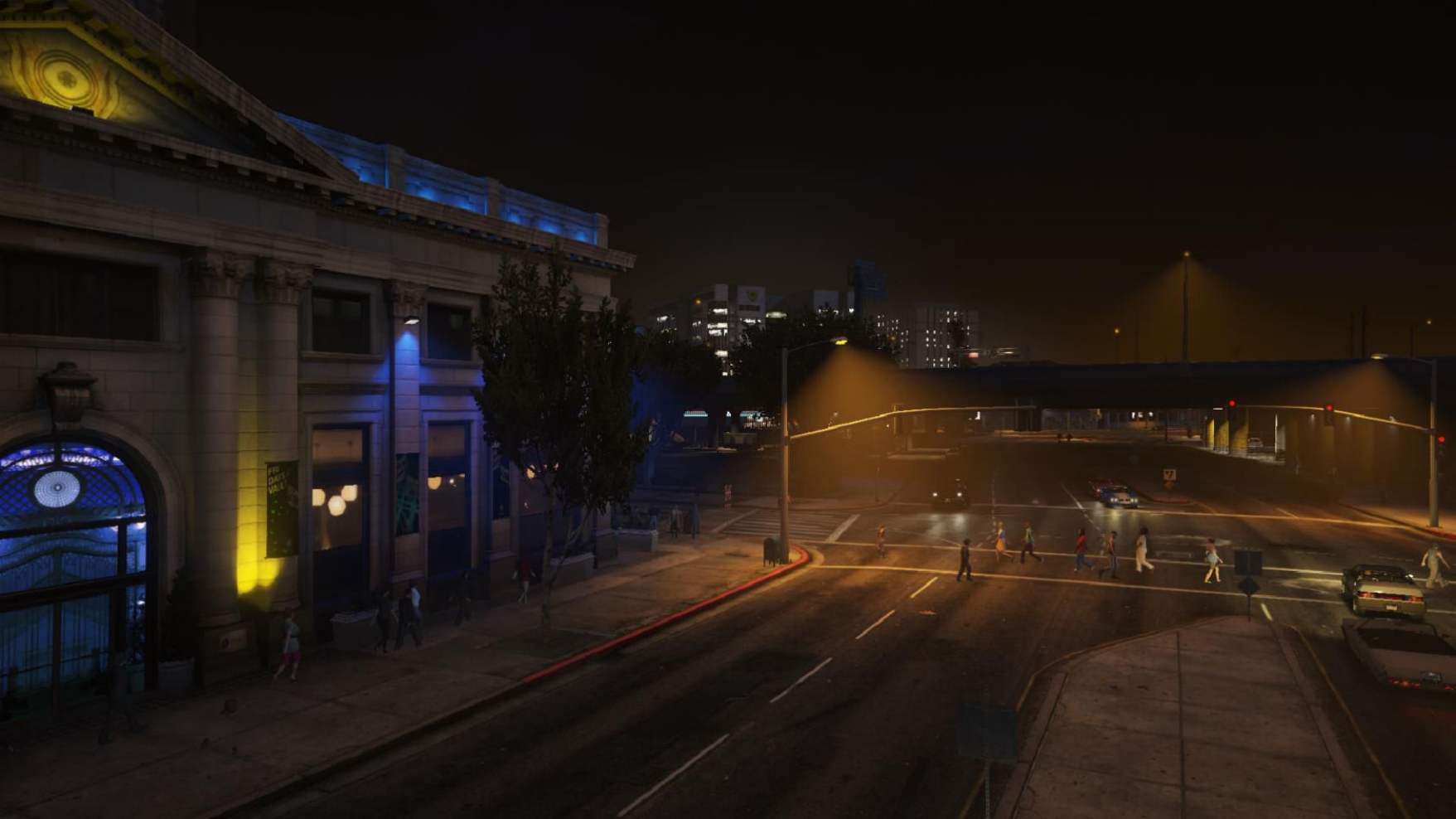} &
        \includegraphics[width=0.40\linewidth]{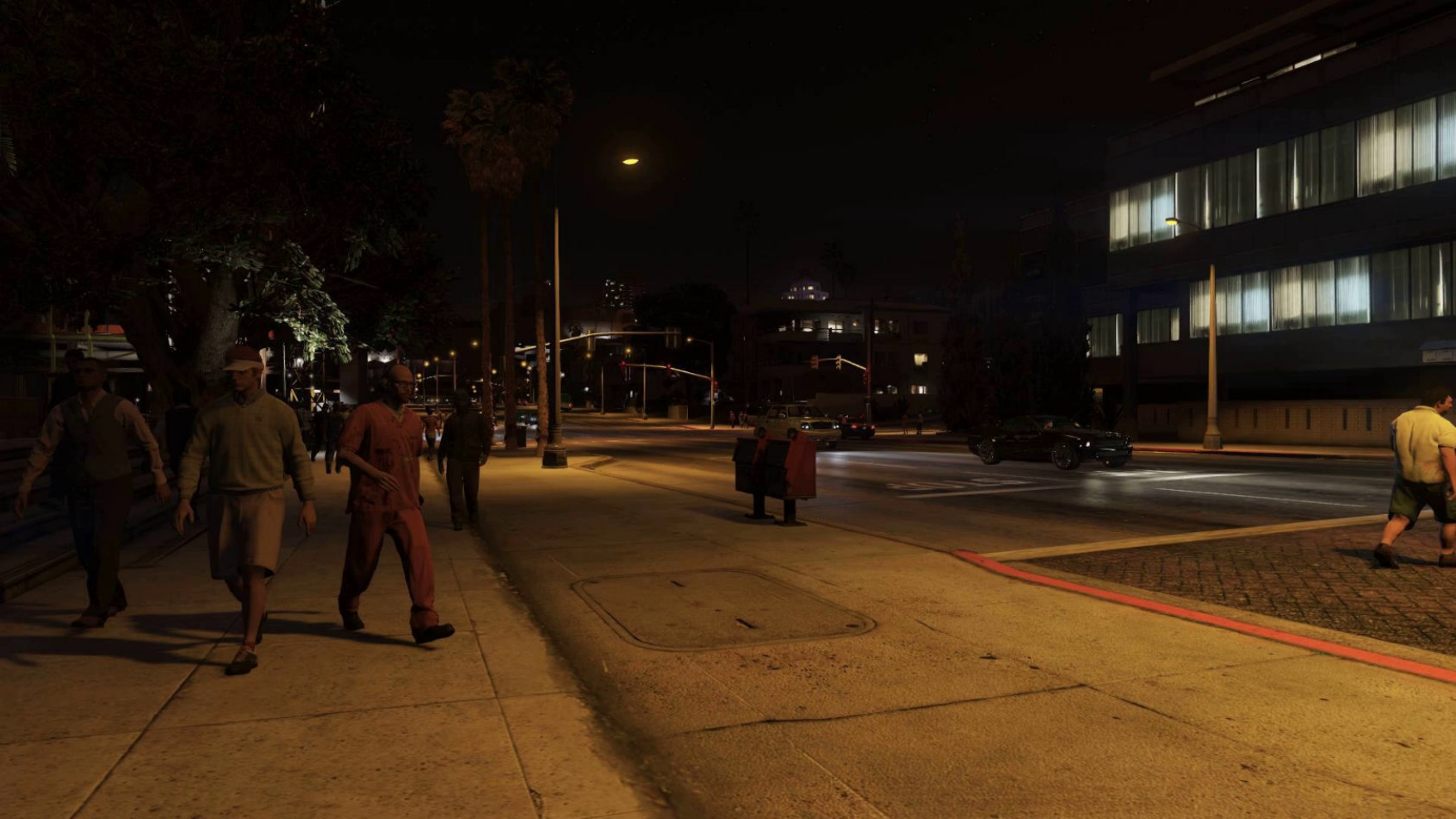} \\

        \includegraphics[width=0.40\linewidth]{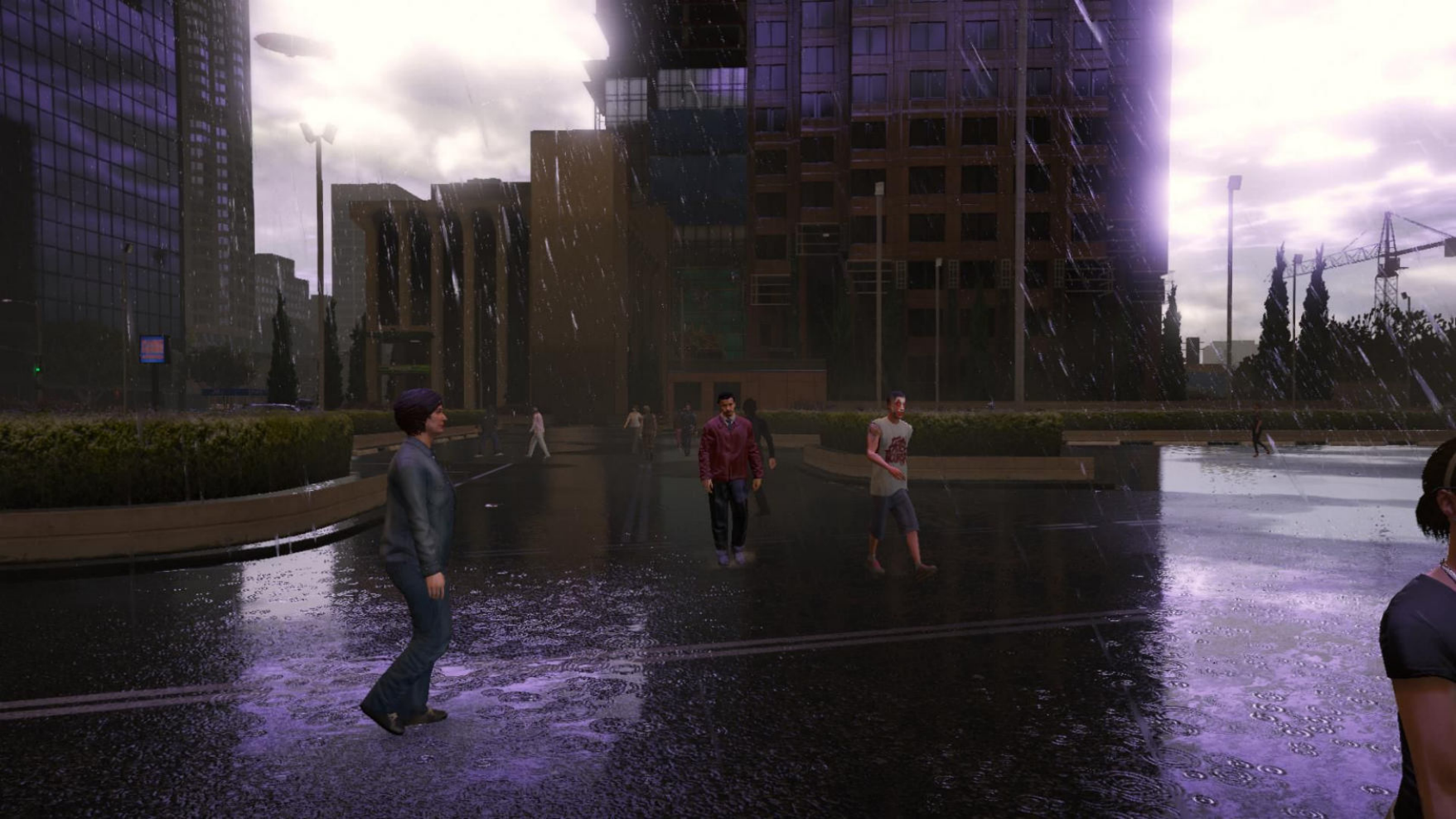} &
        \includegraphics[width=0.40\linewidth]{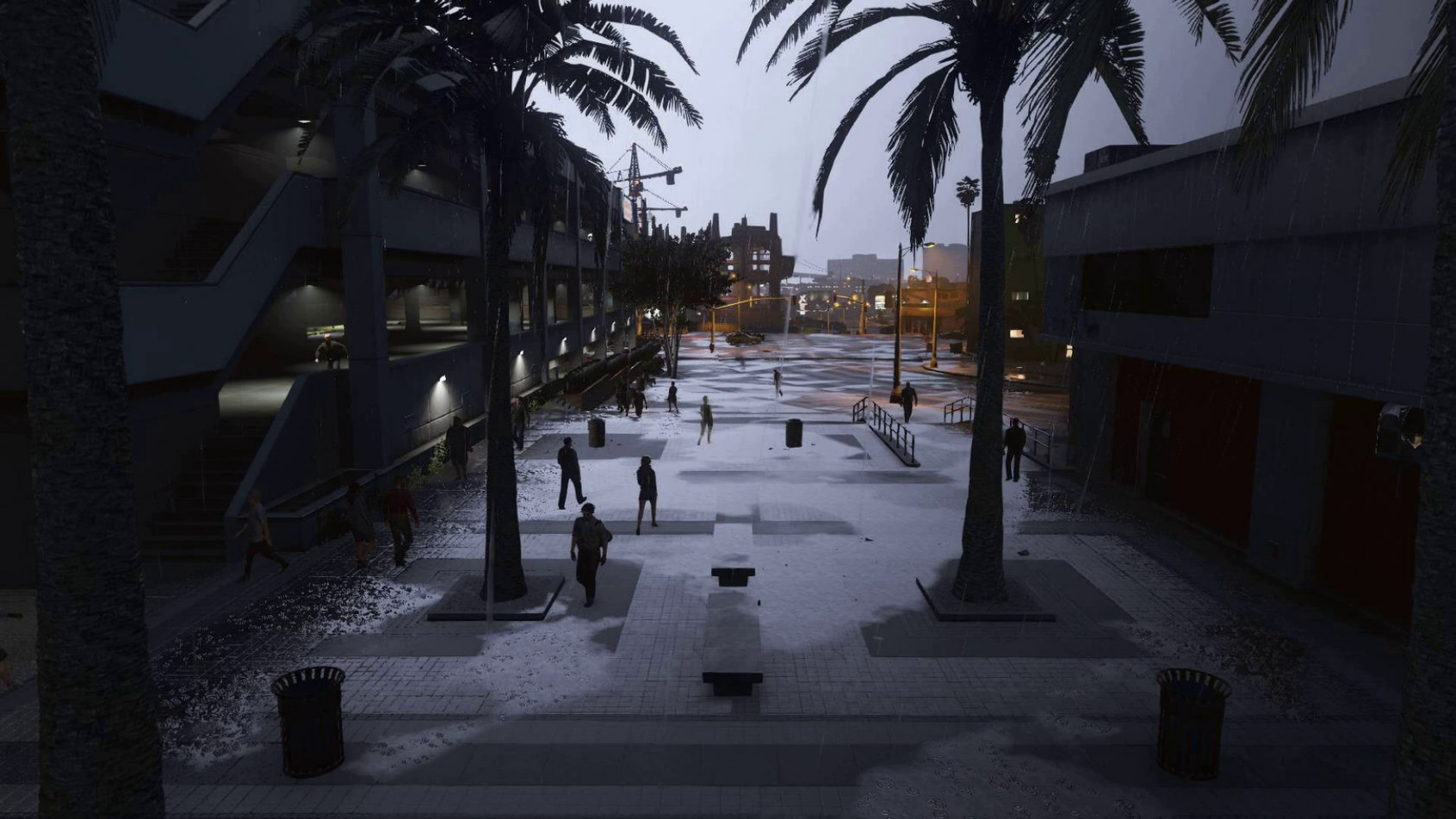} \\
        
        \includegraphics[width=0.40\linewidth]{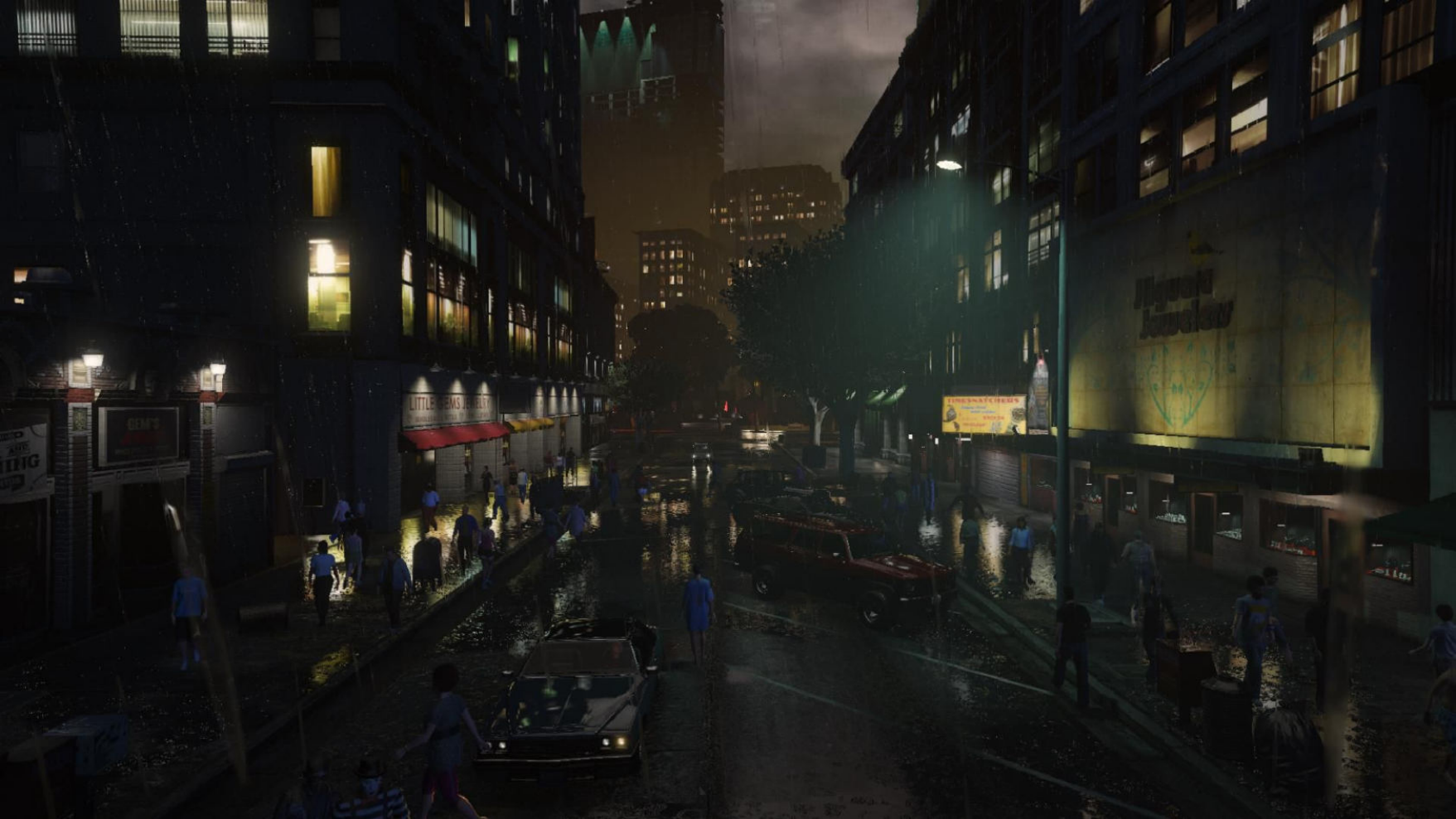} &
        \includegraphics[width=0.40\linewidth]{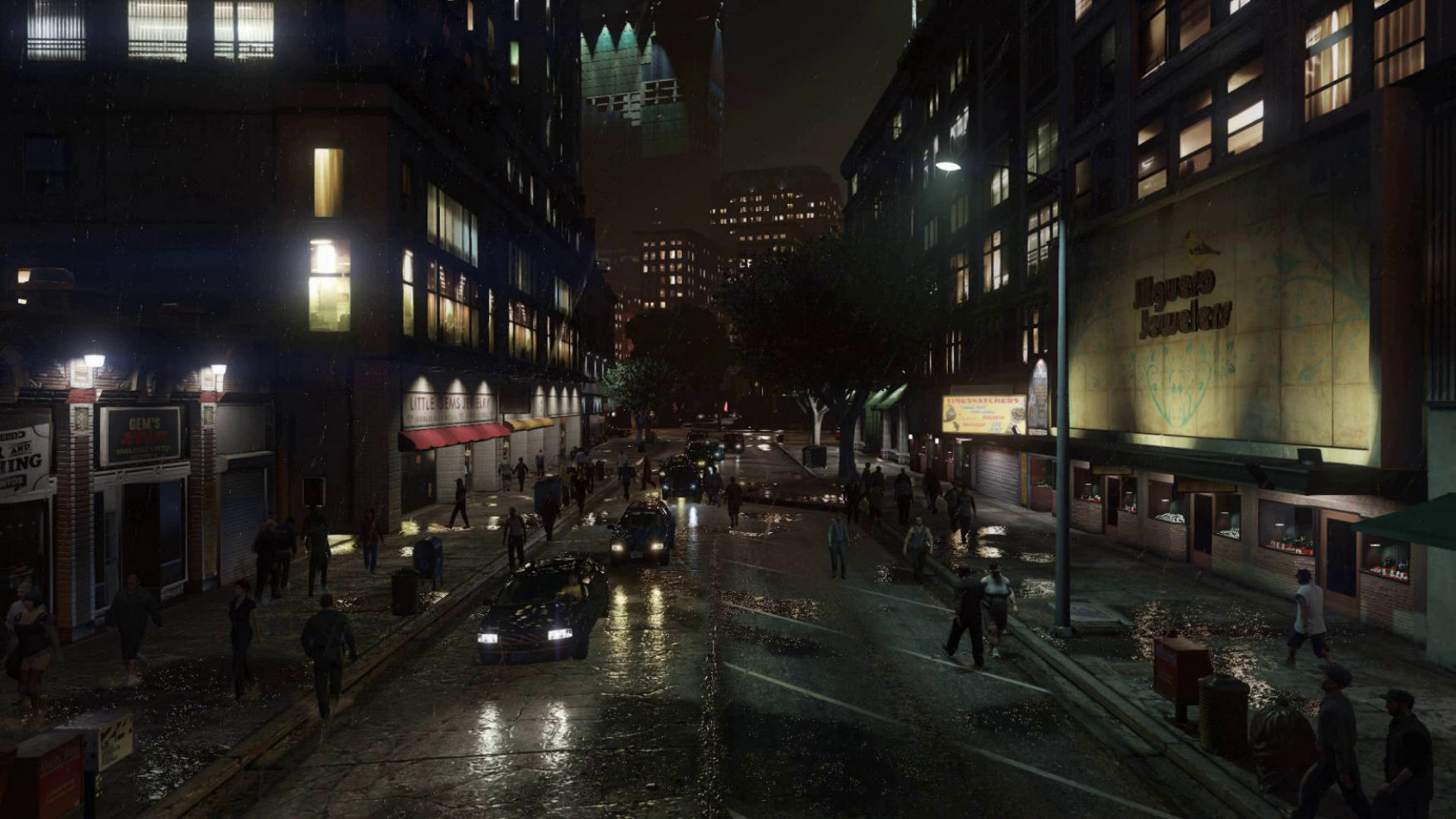} \\
            
        \includegraphics[width=0.40\linewidth]{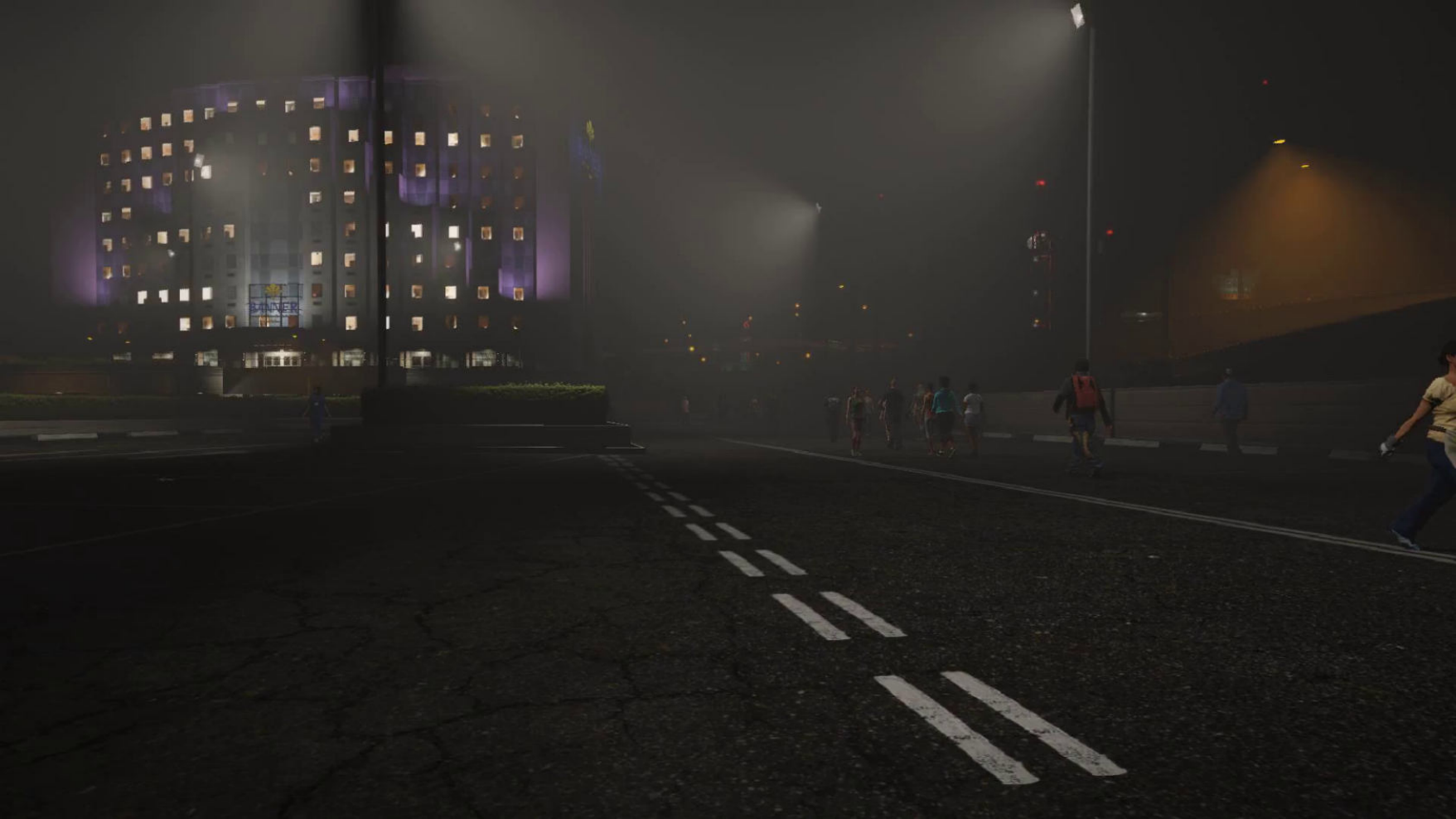} &
        \includegraphics[width=0.40\linewidth]{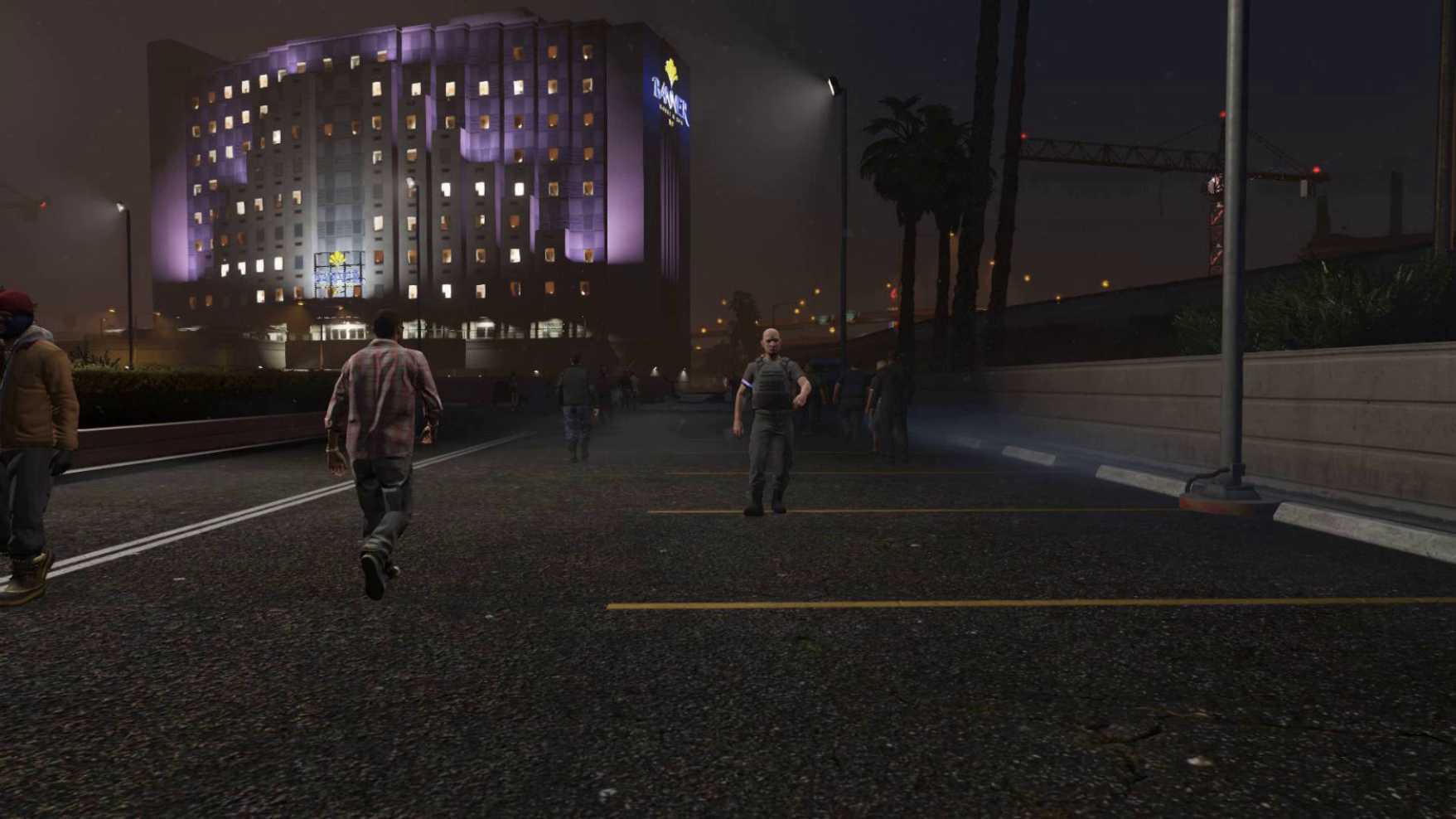} \\
            
        \includegraphics[width=0.40\linewidth]{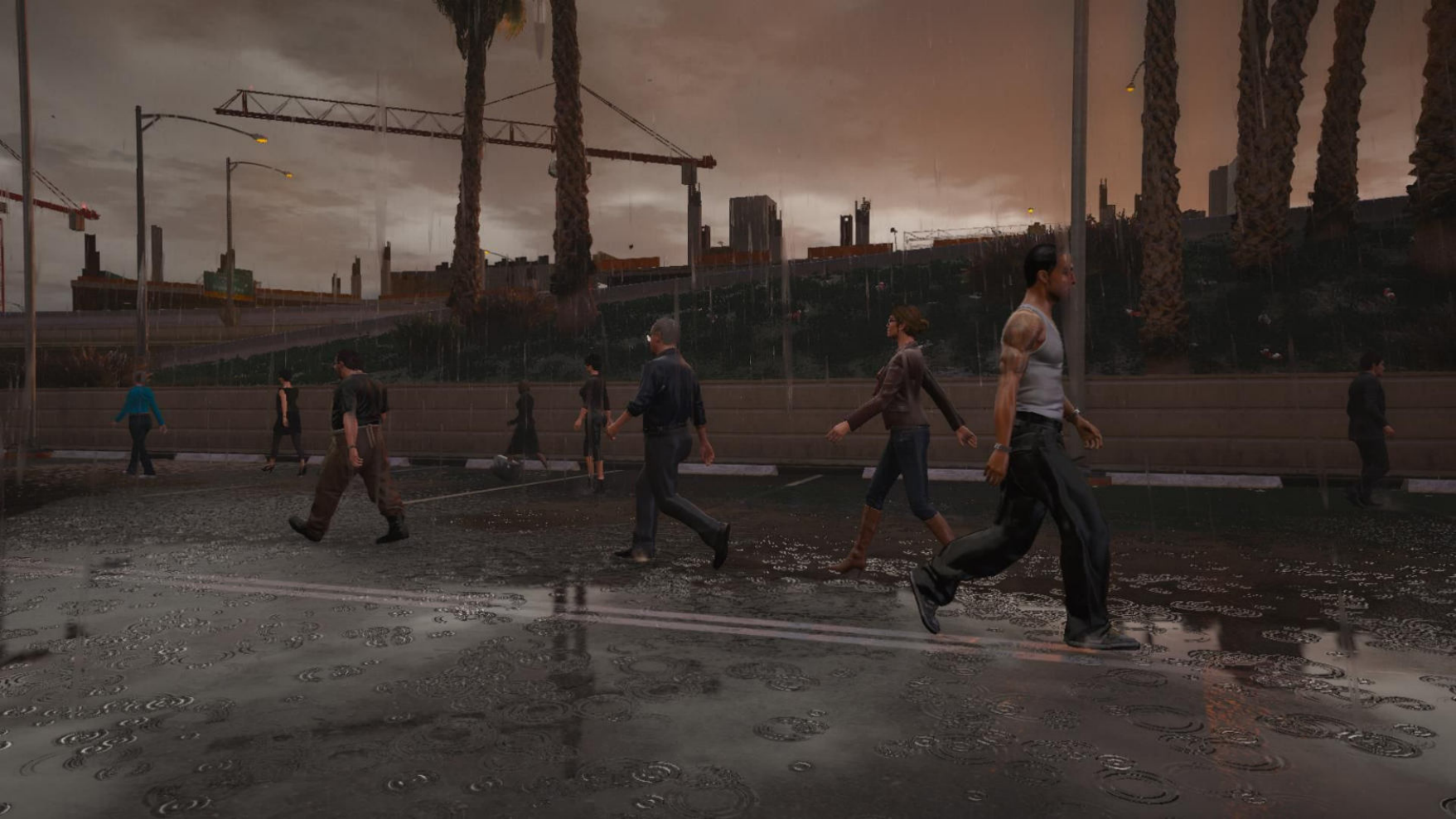} &
        \includegraphics[width=0.40\linewidth]{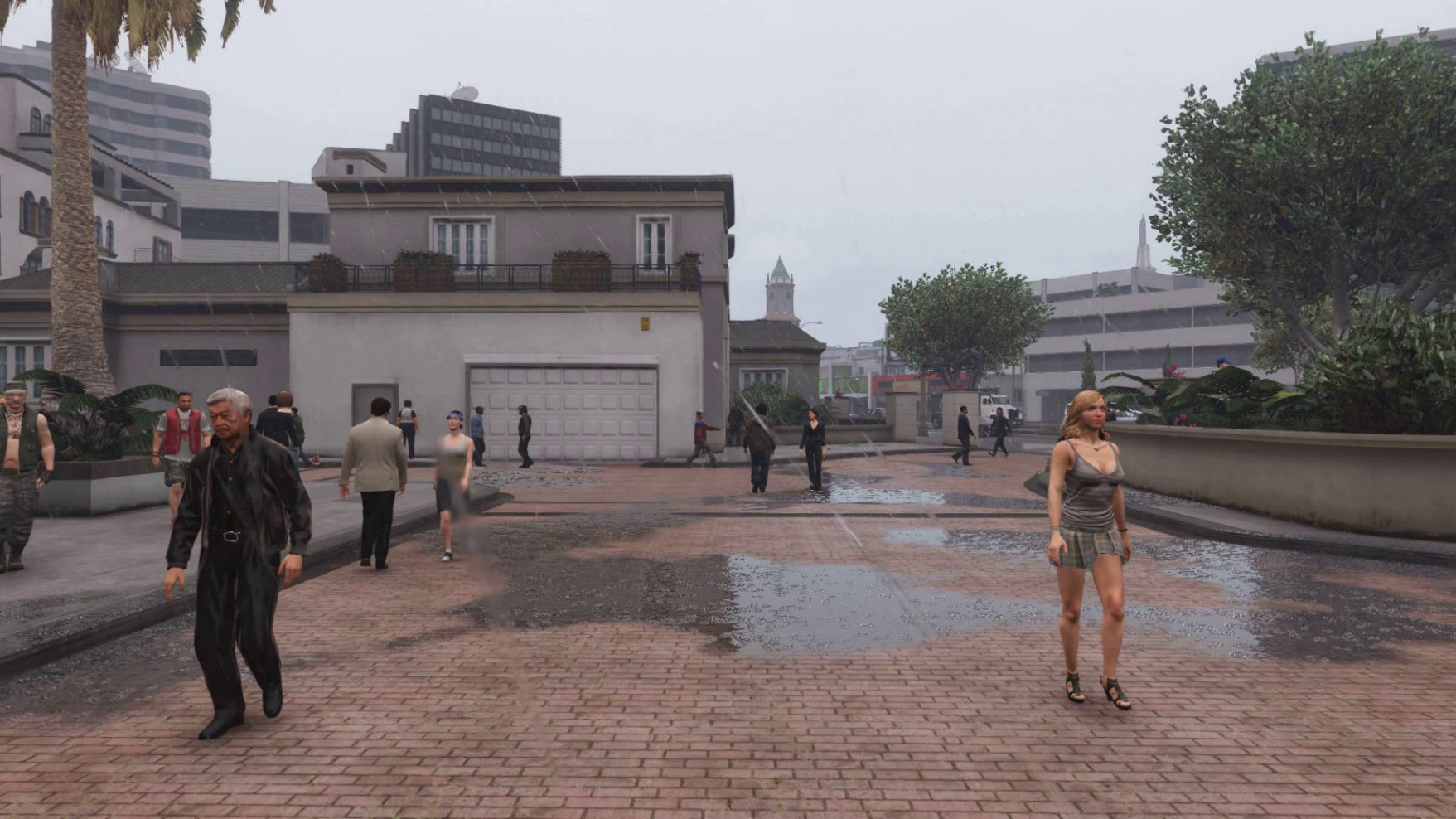} \\
    \end{tabular}
    }
\caption{Examples of frames from \ourdataset{} (left column) and original GTA-V footages (right column), as synthesized via \cite{fabbri2021motsynth}. Images on the left present more diverse weather and lighting conditions and better details.}\label{fig:photorealism-sup}
\end{figure*}

\newpage
\clearpage

\section{Cross Dataset Evaluation}
\label{sec:cross}

In this section, we perform an additional experiment that assesses the performance of models trained on HMR datasets via Cross Dataset Evaluation. 
The next section will introduce the additional datasets we leverage for this study, while Sec.~\ref{sup:cross_dataset_exp} describes the experiment's outcomes.
\begin{figure*}[t]
    
    \centering
    \begin{subfigure}[b]{0.32\textwidth}
        \includegraphics[width=\textwidth]{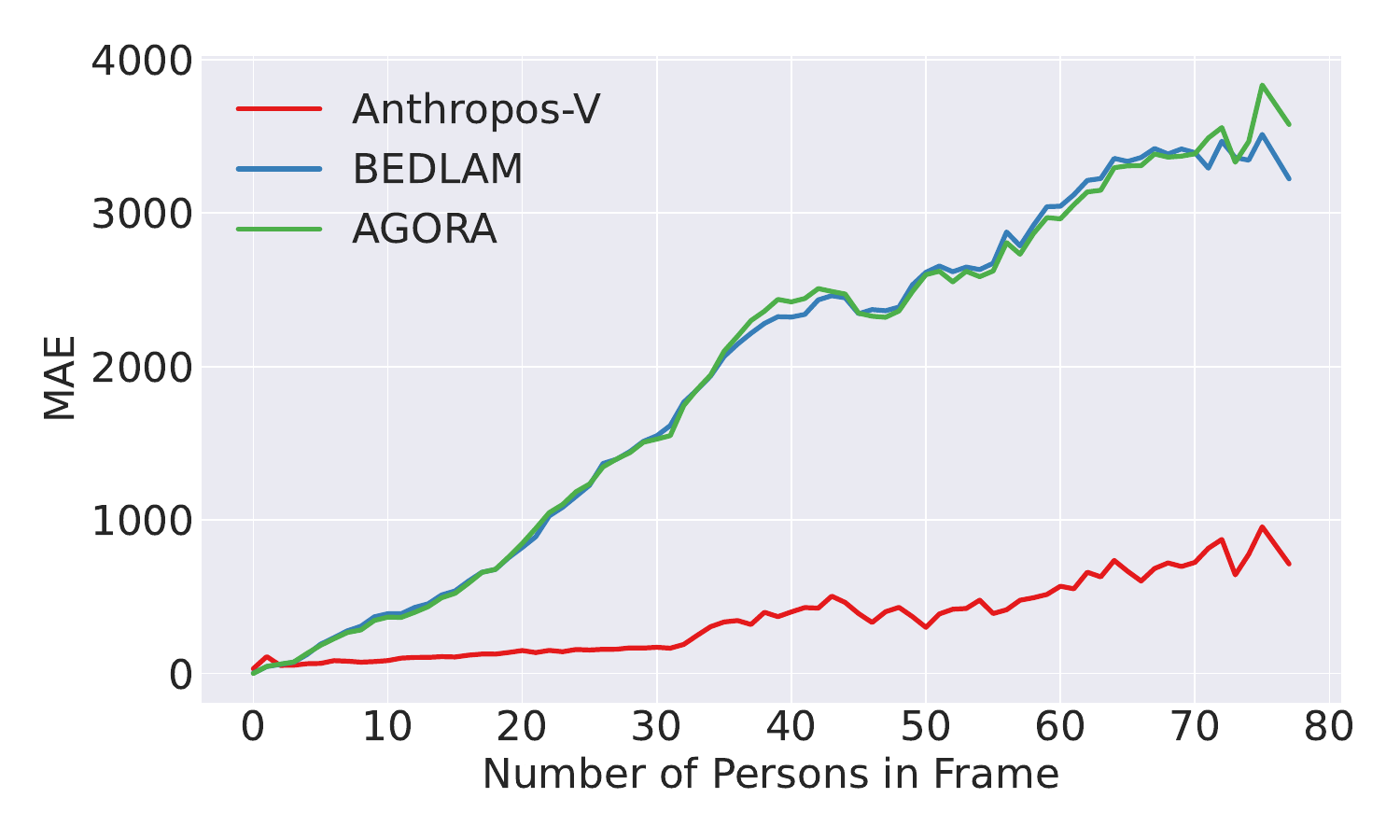}
        \caption{Tested on \ourdataset}
        \label{fig:img1}
    \end{subfigure}
    \hfill
    \vspace{0.15cm}
    \begin{subfigure}[b]{0.32\textwidth}
        \includegraphics[width=\textwidth]{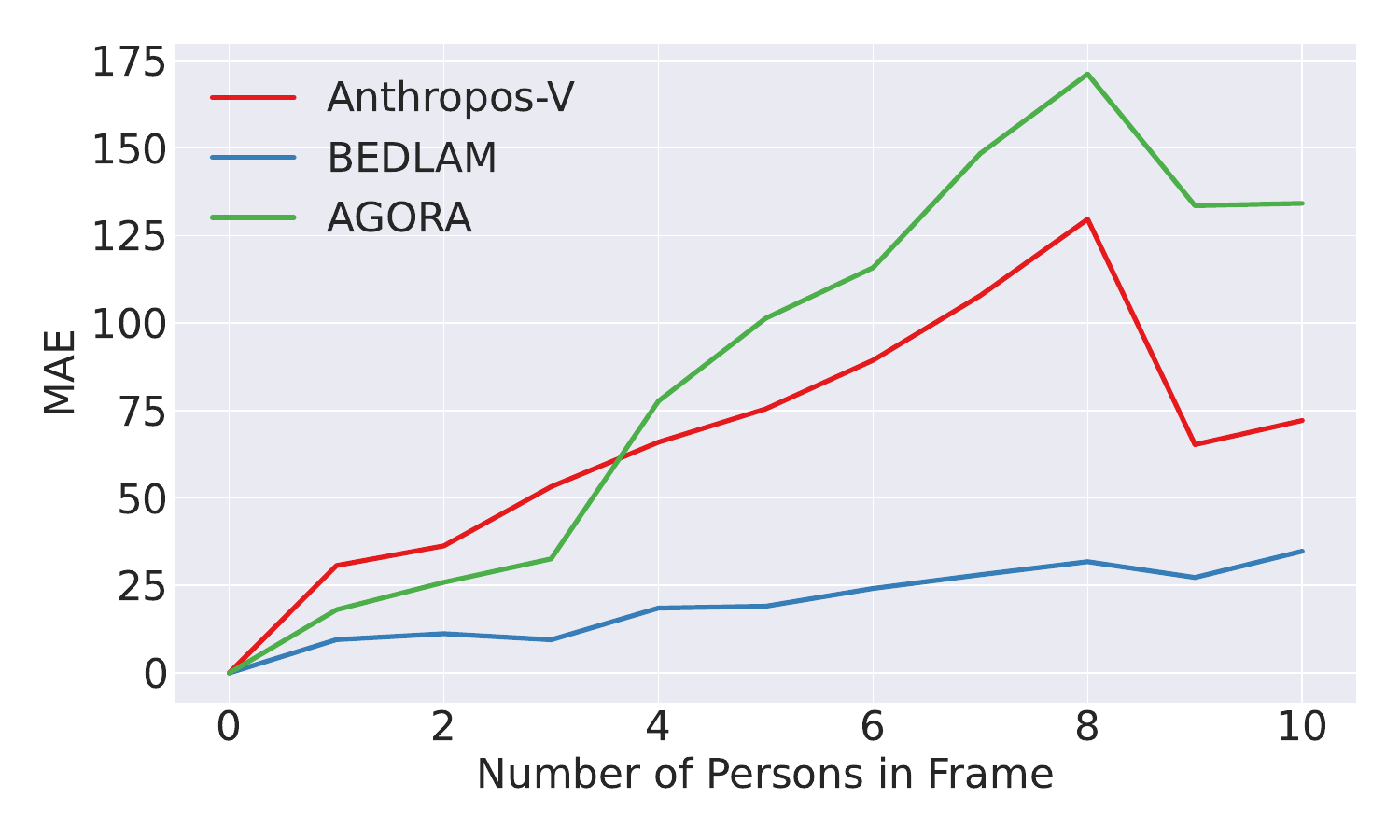}
        \caption{Tested on BEDLAM}
        \label{fig:img2}
    \end{subfigure}
    \hfill
    \vspace{0.15cm}
    \begin{subfigure}[b]{0.32\textwidth}
        \includegraphics[width=\textwidth]{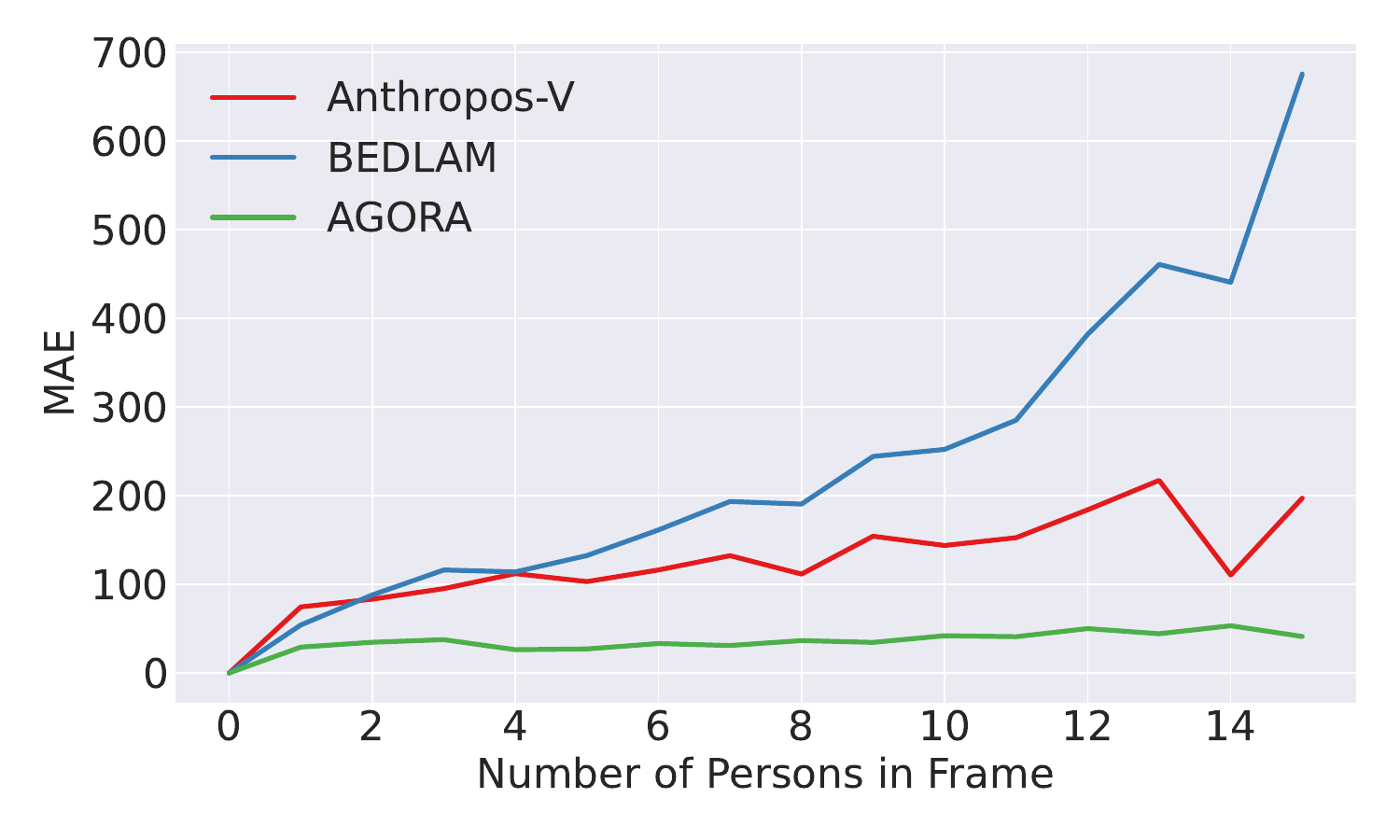}
        \caption{Tested on AGORA}
        \label{fig:img3}
    \end{subfigure}
    \vspace{-1.0em}
    \caption{Error trends of STEERER-V with respect to the growing number of individuals.}

    \label{img:crossdataset}

\end{figure*}

\subsection{HMR Datasets}\label{sup:hmr_datasets}

AGORA~\cite{patel2021agora} is a synthetic image dataset with diverse adult and child characters with SMPL~\cite{loper2015smpl} annotation. The recent BEDLAM~\cite{black2023bedlam} is a comprehensive synthetic video dataset with 271 highly realistic and diverse SMPL-based characters. In contrast to our dataset, they don't target large crowds, having a limited number of people per scene ($\leq 15, 10$ in \cite{patel2021agora} and \cite{black2023bedlam}, respectively); hence, they are not optimal for CVE, as we show in the next section.

\subsection{Cross Dataset Evaluation}\label{sup:cross_dataset_exp}
To assess the capabilities and applications of \ourdataset{}, we conduct a cross-dataset evaluation using the latest human datasets annotated with SMPL meshes, specifically AGORA \cite{patel2021agora} and BEDLAM \cite{black2023bedlam}. It is important to note that these datasets predominantly feature small groups of people, with an average of 3.66 individuals per frame in BEDLAM and 9.08 in AGORA. Since these datasets lack ground-truth volumes, we annotate the volume based on the provided meshes.
In our experiment, we train STEERER-V on all three datasets and evaluate the performance on each dataset's test set. Fig.~\ref{img:crossdataset} displays the error rates for each test set, highlighting how they vary with the increasing number of people in a scene. As shown in Fig.~\ref{fig:img1}, models trained on datasets with smaller groups (represented by the green and blue lines) exhibit less robustness when faced with scenes containing more individuals. Conversely, Fig.~\ref{fig:img2} and Fig.~\ref{fig:img3} demonstrate that STEERER-V, when trained on our crowd dataset, maintains robustness regardless of the increasing number of people in the image.

\section{Decoupling Crowd Counting from Volume Estimation}~\label{sup:decoupling}

The CVE error metrics presented in this paper (MAE/PPMAE) compare per-frame ground truth volumes with model predictions. However, this error stems from two main sources: missed detections and incorrect volume estimations of individuals. To improve CVE models, both of these factors must be addressed. To assess the contribution of each error source to the overall error, we conduct an additional experiment.

We aim to design an evaluation method applicable to all the models proposed in this paper. HD+HMR models are straightforward to adapt by removing the HD component and using ground truth bounding boxes (bbox). However, density-based models require additional steps. For these models, we predict densities for the entire image, then crop the corresponding ground truth bboxes for the individuals involved, isolating their respective volumes. For all models, we consider only non-overlapping bboxes to prevent volume duplication and report the resulting PPMAE.

To avoid penalizing models for detection errors, we exclude any predicted volumes under 10 dm³ from being counted as positives. Additionally, to facilitate easier detection, we removed scenes from the ANTHROPOS-V test set that contain significant occlusions or challenging lighting conditions, creating a refined test set called S1.

Furthermore, we constructed an additional test set, S2, consisting exclusively of bird’s-eye view scenes, which minimize occlusion and further simplify detection.

\begin{table}[ht]
\caption{Results on \ourdataset's S1, S2 and whole test set (FT). All the results are reported in dm${^3}$.} \label{tab:decouple}
\centering
\resizebox{0.9\linewidth}{!}{
\begin{tabular}{l|cc|c}
\toprule
\textbf{Model} & \textbf{PPMAE (S1)} & \textbf{PPMAE (S2)} & \textbf{PPMAE (FT)}\\
\midrule
ReFit~\cite{wang2023refit} & 17.94 & 18.25 & 18.79\\
\midrule
STEERER~\cite{hani2023steerer} & 12.46 & 13.68 & 14.43 \\
\midrule
STEERER-V  & 6.67 & 3.39 & 6.73 \\
\bottomrule
\end{tabular}}
\end{table}

In Table~\ref{tab:decouple}, we present the results of this experiment with simplified detection. Notably, ReFit and STEERER show error levels comparable to those on the full test set (FT), highlighting that the primary source of their error lies in evaluating individuals' volume. In contrast, STEERER-V’s performance improves by 50\% on the easier detection set (S2), suggesting that its error is equally divided between detection and volume estimation. Moreover, as shown in Table~1 of the main paper, when comparing both $C(I)_{B+} \times \overline{V}_D$ and ReFit with their oracular counterparts, a similar ratio emerges, further confirming that volume estimation error accounts for half of the total error.

\section{From Frames to Video} \label{temporal} Lastly, we question if leveraging temporal information can be beneficial in CVE. Specifically, we modify STEERER-V to leverage two neighboring context-frames, one before and one after the target frame. We align features from context-frames to those of the target frame using the method in \cite{huang2024scale} and feed the result into STEERER-V's decoding branch to estimate the total volume in the target frame. We use STEERER-V's pretrained weights, while the feature alignment module is trained from scratch. Despite being an initial attempt to incorporate inter-frame information, this approach proves beneficial for CVE, reducing MAE by 5.27\% and PPMAE by 4.22\%.

{\small
\bibliographystyle{ieee_fullname}
\bibliography{main}
}